%% file: main.tex
\documentclass[11pt]{article}
\usepackage{palatino}
\usepackage[letterpaper,margin=1in]{geometry}

\usepackage{microtype}
\usepackage{graphicx}
\usepackage{subfigure}
\usepackage{booktabs} 

\usepackage{hyperref}

\usepackage{amsmath}
\usepackage{amssymb}
\usepackage{mathtools}
\usepackage{amsthm}
\usepackage{thmtools, thm-restate}
\usepackage{bigints}
\usepackage[capitalize,noabbrev]{cleveref}

\usepackage{amsopn,algorithm,algorithmic,float,bbm,bm,enumerate,color,multirow,gensymb}

\usepackage{epsfig,subfigure,graphicx}
\usepackage{multirow}
\usepackage[]{color-edits}
\usepackage{epsfig,subfigure,graphicx}
\usepackage{comment}
\usepackage{authblk}
\usepackage{afterpage}
\usepackage{multirow}
\allowdisplaybreaks[1]
\usepackage{natbib}

\usepackage[dvipsnames]{xcolor}

\addauthor{ab}{blue}
\addauthor{abb}{red} 
\addauthor{yz}{brown}
\addauthor{tc}{teal}
\input{ICML2022/notation}
\title{Stability Based Generalization Bounds \\for Exponential Family Langevin Dynamics}
\author[1]{Arindam~Banerjee}
\author[2]{Tiancong~Chen}
\author[2]{Xinyan~Li}
\author[2]{Yingxue~Zhou}

\affil[1]{Department of Computer Science,
University of Illinois Urbana-Champaign}
\affil[2]{Department of Computer Science \& Engineering, 
University of Minnesota, Twin Cities}
\affil[ ]{Emails: {\normalsize {\sf arindamb@illinois.edu, \{chen6271,lixx1166,zhou0877\}@umn.edu}}}
\date{}

\begin{document}
\maketitle

\begin{abstract}
\input{arXiv2022/sec/abstract}
\end{abstract}

\section{Introduction}
\label{sec:intro}

\input{arXiv2022/sec/intro}

\section{Expected Stability based Generalization}
\label{sec:exp_stability}
\input{arXiv2022/sec/exp_stability}

\section{Exponential Family Langevin Dynamics}
\label{sec:efld}
\input{arXiv2022/sec/efld}

\section{Experiments}
\label{sec:exp}

\input{arXiv2022/sec/exp}

\section{Conclusions}
\label{sec:conc}
\input{arXiv2022/sec/conc}

{\bf Acknowledgements.} The research was supported by NSF grants IIS 21-31335, OAC 21-30835, DBI 20-21898, and a C3.ai research award. We would like to thank the reviewers for valuable comments and the Minnesota Supercomputing Institute (MSI) for computational resources and support.

\bibliographystyle{apalike}
\bibliography{ICML2022/reference}

\newpage
\appendix
\onecolumn

\section{Related Work}
\label{sec:related_work}
\input{arXiv2022/sec/related}

\section{Analysis and Proofs for Expected Stability (Section~\ref{sec:exp_stability})}
\label{sec:app_exp_stability}
\input{arXiv2022/app/exp_stability}

\section{Analysis and Proofs for EFLD (Section~\ref{sec:efld})}
\label{sec:app_efld}
\input{arXiv2022/app/efld}

\section{Optimization Guarantees for EFLD}
\label{app:opt}

\input{arXiv2022/app/signSGD_opt}

\section{Experiment Details}
\label{app:exp}
\input{arXiv2022/app/app_exp}

\end{document}

%% file: ICML2022/notation.tex
\newcommand{\beq}{\begin{equation}}
\newcommand{\eeq}{\end{equation}}

\newcommand\E{\mathbb{E}}
\renewcommand\P{\mathbb{P}}
\newcommand\I{\mathbb{I}}

\newcommand\R{\mathbb{R}}

\newcommand\1{\mathbbm{1}}

\newcommand{\g}{\mathbf{g}}

\renewcommand{\v}{\mathbf{v}}

\newcommand{\w}{\mathbf{w}}

\newcommand{\p}{\mathbf{p}}

\newcommand{\cN}{{\cal N}}

\newcommand{\cT}{{\cal T}}
\newcommand{\cW}{{\cal W}}

\newcommand{\cZ}{{\cal Z}}

\newcommand{\vertiii}[1]{{\left\vert\kern-0.25ex\left\vert\kern-0.25ex\left\vert #1
    \right\vert\kern-0.25ex\right\vert\kern-0.25ex\right\vert}}

\newcommand{\m}{\boldsymbol{\mu}}
\newcommand{\bmu}{\boldsymbol{\mu}}

\newcommand{\btheta}{\boldsymbol{\theta}}

\newcommand{\ppi}{\boldsymbol{\pi}}

\newcommand{\aalpha}{\boldsymbol{\alpha}}

\newcommand{\bxi}{\boldsymbol{\xi}}

\DeclareMathOperator{\gen}{gen}

\newtheorem{theo}{Theorem}
\newtheorem{prop}{Proposition}
\newtheorem{lemm}{Lemma}

\newtheorem{asmp}{Assumption}

\theoremstyle{definition} 
\newtheorem{exam}{Example}[section]
\newtheorem{remark}{Remark}[section]

\newcommand {\commentout}[1] {}

\def\ints{{{\rm Z} \kern -.35em {\rm Z} }}  
\def\smallints{{{\rm Z} \kern -.3em {\rm Z} }}  
\def\pints{{{\rm I} \kern -.15em {\rm N} }}      
\newcommand{\reals}{\mathbb R}

\newcommand{\RR}{\mathbb R}
\def\cplx{{{\rm I} \kern -.45em {\rm C} }}       
\def\l2{\rm {\mathcal L}^{2}(\reals)}            

\newcommand{\be}{\begin{eqnarray}}
\newcommand{\ee}{\end{eqnarray}}
\newcommand{\bea}{\begin{eqnarray}}
\newcommand{\eea}{\end{eqnarray}}
\newcommand{\beaa}{\begin{eqnarray*}}
\newcommand{\eeaa}{\end{eqnarray*}}
\newcommand{\bnad}{\begin{nad}}
\newcommand{\enad}{\end{nad}}

%% file: arXiv2022/sec/abstract.tex
Recent years have seen advances in generalization bounds for noisy stochastic algorithms,
especially stochastic gradient Langevin dynamics (SGLD) based on stability~\citep{mou2018generalization, Li2020On} and information theoretic approaches~\citep{xu2017information,negrea2019information,steinke2020reasoning}. 
In this paper, we unify and substantially generalize stability based generalization bounds and make three technical contributions. First, we bound the generalization error  in terms of expected (not uniform) stability which arguably leads to quantitatively sharper bounds.
Second, as our main contribution, we introduce Exponential Family Langevin Dynamics (EFLD), a substantial generalization of SGLD, which includes noisy versions of Sign-SGD and quantized SGD as special cases.
We establish data-dependent expected stability based generalization bounds for any EFLD algorithm with a $O(1/n)$ sample dependence and  dependence on gradient discrepancy rather than the norm of gradients, yielding significantly sharper bounds.
Third, we establish optimization guarantees for special cases of EFLD.
%
Further, empirical results on benchmarks
illustrate that our bounds are non-vacuous, quantitatively sharper than existing bounds, and behave correctly under noisy labels.

%% file: arXiv2022/sec/intro.tex
Recent years have seen renewed interest in characterizing generalization performance of learning algorithms in terms of stability which considers change in performance of a learning algorithm based on change of a single training point \citep{hardt2016train,bousquet2002stability,Li2020On,mou2018generalization}. 
For stochastic gradient descent (SGD), \citet{hardt2016train} established generalization bounds based on uniform stability \citep{bousquet2002stability}, although the analysis needed rather small step sizes $\eta_t = 1/t$ which is arguably not useful in practice. While improving the stability analysis for SGD has remained a challenge, advances have been made on noisy SGD algorithms, especially stochastic gradient Langevin dynamics (SGLD) \citep{WellingT11, mou2018generalization,Li2020On} which adds Gaussian noise to the stochastic gradients. In parallel, there has been key developments on related information-theoretic generalization bounds applicable to SGLD type algorithms \citep{negrea2019information,haghifam2020sharpened,xu2017information,russo2016controlling, pensia2018generalization, wang2021analyzing}.

While these developments have led to advances in analyzing generalization of noisy SGD algorithms, and we elaborate on these developments in Appendix \ref{sec:related_work}, they each have certain limitations, e.g., dependence on global Lipschitz constant $L$ \citep{mou2018generalization}, tiny $O(1/L)$ step sizes \citep{Li2020On}, $O(1/\sqrt{n})$ sample dependence \citep{mou2018generalization,negrea2019information}, dependence on gradient norms \citep{Li2020On}, restrictions on nature of mini-batching \citep{wang2021analyzing}, etc. Further, most prior work primarily focuses on SGLD and cannot be readily extended to popular variants such as noisy versions of Sign-SGD or quantized SGD \citep{bernstein2018signsgd,quantized-sgd,jiang2018}.
In this paper, we build on the core strengths of such existing approaches, most notably (a) the $O({1}/{n})$ sample dependence of stability based bounds~\citep{mou2018generalization,Li2020On}, (b) the dependence on
some measures of gradient discrepancy rather than the norm of gradients~\citep{negrea2019information,haghifam2020sharpened}, and (c) no dependence on the global Lipschitz constant $L$, and develop a framework (Section~\ref{sec:exp_stability}) for establishing generalization bounds for a general family of noisy stochastic iterative algorithms which includes SGLD as a special case. Our framework considers generalization based on the concept of {\em expected stability}, rather than uniform stability \citep{hardt2016train,bousquet2002stability, bousquet2020sharper,mou2018generalization,farghly2021timeindependent}, and yields distribution dependent generalization bounds which avoid the worst-case setting of uniform stability. Recall that for any data domain $\cZ$ and a distribution $D$ over the domain, uniform stability considers the worst case difference in loss over two datasets $S_n,S'_n \in \cZ^n$ of size $n$ which differ by one point, i.e., over $\sup_{S_n, S'_n, |S_n \Delta S'_n|=1} \cdots$ \citep{bousquet2002stability,hardt2016train}. In Section~\ref{sec:exp_stability}, we show that one gets a valid generalization bound for stochastic algorithms  by replacing the supremum $\sup$ by an expectation $\E_{S_n,S'_n}$, where $S_n \sim D^n$ and, without loss of generality, $S'_n$ shares the first $(n-1)$ samples with $S_n$ with the $n$-th sample $z'_n \sim D$. Replacing $\sup$ by $\E$ makes the bound distribution dependent
and arguably leads to quantitatively sharper and computable bounds with less assumptions. 
Further, we show that expected stability of general noisy stochastic iterative algorithms
can be bounded by the expectation of a Le Cam Style Divergence (LSD) between  distributions over parameters obtained from $S_n$ and $S'_n$. Thus, getting an expected stability based generalization bound for a specific stochastic algorithm reduces to that of bounding the expected LSD.

In Section~\ref{sec:efld}, we introduce Exponential Family Langevin Dynamics (EFLD), a family of noisy stochastic gradient descent algorithms based on exponential family noise. Special cases of EFLD include SGLD and noisy versions of Sign-SGD or quantized SGD algorithms~\citep{bernstein2018signsgd, bernstein2018signsgdnoise, jin2020stochastic, quantized-sgd}. Our main result provides an expected stability based generalization bound for {\em any} EFLD algorithm with several aforementioned desirable properties: (a) a $O({1}/{n})$ sample dependence, (b) a dependence on the gradient discrepancy, a variant of gradient incoherence~\citep{negrea2019information}, rather than a dependence on the norm of gradients, (c) no dependence on the global Lipschitz constant $L$, and (d) step sizes $\eta_t$ need not be tiny, i.e., $\eta_t = O(1/L)$ is not needed. Existing generalization bounds for SGLD~\citep{Li2020On,negrea2019information} usually use properties of the Gaussian distribution, and do not generalize to EFLD. Our proof technique is new, and uses properties of exponential family distributions. 
%
We also provide optimization guarantees for EFLD, i.e., convergence results for noisy Sign-SGD and 
SGLD.

In Section~\ref{sec:exp}, we present experimental results on benchmark datasets. We illustrate that our bounds for SGLD are non-vacuous and quantitatively tighter than existing bounds~\citep{Li2020On,negrea2019information} due to the desirable dependence on sample size and gradient discrepancy norms, which are empirically shown to be orders of magnitude smaller than gradient norms. We also report results on random labels~\cite{zhang2017understanding} where as the training error goes to zero, our bound correctly captures the increase in generalization error due to increasing fraction of random labels. We also present results for Noisy Sign-SGD and illustrate that our bounds give a quantitatively tight upper bound on the empirical test error across epochs. 

%% file: arXiv2022/sec/exp_stability.tex
In the setting of statistical learning, there is an instance space $\cZ$, a hypothesis space $\cW$, and a loss function $\ell : \cW \times \cZ \mapsto \R_+$. Let $D$ be an unknown distribution of $\cZ$ and let $S_n \sim D^n$ be $n$ i.i.d.~draws from $D$. For any specific hypothesis $\w \in \cW$, the population and empirical loss are respectively given by
\begin{equation}
L_D(\w) \triangleq \E_{z \sim D}[\ell(\w,z)]~, \qquad \text{and} \qquad L_S(\w) \triangleq \frac{1}{n} \sum_{i=1}^n \ell(\w,z_i)~.  
\end{equation}
For any distribution $P$ over the hypothesis space, we respectively denote the expected population and empirical loss as
\begin{equation}
L_D(P) \triangleq \E_{z \sim D}\E_{\w \sim P}[\ell(\w,z)]~, \quad \text{and} \quad  L_S(P) \triangleq \frac{1}{n} \sum_{i=1}^n \E_{\w \sim P}[\ell(\w,z_i)]~.
\end{equation}
We consider a randomized algorithm $A$ which works with $S_n =\{z_1,\ldots,z_n\} \sim D^n$ and creates a distribution over the hypothesis space $\cW$. For convenience, we will denote the distribution as $A(S_n)$. The focus of our analysis is to bound the generalization error of $A$ defined as: 
\begin{align}
\gen(A(S_n)) \triangleq L_D(A(S_n)) - L_S(A(S_n))~. 
\end{align} 
We will assume $A$ is permutation invariant, i.e., the ordering of samples in $S_n$ does not modify $A(S_n)$, an assumption satisfied by most learning algorithms. 
All technical proofs for results in this section are in Appendix~\ref{sec:app_exp_stability}.

\subsection{Bounds based on Expected Stability}
We start our analysis by noting that the expected generalization error can be upper bounded by {\em expected stability} based on the Hellinger divergence $H(P \| P')$ between two distributions given by~\citep{save16,Li2020On}: $H^2(P\| P') = \frac{1}{2}\int_{\w} (\sqrt{p(\w)} - \sqrt{p'(\w)})^2 d\w$.
\begin{restatable}{prop}{prophell}
Let $S_n \sim D^n$ and let $S'_{n}$ be a dataset obtained by replacing $z_n \in S_n$ with $z'_n \sim D$. Let $A(S_n),A(S'_{n})$ respectively denote the distributions over the hypothesis space $\cW$ obtained by running randomized algorithm $A$ on $S_n,S'_{n}$.
Assume that for all $S_n \in \cZ^n, z \in \cZ$, $\E_{W \sim  A(S_n)}[\ell^2(W,z)] \leq \frac{c^2_0}{4}$ for some constant $c_0>0$.  With $H(\cdot,\cdot)$ denoting the Hellinger divergence, we have
\begin{equation}
\begin{aligned}
| \E_{S_n \sim D^n}[L_D(A(S_n)) - L_S(A(S_n))] | ~\leq ~ c_0 \E_{S_n \sim D^n} \E_{z'_n \sim D} \sqrt{2 H^2\big(A(S_n), A(S'_{n})\big)}~. 
\end{aligned}
\end{equation}
\label{prop:hell}
\end{restatable}
\begin{remark}
Proposition~\ref{prop:hell} does not need bounded losses, only the {\em second moment} of $\ell(W,z), W \sim A(S_n), \forall S_n, z$ needs to be bounded. In comparison, recent information theoretic bounds~\citep{haghifam2020sharpened,xu2017information} assume $\ell(\w,Z), Z \sim D, \forall \w \in \cW$ to be {\em sub-Gaussian}. Note that these assumptions are satisfied by bounded losses. \qed 
\end{remark}
\begin{remark}
The bound in Proposition~\ref{prop:hell} is in terms of {\em expected stability} where we consider $\E_{S_n \sim D^n} \E_{z'_n \sim D}[\cdots]$, an important departure from bounds based on {\em uniform stability} \citep{elisseeff2005stability, bousquet2002stability, mou2018generalization,bousquet2020sharper}
where one considers $\sup_{S_n,S_n' \in \cZ^n, |S_n \Delta S_n'| = 1}[\cdots]$. 
Replacing $\sup$ by $\E$ makes the bounds distribution dependent, avoids the worst case analysis associated with uniform stability, and arguably leads to quantitatively tighter bounds. \qed 
\end{remark}

\subsection{Expected Stability of Noisy Iterative Algorithms}
\label{sec: exp_stab_nsi}
We consider a general family of noisy stochastic iterative (NSI) algorithms. Given $S_n \sim D^n$, such iterative algorithms have two (additional) sources of randomness in each iteration $t$: 
\begin{enumerate}[(a)] 
\item a stochastic mini-batch of samples $S_{B_t}$, with $|S_{B_t}| = b \leq n$, drawn uniformly at random with replacement from $S_n$; and 
\item noise $\bxi_t$ suitably included in the iterative update. 
\end{enumerate}
In our exposition, $B$ will denote a subset of indices to samples and $S_B$ will denote the corresponding mini-batch of samples based on the subset of indices in $B$. In (a) above, $B_t \subseteq [n]$ with $|B_t|=b$ and $S_{B_t} \subseteq S_n$ with $|S_{B_t}|=b$.

Given a trajectory (realization) of past iterates $W_{0:(t-1)} = \w_{0:(t-1)}$, the new iterate $W_t$ is drawn from a distribution $P_{B_t,\bxi_t|\w_{0:(t-1)}}$ over $\cW$:
\begin{align}
\label{eq:nmia}
W_{t} \sim P_{B_t,\bxi_t|\w_{0:(t-1)}}(W) ~.
\end{align}
We will often drop conditioning $\w_{0:(t-1)}$ to avoid clutter. 

Let $\bar{S}_{n+1} \sim D^{n+1}$ with $\bar{S}_{n+1} = \{z_1,\ldots,z_{n+1}\}$. Let $S_0 = \{z_1,\ldots,z_{n-1}\}$. $S_n,S'_{n}$ are size $n$ subsets of $\bar{S}_{n+1}$ with $S_n= S_0 \cup \{z_n\}$ and $S'_{n} = S_0 \cup \{z'_n\}$, where $z'_n = z_{n+1}$. 
The algorithms we consider use a mini-batch of size $b$ in each iteration uniformly sampled from $S_n$ or $S'_n$. Let $G$ denote the set of all mini-batch index subsets of size $b$ that can be drawn from $S_{n}$,
$G_0$ denote the set of all mini-batch index subsets of size $b$ that can be drawn from $S_0$, and
$G_1$ denote the set of all mini-batch index subsets of size $b$ that can be drawn from $S_n$ which includes the last sample~$z_n$.
Formally, with $2^{n]}$ denoting the set of all subsets of $[n]=\{1,\ldots,n\}$
\begin{align}
G & \triangleq \left\{ B \subseteq [n] \mid |B|=b, S_B \subseteq S_n \right\}~, \\
G_0 & \triangleq \left\{ B \subseteq [n] \mid |B|=b, S_B \subseteq S_0 \right\}~, \\
G_1 & \triangleq \left\{ B \subseteq [n] \mid |B|=b, S_B \subseteq S_n, z_n \in S_B \right\}~.
\end{align}
Note that $|G_0| = \binom{n-1}{b}$, $|G_1| = \binom{n-1}{b-1}$, and 
$|G_0| + |G_1| = \binom{n-1}{b} + \binom{n-1}{b-1} = \binom{n}{b} = |G|$. Further, note that one can replace $S_n$ in the definition of $G, G_1$ with $S'_n$ when analyzing a stochastic algorithm run on $S'_n$, and the equation $|G_0|+|G_1|=|G|$ stays the same.

Based on \eqref{eq:nmia}, let $P_{0:(t-1)}$ denote the joint distribution over $W_{0:(t-1)} = (W_0,\ldots,W_{t-1})$, and 
let $P_{t|} := P_{B_t,\bxi_t|\w_{0:(t-1)}}$ compactly denote the conditional distribution on $W_t$. 
Let $P_T, P'_T$ denote the marginal distributions over $W \in \cW$ after $T$ steps of the algorithm based on $S_n,S'_{n}$ respectively. For randomized algorithms of the form \eqref{eq:nmia}, from Proposition \ref{prop:hell} we first bound the Hellinger divergence with KL-divergence, i.e., $2H^2(P_T,P'_T) \leq KL(P_T,P'_T)$ (Proposition~\ref{prop:hellkl} in Appendix~\ref{sec:app_exp_stability}), and then 
use the following chain rule \citep{pensia2018generalization,negrea2019information,haghifam2020sharpened} to bound the KL-divergence between $P_T$ and $P'_T$:
\begin{align}
KL(P_T \| P'_T) \leq KL(P_{0:T} \| P'_{0:T} ) = \sum_{t=1}^T \E_{P_{0:(t-1)}}\left[ KL(P_{t|} \| P'_{t|}) \right]~.
\end{align}
%
We can bound the per-step conditional KL-divergences $KL(P_{t|} \| P'_{t|})$ 
in terms of a Le Cam Style Divergence (LSD). While the classical Le Cam divergence~\citep{save16} is $LSD(P\|P') \triangleq \frac{1}{2} \int \frac{(dP - dP')^2}{dP + dP'}$ (where $dP$ denotes the density), our bounds are in terms of
\begin{equation}
\begin{split}
\label{eq:lsd}
LSD(P_{t|}| \| P'_{t|}) & := \underset{B_t \in G_1}{\E} \underset{A_t \in G_0}{\E}  \left[ \Lambda(B_t,A_t) \right]~, \\
\text{where} \quad \Lambda(B_t,A_t) & := \int_{\bxi_t} \frac{(dP_{B_t,\bxi_t} - dP'_{B_t,\bxi_t})^2}{dP_{A_t,\bxi_t}} d \bxi_t ~.
\end{split}
\end{equation}
Note that $P_{B_t,\bxi_t}$ and $P'_{B_t,\bxi_t}$ represent the conditional distribution of $W_t$ for $S_n$ and $S_n^\prime$ respectively since the mini-batch $S_{B_t}$ of $S_n$ and $S_n^\prime$ differs in the $n$-th sample. 
Then, we have the following LSD based generalization bound.
\begin{restatable}{lemm}{theostab}
In the setting of Proposition~\ref{prop:hell} consider a noisy stochastic iterative algorithms of the form~\eqref{eq:nmia} with mini-batch size $b \leq n/2$. Then, with $c_1 = \sqrt{2} c_0$ (with $c_0$ as in Proposition~\ref{prop:hell}) and $\Lambda(B_t,A_t)$ as in \eqref{eq:lsd}, we have
\begin{equation}
\begin{aligned}
| \E_{S_n}[L_D(A(S_n)) - L_{S}(A(S_n))] | ~\leq ~c_1 \frac{b}{n} \E_{S_n} \E_{z'_n} \sqrt{ \sum_{t=1}^T  \underset{W_{0:(t-1)}}{\E} \underset{B_t \in G_1}{\E} \underset{A_t \in G_0}{\E} \left[ \Lambda(B_t,A_t) \right] }~.
\end{aligned}
\end{equation}
\label{theo:stab}
\end{restatable}
\begin{remark}
Though not stated explicitly, \citet{Li2020On} essentially has this result for SGLD and inspired our work. Our proofs are significantly simpler, does not make any additional assumptions, and illustrates applicability to general noisy iterative algorithms of the form~\eqref{eq:nmia} not just SGLD with Gaussian noise as in \cite{Li2020On}. \qed 
\end{remark}
\begin{remark}
The bound depends on expectations over samples $S_n,z'_n$, trajectories $W_{0:(t-1)}$, and mini-batches $B_t,A_t$. Unlike uniform stability and other worst case analysis, there is no $\sup$ over samples, trajectories, or mini-batches. \qed 
\end{remark}
\begin{remark}
The bound seems to worsen with $b$, the size of the mini-batch. As we show in Section~\ref{sec:efld}, the LSD terms $\Lambda(\cdot,\cdot)$ have a $\frac{1}{b^2}$ dependence for SGLD and its generalizations we introduce, so the leading $b$ is neutralized. \qed 
\end{remark}
\begin{remark}
A high probability version of the result based on an exponential version of the Efron-Stein inequality~\cite{boucheron2013concentration} is presented in Appendix~\ref{ssec:highp}. \qed 
\end{remark}


%% file: arXiv2022/sec/efld.tex

Recent years have seen advances in establishing generalization bounds for SGLD \citep{Li2020On, pensia2018generalization, negrea2019information, haghifam2020sharpened} which adds isotropic Gaussian noise at every step of SGD:
\begin{align}
\label{eq:sgld}
\w_{t+1} = \w_{t} - \eta_t \nabla \ell(\w_t, S_{B_t}) + \mathcal{N}\left(0, \sigma_t^2 \mathbb{I}\right)~,
\end{align}
where $ \nabla \ell(\w_t, S_{B_t})$ is the stochastic gradient on mini-batch $B_t$, $\eta_t$ is the step size, and $\sigma_t^2$ is noise vairance.  We introduce a substantial generalization of SGLD called Exponential Family Langevin Dynamics (EFLD) which uses general exponential family noise in noisy iterative updates of the form~\eqref{eq:nmia}. In addition to being a mathematical generalization of the popular SGLD, the proposed EFLD provides flexibility to use noisy gradient algorithms with different representation of the gradient, e.g., skewed Rademacher noise for Sign-SGD, discrete distribution for quantized or finite precision SGD, etc. \citep{canonne2020discrete, quantized-sgd, jiang2018, yang2019swalp}. All technical proofs for results in this section are in Appendix~\ref{sec:app_efld}.

\subsection{Exponential Family Langevin Dynamics (EFLD)}\label{subsec:efld}

Exponential families \citep{barndorff2014information,brown1986fundamentals,wainwright2008graphical} constitute a large family of parametric distributions which include Gaussian, Bernoulli, gamma, Poisson, Dirichlet, etc., as special cases. Exponential families are typically represented in terms of natural parameters $\btheta$, and we consider component-wise independent distributions with scaled natural parameter $\btheta_{\alpha} = \btheta/\alpha$ with scaling $\alpha > 0$, i.e.,
\begin{align*}
p_{\psi}(\bxi, \btheta_{\alpha})  = \exp(\langle \bxi, \btheta_{\alpha} \rangle - \psi(\btheta_{\alpha})) \ppi_{0,\alpha}(\bxi)  = \prod_{j=1}^p \exp( \xi_j \theta_{j\alpha} - \psi_j(\theta_{j\alpha})) \pi_{0,\alpha}(\xi_j)~,
\vspace*{-2mm}
\end{align*}
where $\bxi \in \R^p$ is the sufficient statistic, $\psi(\btheta_{\alpha})= \sum_{j=1}^p \psi_j(\theta_{j\alpha})$ is the log-partition function, and $\ppi_{0,\alpha}(\xi) = \prod_{j=1}^p \pi_{0,\alpha}(\xi_j)$ is the base measure. 
$\psi$ is a smooth convex function by construction \citep{barndorff2014information,banerjee2005clustering,wainwright2008graphical} which implies $\nabla_{\btheta_\alpha}^2 \psi(\btheta_{\alpha}) \leq c_2 \I$ for some constant $c_2 > 0$.

Exponential family Langevin dynamics (EFLD) uses noisy stochastic gradient updates similar to SGLD, but using exponential family noise rather than Gaussian noise as in SGLD. In particular, for mini-batch $S_{B_t}$, EFLD updates are as follows: with step size $\rho_t > 0$
\begin{equation}
    \w_t = \w_{t-1} - \rho_t \bxi_t~, \qquad \bxi_t \sim p_{\psi}(\bxi; \btheta_{B_t,\alpha_t})~,
\vspace{-3mm}    
\label{eq:efld1}
\end{equation}
where 
\begin{align}
p_{\psi}(\bxi; \btheta_{B_t,\alpha_t})  = \exp(\langle \bxi, \btheta_{B_t,\alpha_t} \rangle - \psi(\btheta_{B_t,\alpha_t})) \ppi_{0,\alpha}(\bxi)   ~,~
\btheta_{B_t,\alpha_t}  \triangleq \frac{\btheta_{B_t}}{\alpha_t} = \frac{\nabla \ell(\w_{t-1}, S_{B_t})}{\alpha_t}~. 
\label{eq:efld2}
\vspace*{-2mm}
\end{align}
For EFLD, the natural parameter $\btheta_{B_t,\alpha_t}$ at step $t$ is simply a scaled version of the mini-batch gradient $\nabla \ell(\w_{t-1}, S_{B_t})$. 
EFLD becomes SGLD when the exponential family is {\em Gaussian}~\cite{Li2020On}. 
EFLD becomes noisy sign-SGD \citep{bernstein2018signsgd, bernstein2018signsgdnoise} when the exponential family is a {\em skewed Rademacher distribution} over $\{-1,+1\}$ with $P(\xi_j=+1) = \frac{\exp(\nabla \ell_j)}{\exp(-\nabla \ell_j) + \exp(\nabla \ell_j)}, P(\xi_j=-1)= 1 - P(\xi_j=+1)$ where $\nabla \ell_j = [\nabla \ell(\w_{t-1}, S_{B_t})/\alpha_t]_j$ which becomes Sign-SGD as $\alpha_t \rightarrow 0$. We briefly discuss the case of SGLD here and discuss additional examples including skewed Rademacher (noisy sign-SGD) and Bernoulli in Appendix \ref{app:example}.

\begin{exam}[Gaussian] 
From the EFLLD perspective, SGLD uses scaled Gaussian noise with $\psi(\btheta)= \|\btheta\|_2^2/2, \btheta_{\alpha} = \btheta/\alpha, \alpha = \sigma/\eta$, 
$\ppi_{0,\alpha}(\bxi) = \frac{1}{\sqrt{(2\pi)^p \alpha^p}} \exp(-\|\bxi\|_2^2/2\alpha^2)$ so that $p_{\psi}(\bxi; \btheta_{B,\alpha}) = \mathcal{N}(\btheta_{B}, \alpha^2 \I_d)$. 
In particular, the distribution from the natural parameter form is:
\begin{equation}
    p_{\btheta/\alpha}(\xi) = \exp( \langle\xi, \btheta\rangle / \alpha^2 - \|\theta\|_2^2/(2\alpha^2) ) \times \frac{1}{\sqrt{2\pi}\alpha} \exp(-\|\xi\|_2^2/2\alpha^2)
    = \frac{1}{\sqrt{2\pi}\alpha} \exp(-\|\bxi-\bmu\|_2^2/2\alpha^2)~,
\end{equation}
where the expectation parameter $\bmu = \nabla \psi(\btheta) = \btheta$.
By choosing stepsize $\rho_t = \eta_t$ in the update in \eqref{eq:efld1}, $\rho_t \bxi_t$ is distributed as $\mathcal{N}(\eta_t \btheta_{B_t}, \eta_t^2 \alpha_t^2 \I_d) = \mathcal{N}(\eta_t\nabla \ell(\w_{t-1}, S_{B_t}), \sigma_t^2 \I_d)$ since $\eta_t \alpha_t = \sigma_t$. Thus the EFLD update in \eqref{eq:efld1} reduces to the SGLD update: 
\begin{align*}
\w_{t} = \w_{t-1} - \eta_t \nabla \ell(\w_{t-1}, S_{B_t}) + \mathcal{N}\left(0, \sigma_t^2 \mathbb{I}_{d}\right)~,
\end{align*}
illustrating that SGLD is a special case of EFLD. \qed 
\label{exam:sgld}
\end{exam}

%


\subsection{Expected Stability of EFLD}
\input{arXiv2022/sec/stability}

\subsection{Optimization Guarantees for EFLD}

We now establish optimization guarantees for two examples of EFLD, i.e., Noisy Sign-SGD with skewed Rademacher noise over $\{-1,+1\}$ and SGLD with Gaussian noise. The details and the proof for results in this subsection are relegated in Appendix \ref{app:opt}.

\noindent {\bf Noisy Sign-SGD.} For noisy Sign-SGD with mini-batch $B_t$ and scaling $\alpha_t$, mini-batch Noisy Sign-SGD updates as $\w_{t} = \w_{t-1} - \eta_t \bxi_t$, where each component $j\in [p]$
\begin{equation*}\label{eq:signSGD_mini_batch}
\bxi_{t,j} \sim p_{\btheta_{B_t,\alpha_t,j}}(\xi_j) =  \frac{\exp(\xi_j \theta_{B_t,\alpha_t,j})}{\exp(-\theta_{B_t,\alpha_t,j})+\exp(\theta_{B_t,\alpha_t,j})},
\end{equation*}
where $\xi_j\in\{-1,+1\}$ and $\btheta_{B_t,\alpha_t} = \nabla \ell(\w_{t-1}, S_{B_t})/\alpha_t$ is the scaled mini-batch gradient.
The full-batch version uses parameters $\btheta_{B_t,\alpha_t} = \nabla L_S(\w_{t-1})/\alpha_t$.
For full batch gradient descent, we assume that the loss is smooth.
\begin{restatable}{asmp}{asmpsmooth}\label{asmp: smooth}
The loss function $L_S(\w) = \frac{1}{n} \sum_{i=1}^n \ell(\w,z_i)$ satisfies: $\forall \w, \w^\prime$, for some non-negative constants $\vec{K}:=\left[K_{1}, \ldots, K_{p}\right]$, we have
$L_S(\w) \leq L_S(\w^\prime) + \nabla L_S(\w^\prime)^T(\w - \w^\prime) + \frac{1}{2} \sum_{i} K_i (w_i - w^\prime_i)^2$. 
\end{restatable}

For mini-batch analysis, we assume the mini-batch gradients are unbiased, symmetric, and sub-Gaussian.
\begin{restatable}{asmp}{asmpbatch}\label{asmp:batch}
Given $\w_{t-1}$, the mini-batch gradient $\nabla \ell(\w_{t-1}, S_{B_t})$ is
(a)  unbiased,  i.e., $\E_{B_t|\w_{t-1}}\nabla \ell(\w_{t-1}, S_{B_t}) = \nabla L_S(\w_{t-1})$; (b) symmetric, i.e., the density $p_{B_t|\w_{t-1}}(\bxi)$ of $\bxi \equiv \nabla \ell(\w_{t-1}, S_{B_{t}})$ is symmetric and 
(c) sub-Gaussian, i.e., for any $\lambda >0$, {any $\v$ s.t. $\|\v\|_2 =1$}, $
\E_{B_t|\w_{t-1}}[\exp \lambda\langle \v,\nabla \ell(\w_{t-1}, S_{B_{t}})-\nabla  L_S(\w_{t-1})\rangle] \leq \exp(\lambda^2\kappa_t^2/2)~,
$
for some constant $\kappa_t > 0$.
\end{restatable}
The smoothness assumption in Assumption \ref{asmp: smooth} is standard in non-convex optimization especially for sign-SGD literature \citep{bernstein2018signsgd,bernstein2018signsgdnoise}. Assumption \ref{asmp:batch} for the mini-batch setting helps the theoretical analysis, where (a) is satisfied when the batches $S_{B_t}$ are taken uniformly from samples $S$ as the standard training does; (b) assumes symmetry of the mini-batch gradients; and (c) is similar and stronger assumption compared to Assumption 3 in \citet{bernstein2018signsgd}, where they assume bounded variance for stochastic gradient and our assumption implies suitably bounded higher moments of $\nabla \ell(\w_{t-1}, S_{B_{t}})-\nabla  L_S(\w_{t-1})$, which is referred as the minibatch noise in recent noisy SGD literature e.g. \citet{damian2021label}. 
Similar to such literature, if we consider mini-batch stochastic gradient be modeled as the average of $|B_t|$ calls to the full-batch gradient, from concentration property $\kappa_t$ is scaled by $1/\sqrt{|B_t|}$. 

Based on the assumptions, we have following optimization guarantee for mini-batch noisy Sign-SGD, the full-batch version can be found in Appendix \ref{app:opt}.

\begin{restatable}{theo}{theosignSGDconv}\label{theo:signSGD}
The following holds for any $S$, any initialization $\w_0$, and the expectation is taken over the randomness of algorithm: if Assumption~\ref{asmp: smooth} and \ref{asmp:batch} hold, for mini-batch noisy Sign-SGD with step size $\eta_t = 1/\sqrt{T}$, and $\alpha_t$ satisfying $c \geq \alpha_t \geq \max[\sqrt{2}\kappa_t, 4\|\nabla L_S(\w_t)\|_{\infty}]$, we have
\begin{equation*}\label{eq:sign_SGD_mini_batch_convergence_1}
    \E \|\nabla L_S(\w_R)\|_2^2 \leq O\left(\frac{1}{\sqrt{T}}\right) + O\left(\frac{\|\vec{K}\|_1}{\sqrt{T}}\right)~,
\end{equation*}
where $\w_R$ is uniformly randomly sampled from $\{\w_t\}_{t=1}^T$.
\end{restatable}

\begin{figure*}[t] 
\centering
\subfigure[MNIST, $\alpha_t^2 \approx 0.1$]{
 \includegraphics[width=0.23\textwidth]{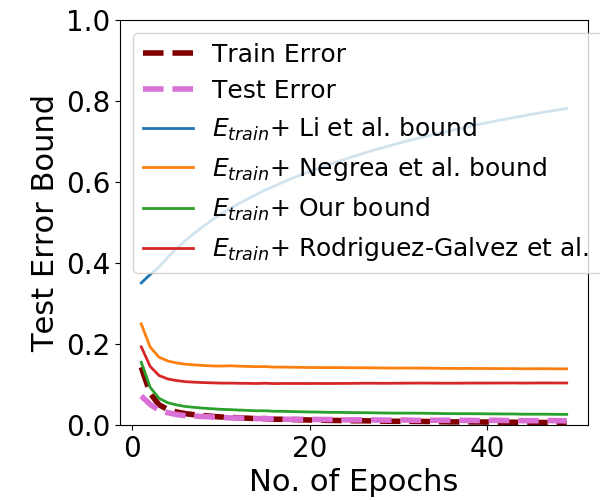}
 } 
 \subfigure[CIFAR-10, $\alpha_t^2 \approx 0.1$]{
 \includegraphics[width=0.23\textwidth]{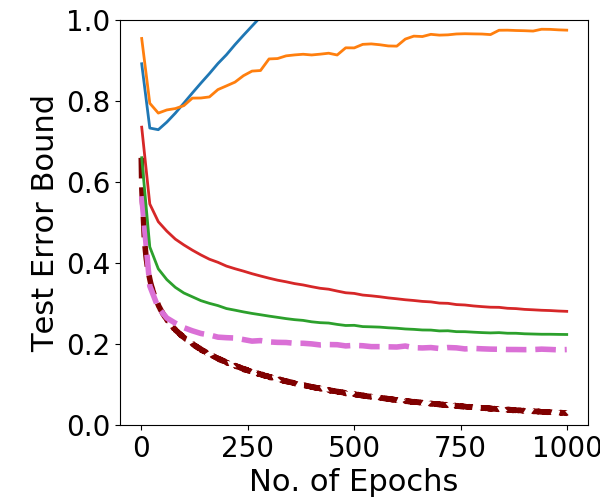}
 }
\subfigure[Fashion, $\alpha_t^2 \approx 0.1$]{
 \includegraphics[width=0.23\textwidth]{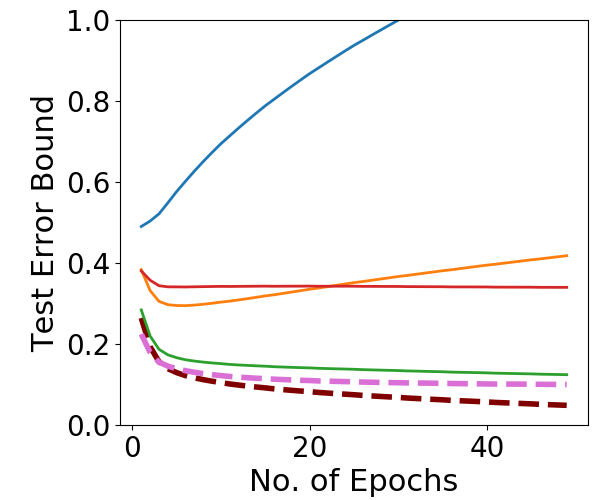}
 } 
 \subfigure[Fashion, $\alpha_t^2 \approx 0.01$]{
 \includegraphics[width=0.23\textwidth]{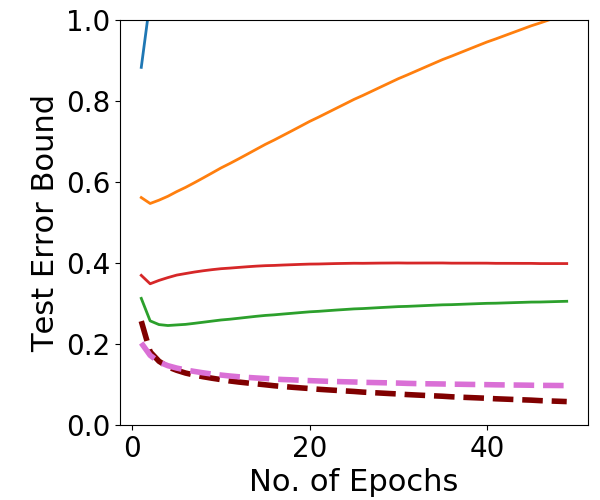}
 } 
 \subfigure[MNIST, $\alpha_t^2 \approx 0.1$]{
 \includegraphics[width=0.23\textwidth]{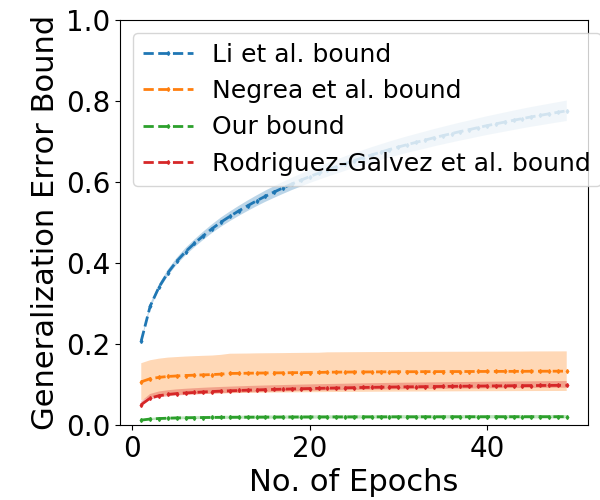}
 } 
 \subfigure[CIFAR-10, $\alpha_t^2 \approx 0.1$]{
 \includegraphics[width=0.23\textwidth]{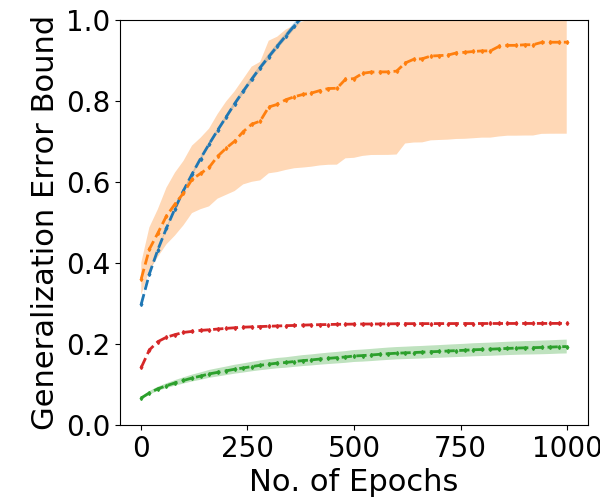}
 }
\subfigure[Fashion, $\alpha_t^2 \approx 0.1$]{
 \includegraphics[width=0.23\textwidth]{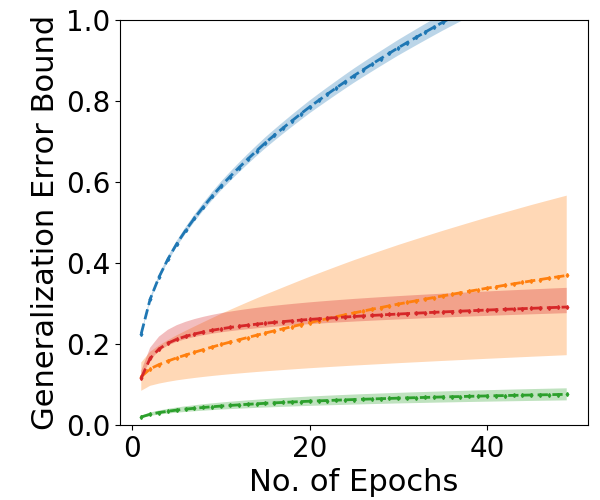}
 } 
 \subfigure[Fashion, $\alpha_t^2 \approx 0.01$]{
 \includegraphics[width=0.23\textwidth]{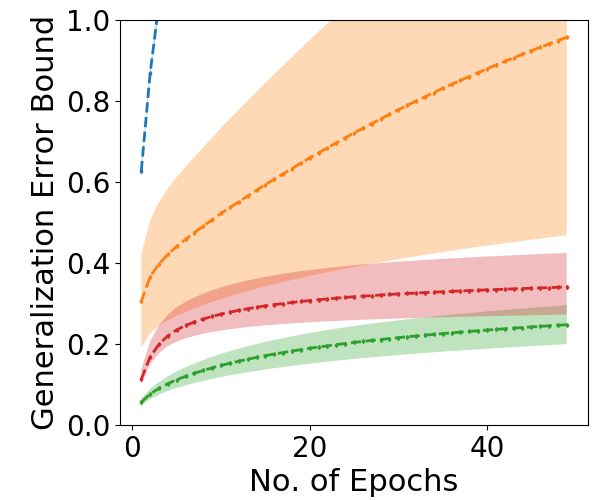}
}
\caption[]{Results for training CNN using SGLD on MNIST, Fashion-MNIST and CIFAR-10. 
(a)-(d) shows our bound is non-vacuous and can be used to bound the empirical test error. (e)-(h) compare our bound with the existing bounds and show the effect on $\alpha_t^2$. Our bounds are numerically sharper than existing bounds, and larger $\alpha_t^2$ leads to tighter generalization bounds which is consistent with Theorem~\ref{theo:efldstab}. 
}
\label{fig:sgld_bound}
\end{figure*}

\noindent {\bf Stochastic Gradient Langevin Dynamics (SGLD).} We acknowledge that following optimization result for SGLD exists in various forms, as noisy gradient descent algorithms with Gaussian noise have been studied in literature such as differential privacy, where SGLD can be viewed as DP-SGD  \citep{bassily2014private, wang2019differentially} and the proof technique boils down to bounding the stochastic variance of the noisy gradient \citep{shamir2013stochastic}.
\begin{restatable}{theo}{theoSGLDopt}
The following holds for any $S$, any initialization $\w_0$, and the expectation is taken over the randomness of algorithm: under Assumptions \ref{asmp: smooth} and \ref{asmp:batch} with $K_i = K, \forall i \in [p]$,  SGLD, i.e., EFLD with Gaussian noise, $\rho_t = \eta_t$, $\alpha_t = \sigma_t/\eta_t$,
and step size $\eta_t = \frac{1}{\sqrt{T}}$ satisfies 
\begin{align*}
\mathbb{E} \| \nabla L_S(\w_R)\|^2
\leq & O\left(\frac{1 }{\sqrt{T}}\right) 
 + O\left(\frac{  \frac{K}{2T} \sum_{t=1}^T (p \alpha_t^2 +   \kappa_t^2) }{\sqrt{T}} \right) ~,
\end{align*}
where $\w_R$ is uniformly randomly sampled from $\{\w_t\}_{t=1}^T$.
\end{restatable}
The error rate of SGLD depends on the noise level $\alpha_t$ and the sub-Gaussian parameter $\kappa_t$.  The bound has a $O(1/\sqrt{T})$ rate as long as the average noise level and sub-Gaussian parameter are bounded by a constant. 
Similar to differentially private SGD, the convergence rate depends on the dimension of the gradient $p$ due to the isotropic Gaussian noise. 
Special noise structures such as anisotropic noise that align with the gradient structure can improve the dependence on dimension \citep{kairouz2020dimension,zhang2021wide,asi2021private, zhou2020bypassing}.


%% file: arXiv2022/sec/stability.tex

From Lemma~\ref{theo:stab}, conditioned on a trajectory $W_{0:(t-1)}=\w_{0:(t-1)}$, mini-batches $S_{B_t}, S_{A_t}$, 
we can get an expected stability based generalization bound by suitably bounding the expected LSD $\E[\Lambda(B_t,A_t)]$ as in \eqref{eq:lsd}.
For EFLD, 
we have the following bound on the per step LSD $\Lambda(B_t,A_t)$.

\begin{restatable}{theo}{theoefld}
\label{theo:efld}
For a given set $\bar{S}_{n+1} \sim D^{n+1}$ and $\w_{t-1}$ at iteration $(t-1)$, let 
\begin{align*}
\Delta_{t|\w_{t-1}}(\bar{S}_{n+1}) = \max_{z, z' \in  \bar{S}_{n+1}} \|\nabla \ell(\w_{t-1}, z) - \nabla \ell(\w_{t-1}, z')\|_{2}~.
\end{align*}
Further, for a $c_2$-smooth log-partition function $\psi$, let the scaling $\alpha_{t|\w_{t-1}}$ be data-dependent such that $\alpha_{t|\w_{t-1}}^2 \geq  8 c_2 \Delta_{t|\w_{t-1}}^{2}(\bar{S}_{n+1})$. Then, for $\Lambda(A_t, B_t)$ as in \eqref{eq:lsd} we have
\begin{equation}
\begin{split}
\Lambda  (A_t, B_t) & \leq 5 c_2 \| \btheta_{B_t,\alpha_t} - \btheta_{B'_t,\alpha_t} \|_2^2  \\
& = \frac{5c_2}{2\alpha_{t|\w_{t-1}}^2} \left[\left\|\nabla \ell\left(\w_{t-1}, {S_{B_t}}\right)  -  \nabla \ell\left(\w_{t-1}, S'_{B_t}\right)\right\|_{2}^{2}\right]~.
\end{split}
\end{equation}
\end{restatable}
\begin{remark}
Theorem~\ref{theo:efld} shows that per step LSD $\Lambda(B_t,A_t)$ can be bounded by the (scaled) mini-batch gradient discrepancy. The result holds for {\em all} EFLD algorithms of the form \eqref{eq:efld1}-\eqref{eq:efld2}. \qed 
\end{remark}
\begin{remark}
Since $S_{B_t}$ and  $S'_{B_t}$ only differ at samples $z_n$ and $z'_n=z_{n+1}$, $\nabla \ell(\w_{t-1}, S_{B_t})  -  \nabla \ell(\w_{t-1}, S'_{B_t} ) = \frac{1}{b} (\nabla \ell (\w_{t-1}, z_n) - \nabla \ell(\w_{t-1},z'_n))$. The $1/b$ scale factor neutralizes the leading $b$ term in Lemma~\ref{theo:stab}. \qed 
 \end{remark}

Theorem~\ref{theo:efld} can now be directly applied to Lemma~\ref{theo:stab} to get expected stability based generalization bounds for EFLD.
\begin{restatable}{theo}{theoefldstab}
\label{theo:efldstab}
In the setting of Proposition~\ref{prop:hell} consider an exponential family Langevin dynamics (EFLD) algorithm of the form~\eqref{eq:efld1}-\eqref{eq:efld2} with a $c_2$-smooth log-partition function $\psi$.  Then, for mini-batch size $b \leq n/2$, with $c = c_0 \sqrt{5 c_2}$ (with $c_0$ as in Proposition~\ref{prop:hell}) and $\alpha_{t|}^2 \geq 8 c_2 \Delta_{t|}^2(\bar{S}_{n+1})$ (as in Theorem~\ref{theo:efld}, with the conditioning on $\w_{t-1}$ hidden to avoid clutter), we have
\begin{equation}
\begin{aligned}
| \E_{S}[L_D(A(S)) - L_S(A(S))] | \leq  \frac{c}{n} ~\underset{\bar{S}_{n+1}}{\E} \sqrt{ \sum_{t=1}^T  \underset{W_{0:(t-1)}}{\E}  \frac{\left\|\nabla \ell\left(W_{t-1}, {z_n} \right)  -  \nabla \ell\left(W_{t-1}, {z'_n}\right)\right\|_{2}^{2}}{\alpha_{t|W_{t-1}}^2}   }.
\end{aligned}
\end{equation}
\end{restatable}

\begin{remark}
The key term in the bound is the expected gradient discrepancy only on the sample $z_n, z'_n$ where $S_n,S'_n$ differ. Further, the only dependence on the specific exponential family is through the smoothness constant $c_2$. \qed 
\end{remark}

\begin{remark}
Since SGLD is a special case of EFLD, Theorem \ref{theo:efldstab} gives a generalization bound for SGLD.
The bound has effectively the same dependence on $n$ and $T$ as the bound in \citet{Li2020On}. 
However, the bound is quantitatively much sharper since the \emph{gradient norm} term $\frac{1}{n}\sum_{z \in S}\| \nabla \ell(\w_t,z)\|^2$ in \citet{Li2020On} gets replaced by the {\em gradient discrepancy} term $\|\nabla \ell(\w_t,z) - \nabla \ell(\w_t, z^\prime) \|^2$. As illustrated in our experiments (Section~\ref{sec:exp}), the gradient discrepancy is orders of magnitude smaller than the gradient norm.
The bound in \citet{negrea2019information} depends on a related \emph{gradient incoherence} which we found to be empirically smaller than gradient discrepancy in our experiments (Section~\ref{sec:exp}). However, their bound has a ${1}/{\sqrt{n}}$ sample dependence, which is worse than the ${1}/{n}$ dependence in our bound. \citet{wang2021analyzing} also obtained a $1/n$ rate in their bound depending on the sum of gradient variances. However, their bound scales inversely with $b$, since gradient variance increases as $b$ decreases. In contrast, our bound is suitable for small batch size as well. \citet{Lei2020} considered ``on average stability'' based generalization bounds, but has a dependence either on the global Lipschitz constant $L$ or on some form of convexity. Our bound does not depend on $L$, and works for any non-convex and non-smooth loss.  \qed
\end{remark}

\begin{remark}
Our bounds hold for non-convex and/or non-smooth loss functions. \citet{hardt2016train} developed uniform stability based generalization bounds for SGD for smooth losses. To compare with \citet{hardt2016train} for the non-convex case, note that by construction, 
$\gamma_t^2 := \frac{\|\nabla \ell (w,z_n) - \nabla \ell (w,z_n')\|_2^2}{\alpha^2} \leq \frac{\Delta^2(\bar{S}_{n+1})}{\alpha^2} \leq \frac{1}{8c_2}$,
so that our bound in Theorem~\ref{theo:efldstab} can be upper bounded by
$\frac{c}{n} \underset{\bar{S}_{n+1}}{\E} \sqrt{ \sum_{t=1}^T \underset{W_{0:(t-1)}}{\E}\left[\frac{1}{8c_2} \right]} \leq \frac{\sqrt{5}c_0}{2\sqrt{2}} \frac{\sqrt{T}}{n}$, 
since $c=c_0 \sqrt{5c_2}$. For a $\beta=1$ smooth loss, with step size $\eta_t = 1/t$, \citet{hardt2016train} gets a $O(\sqrt{T}/n)$ bound by their Theorem 3.12. However, {\em we do not need the loss to be smooth} and {\em we work with constant step sizes}. Their results do not extend to non-smooth losses or constant step sizes. \qed 
\end{remark}

\begin{remark}
EFLD can be extended to work with anisotropic noise by using $\btheta_{B_t,\aalpha_t} = \nabla \ell(\w_{t-1},S_{B_t}) \oslash \aalpha_t$ in \eqref{eq:efld2} where $\aalpha_t \in \R^p$ determines different scaling for each dimension and $\oslash$ denotes Hadamard division. Theorems~\ref{theo:efld} and \ref{theo:efldstab} can be extended to such anisotropic noise by using $\aalpha$-scaled norms for the gradient discrepancy, i.e., $\| \g - \g'\|_{2,\aalpha}^2 = \sum_j (g_j - g'_j)^2/\alpha_j^2$. \qed 
\end{remark}

\begin{remark}
The lower bound on $\alpha_t$ in Theorem~\ref{theo:efldstab} is a data-dependent quantity $\Delta_{t|}(\bar{S}_{n+1})$. 
For SGLD in \eqref{eq:sgld}, since $\alpha_{t|} = \sigma_t/\eta_t$ (see Example~\ref{exam:sgld}), the condition $\alpha^2_{t|} \geq c_3^2 \Delta_{t|}^2(\bar{S}_{n+1})$ for some constant $c_3$ implies $\eta_t \leq \sigma_t/c_3 \Delta_{t}(\bar{S}_{n+1})$, a much more benign (and computable) condition on the step size compared to those in the related work~\cite{mou2018generalization,Li2020On,hardt2016train} which require step size to be bounded by ${\sigma_{t}}/{L}$, where $L$ is the global Lipschitz constant for the loss $\ell$. Note that $\Delta_{t|}(\bar{S}_{n+1}) \ll L$ because $L$ is a uniform bound. 
Further, $\Delta_{t|}(\bar{S}_{n+1})$ is expected to decrease over iterations, i.e., as $t$ increases, and gradients get smaller. \qed
\vspace*{-2mm} 
 \end{remark}

\subsection{Proof Sketches of Main Results: Theorems~\ref{theo:efld} and \ref{theo:efldstab}}
\vspace*{-3mm} 
We focus on Theorem~\ref{theo:efld}. To avoid clutter, we drop the subscript $t$ for the analysis and note that the analysis holds for any step $t$.  When the densities $dP_{B,\bxi} = p_{\psi}(\bxi;\btheta_{B,\alpha})$ and $dP'_{B,\bxi} = p_{\psi}(\bxi;\btheta'_{B,\alpha})$, i.e., densities in the same exponential family but with different parameters 
$\btheta_{B,\alpha}$ and $\btheta'_{B,\alpha}$ because of the difference in the mini-batches,
by mean-value theorem, for each $\bxi$, we have 
\vspace{-2mm}
\begin{equation*}
p_{\psi}(\bxi; \btheta_{B,\alpha}) - p_{\psi}(\bxi; \btheta'_{B,\alpha}) 
= \langle \btheta_{B,\alpha} -\btheta'_{B,\alpha}, \nabla_{\tilde{\btheta}_{B,\alpha}} p_{\psi}(\bxi;\tilde{\btheta}_{B,\alpha})\rangle~,
\vspace{-2mm}
\end{equation*}
for some $\tilde{\btheta}_{B,\alpha} = \gamma_{\bxi} \btheta_{B,\alpha} + (1-\gamma_{\bxi}) \btheta'_{B,\alpha}$ where $\gamma_{\bxi} \in [0,1]$ with the subscript $\bxi$ illustrating dependence on $\bxi$. Then,
\vspace{-2mm}
{\small
\begin{align}
\hspace*{-3mm}
\Lambda_{A,B} &=  \bigintsss_\bxi \frac{\big( p_{\psi}(\bxi; \btheta_{B,\alpha}) - p_{\psi}(\bxi; \btheta'_{B,\alpha}) \big)^2}{p_{\psi}(\bxi; \btheta_{A,\alpha})} d\bxi \nonumber \\
 =& \bigintsss_\bxi \frac{\langle \btheta_{B,\alpha} -\btheta'_{B,\alpha}, \nabla_{\tilde{\btheta}_{B,\alpha}} p_{\psi}(\bxi;\tilde{\btheta}_{B,\alpha})\rangle ^2}{p_{\psi}(\bxi; \btheta_{A,\alpha})} d\bxi \nonumber \\
= &\bigintsss_\bxi \frac{\langle \btheta_{B,\alpha} -\btheta'_{B,\alpha}, \bxi - \nabla_{\tilde{\btheta}_{B,\alpha}} \psi(\bxi;\tilde{\btheta}_{B,\alpha}) \rangle ^2~p_{\psi}^2(\bxi;\tilde{\btheta}_{B,\alpha})}{p_{\psi}(\bxi; \btheta_{A,\alpha})} d\bxi,
\label{eq:expcore1}
\vspace{-3mm}
\end{align}}
where since $p_{\psi}(\bxi;\tilde{\btheta}_{B,\alpha}) = \exp(\langle \bxi, \tilde{\btheta}_{B,\alpha} \rangle - \psi(\tilde{\btheta}_{B,\alpha})) \pi_0(\bxi)$ we have 
$$ \nabla_{\tilde{\btheta}_{B,\alpha}} p_{\psi}(\bxi;\tilde{\btheta}_{B,\alpha}) =\allowbreak (\bxi - \nabla_{\tilde{\btheta}_{B,\alpha}} \psi(\bxi;\tilde{\btheta}_{B,\alpha}) ) p_{\psi}(\bxi;\tilde{\btheta}_{B,\alpha}).$$

\noindent {\bf Handling Distributional Dependence of $\tilde{\btheta}_{B,\alpha}$.}
Note that it is difficult to proceed with the analysis with the density term depending on parameter $\tilde{\btheta}_{B,\alpha}$ since $\tilde{\btheta}_{B,\alpha}$ depends on $\bxi$ and there is an outside integral over $\bxi$ in \eqref{eq:expcore1}. So, we first bound the density term depending on $\tilde{\btheta}_{B,\alpha}$ in terms of exponential family densities with parameters $\btheta_{B,\alpha}$ and $\btheta'_{B,\alpha}$ essentially using $c_2$-smoothness of $\psi$.
\begin{restatable}{lemm}{exptilde}
With $\tilde{\btheta}_{B,\alpha} = \gamma_{\bxi} \btheta_{B,\alpha} + (1-\gamma_{\bxi}) \btheta'_{B,\alpha}$ for some $\gamma_{\bxi} \in [0,1]$, we have 
\small
\begin{align*}
\frac{\exp\left[\langle \bxi, \tilde{\btheta}_{B, \alpha} \rangle - \psi(\tilde{\btheta}_{B,\alpha}) \right]}{ \max\left( \exp\big[\langle \bxi, \btheta_{B,\alpha} \rangle - \psi(\btheta_{B,\alpha}) \big] , \exp\left[\langle \bxi, \btheta'_{B, \alpha} \rangle - \psi(\btheta'_{B,\alpha}) \right]\right) }
\leq \exp\left[ c_2 \| \btheta_{B,\alpha} - \btheta'_{B,\alpha} \|_2^2 \right] ~.
\end{align*}
\normalsize
\label{lemm:exp_step2}
\end{restatable}
In other words, for any $\bxi$ we have
\begin{align*}
p_{\psi}(\bxi;\tilde{\btheta}_{B,\alpha}) \leq \exp\left[ c_2 \| \btheta_{B,\alpha} - \btheta'_{B,\alpha} \|_2^2 \right] \max \left( p_{\psi}(\bxi;\btheta_{B,\alpha}), p_{\psi}(\bxi;\btheta'_{B,\alpha}) \right)~.
\end{align*}
Since the parameters $\btheta_{B,\alpha}, \btheta'_{B,\alpha}$ in the right-hand-side depend on $\bxi$, the outside integral over $\bxi$ in \eqref{eq:expcore1} will not pose any unusual challenges.

\noindent {\bf Bounding the Density Ratio.} Next we focus on the density ratio $p^2_{\psi}(\bxi,\tilde{\btheta}_{B,\alpha})/p_{\psi}(\bxi;\btheta_{A,\alpha})$ in \eqref{eq:expcore1}. By Lemma~\ref{lemm:exp_step2}, it suffices to focus on $p^2_{\psi}(\bxi,\btheta_{B,\alpha})/p_{\psi}(\bxi;\btheta_{A,\alpha})$ or the equivalent term for $\btheta'_{B,\alpha}$. We show that the density ratio can be bounded by another distribution in the same exponential family $p_{\psi}$ with parameters $(2\btheta_{B,\alpha} - \btheta_{A,\alpha})$.
\begin{restatable}{lemm}{expdr}
\vspace*{-1mm}
For any $\bxi$, we have 
\begin{align*}
\frac{\exp\left[\langle \bxi, 2\btheta_{B, \alpha} \rangle - 2\psi(\btheta_{B,\alpha}) \right] }{\exp\left[\langle \bxi, \btheta_{A,\alpha} \rangle - \psi(\btheta_{A,\alpha}) \right]} \leq \exp \left[ 2c_2 \| \btheta_{B,\alpha} - \btheta_{A,\alpha} \|_2^2 \right] \exp \left[  \langle \bxi, (2\btheta_{B,\alpha} - \btheta_{A,\alpha} \rangle - \psi( 2\btheta_{B,\alpha} - \btheta_{A,\alpha} )\right].
\end{align*}
\label{lemm:exp_step3}
\end{restatable}
In other words, for any $\bxi$ we have
{\small
\begin{align*}
\frac{p^2_{\psi}(\bxi,\btheta_{B,\alpha})}{p_{\psi}(\bxi;\btheta_{A,\alpha})} \leq \exp \left[ 2c_2 \| \btheta_{B,\alpha} - \btheta_{A,\alpha} \|_2^2 \right] p_{\psi}(\bxi; 2\btheta_{B,\alpha} - \btheta_{A,\alpha})~.
\end{align*}}
The analysis for the term $p^2_{\psi}(\bxi,\btheta'_{B,\alpha})/p_{\psi}(\bxi;\btheta_{A,\alpha})$ is exactly the same. 

\noindent {\bf Bounding the Integral.} Ignoring multiplicative terms which do not depend on $\bxi$ for the moment, the analysis needs to bound an integral term of the form
\vspace{-2mm}
\begin{equation*}
\bigintsss_\bxi \langle \btheta_{B,\alpha} -\btheta'_{B,\alpha}, \bxi - \nabla \psi(\bxi;\tilde{\btheta}_{B,\alpha}) \rangle ^2 ~p_{\psi}(\bxi;2 \btheta_{B,\alpha} - \btheta_{A,\alpha}) d\bxi~,
\vspace{-2mm}
\end{equation*}
and a similar term with $p_{\psi}(\bxi;2 \btheta'_{B,\alpha} - \btheta_{A,\alpha})$. First, note that  $\nabla \psi(\bxi;\tilde{\btheta}_{B,\alpha}) = \tilde{\m}_{B,\alpha}$, the expectation parameter for $p_{\psi}(\bxi;\tilde{\btheta}_{B,\alpha})$~\citep{wainwright2008graphical,banerjee2005clustering}. The integral, however, is with respect to $p_{\psi}(\bxi;2\btheta_{B,\alpha}-\btheta_{A,\alpha})$, not $p_{\psi}(\bxi;\tilde{\btheta}_{B,\alpha})$. We handle this discrepancy by using
\begin{align*}
& \langle \btheta_{B,\alpha} -\btheta'_{B,\alpha}, \bxi - \nabla \psi(\bxi;\tilde{\btheta}_{B,\alpha}) \rangle^2 \\
& = \langle \btheta_{B,\alpha} -\btheta'_{B,\alpha}, (\bxi - \E[\bxi]) + (\E[\bxi] - \nabla \psi (\bxi; \tilde{\btheta}_{B,\alpha})) \rangle^2 \\
& \leq 2 \langle \btheta_{B,\alpha} -\btheta'_{B,\alpha}, \bxi - \E[\bxi] \rangle^2+2 \langle \btheta_{B,\alpha} -\btheta'_{B,\alpha}, \E[\bxi] - \nabla \psi (\bxi; \tilde{\btheta}_{B,\alpha}) \rangle^2~,
\end{align*}
where the expectation $\E[\bxi]$ is with respect to $p_{\psi}(\bxi;2\btheta_{B,\alpha}-\btheta_{A,\alpha})$.
Quadratic form for the first term yields the covariance $\E[(\bxi- \E[\bxi])(\bxi - \E[\bxi])^T] = \nabla^2 \psi(\btheta_{2\btheta_{B,\alpha} - \btheta_{A,\alpha}}) \leq c_2 \I$, by smoothness and since the covariance matrix of an exponential family is the Hessian of the log-partition function~\citep{wainwright2008graphical}. Since $\E[\bxi] = \nabla \psi ( 2\btheta_{B,\alpha} - \btheta_{A,\alpha})$, the second term depends on the difference of gradients $\nabla \psi ( 2\btheta_{B,\alpha} - \btheta_{A,\alpha}) - \nabla \psi(\tilde{\btheta}_{B,\alpha})$ which, using smoothness and additional analysis, can be bounded by the norm of $(\btheta_{B,\alpha} - \btheta_{A,\alpha})$. All the pieces can be put together to get the bound in Theorem~\ref{theo:efld}, which when used in Lemma~\ref{theo:stab}
yields Theorem~\ref{theo:efldstab}.

%% file: arXiv2022/sec/exp.tex
In this section, we conduct experiments to evaluate our generalization error bounds. For SGLD, we compare our bound in Theorem \ref{theo:efldstab} with existing bounds in \citet{Li2020On, negrea2019information, rodriguez2021random} for various datasets. Note that the bound presented in \citet{rodriguez2021random} is an extension of that in \citet{haghifam2020sharpened} from full-batch setting to mini-batch setting. We also evaluate the optimization performance of proposed Noisy Sign-SGD, comparing it with the original Sign-SGD \citep{bernstein2018signsgd} and present the corresponding generalization bound in Theorem \ref{theo:efldstab}. 

The details of model architectures, learning rate schedules, hyper-parameter selections, and additional experimental results can be found in Appendix \ref{app:exp}. Evaluation of the expectation in Theorem~\ref{theo:efldstab} is done based on (re)sampling (Appendix~\ref{app:exp}). We emphasize that the goal for the experiments is to do a comparative study relative to existing bounds. We note that the empirical performance of the methods can potentially be improved with better architectures and training strategies, e.g., deeper/wider networks, data augmentation, batch/layer normalization, etc. 

\begin{figure}[t!]
    \centering
      \subfigure[MNIST, $\alpha_t^2 \approx 0.1$]{
      \includegraphics[width=0.4\textwidth]{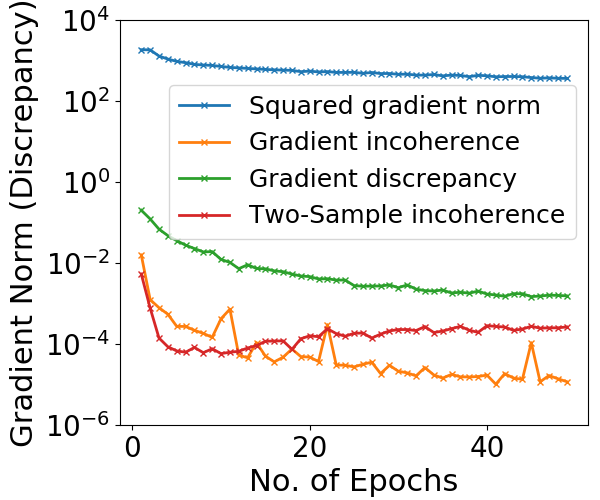}} 
      \hspace*{10mm}
      \subfigure[CIFAR-10, $\alpha_t^2 \approx 0.1$]{\includegraphics[width=0.4\textwidth]{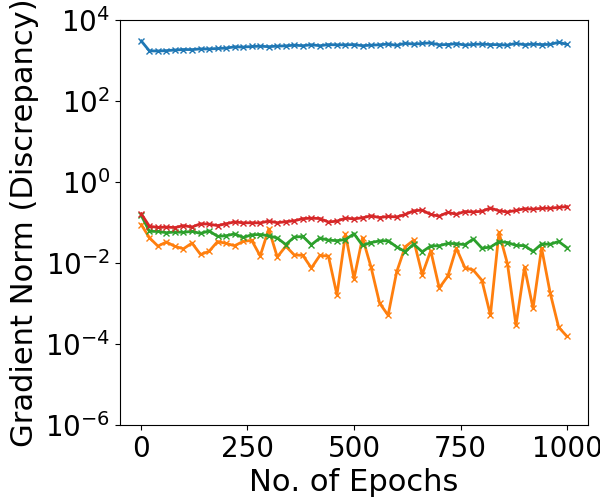}}
    \caption[]{Comparison of the squared gradient norm \cite{Li2020On}, the gradient incoherence \cite{negrea2019information}, the two-sample incoherence \cite{rodriguez2021random}, and the gradient discrepancy in our bound. Incoherence or discrepancy based quantities are orders of magnitude smaller than the gradient norm. }
    \label{fig:sgld_key_quantities}
\end{figure}

\subsection{Stochastic Gradient Langevin Dynamics}
\noindent \textbf{Comparison with existing work.} We have derived generalization error bounds that depend on the data-dependent \emph{gradient discrepancy}, i.e., $\left\|\nabla \ell\left(\w_{t}, z_n \right)  -  \nabla \ell\left(\w_{t}, z'_n\right)\right\|_{2}^{2}$. Existing bounds in \citet{Li2020On} and \citet{negrea2019information} have also improved the Lipschitz constant in \citet{mou2018generalization} to a data-dependent quantity. All these generalization error bounds can be added to the empirical training error to get bounds on the empirical test error. As shown in Figure~\ref{fig:sgld_bound} (a)-(d), our bound is able to generate a much tighter upper bound on the test error. The improvements is mainly due to the fact that we replace the squared \emph{gradient norm} in \citet{Li2020On}, the squared norm of \emph{gradient incoherence}
in  \citet{negrea2019information}, and that of \emph{two-sample incoherence} in \citet{rodriguez2021random} with the gradient discrepancy while maintaining a $1/n$ sample dependence. Figure~\ref{fig:sgld_bound} (e)-(h) shows that our bounds are much sharper than those of \citet{Li2020On} because our gradient discrepancy norms (Figure~\ref{fig:sgld_key_quantities}) are usually 2-4 order of magnitude smaller than the squared gradient norms in \citet{Li2020On}. Our bounds are also sharper than those of \citet{negrea2019information} and \citet{rodriguez2021random} due to our $O(1/n)$ sample dependence compared to their $O(1/\sqrt{n})$ dependence. Although the gradient incoherence in \citet{negrea2019information} is can be about 1 to 2 order of magnitude smaller than the gradient discrepancy for simple problems such as MNIST (Figure~\ref{fig:sgld_key_quantities}(a)), the difference between the gradient incoherence and our gradient discrepancy reduces as the problem becomes harder (see results for CIFAR-10 in Figure~\ref{fig:sgld_key_quantities}(b)).

\begin{figure}[t] 
\centering
\subfigure[Training Error]{
\includegraphics[width=0.23\textwidth]{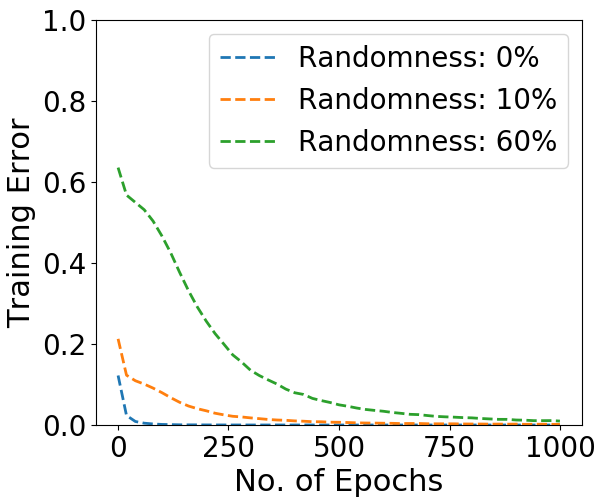}
 }
\subfigure[Gradient Discrepancy]{
\includegraphics[width=0.23\textwidth]{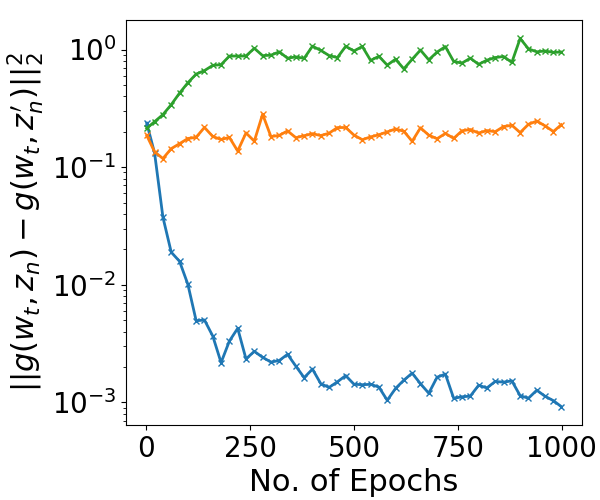}
 }\vspace*{-2mm}
\subfigure[Generalization Bound]{
\includegraphics[width=0.23\textwidth]{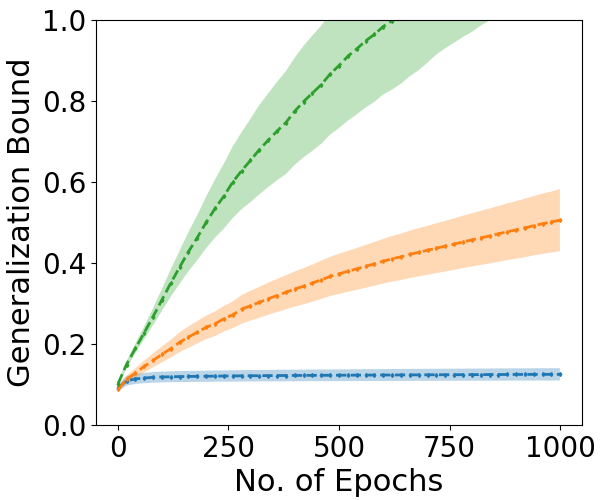}
 }
\subfigure[Test Error Bound]{
\includegraphics[width=0.23\textwidth]{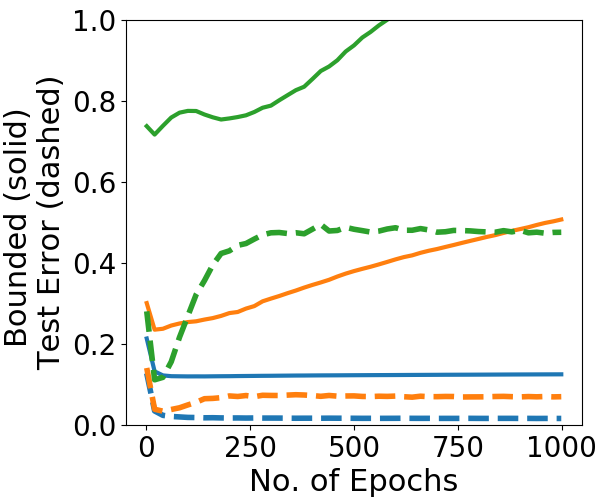}
 }  
\caption[]{Results for training CNN using SGLD on a subset of MNIST ($n=10000$) with different label randomness.(a) Training error: While training takes more epochs with random labels, training error does go to zero. (b) Gradient discrepancy: As the randomness increases, so does the gradient discrepancy, 
leading to the increase in the generalization bound. (c) Generalization bound: As the randomness increases, so does the generalization bound in Theorem \ref{theo:efldstab}. (d) Test error and bound: As the randomness increases, the empirical test error (dashed lines) increases and so does the test error bound (solid lines); overall, the bounds stay valid.
}
\label{fig:bound_random}
\end{figure}

\noindent \textbf{Effect of Random Labels.} Motivated by \citet{zhang2017understanding}, we train CNN with SGLD on a smaller subset of MNIST dataset ($n=10,000$) with randomly corrupted labels. The corruption fraction varies from $0\%$ (no label corruption) to $60\%$. As shown in Figure~\ref{fig:bound_random} (a), for long enough training time, all experiments with different levels of label randomness can achieve almost zero training error. However, increase in random labels leads to increase in gradient discrepancy (Figure~\ref{fig:bound_random}(b)) which in turn leads to increase in the generalization bound (Figure~\ref{fig:bound_random}(c)). 
%
%
As a result, as the empirical error rate increases with increase in label randomness (Figure \ref{fig:bound_random}(d) dashed lines), we get the correct increase in the test error bound  (solid lines).

\begin{figure}[t] 
\centering
\subfigure[CNN, MNIST]{
 \includegraphics[width=0.23\textwidth]{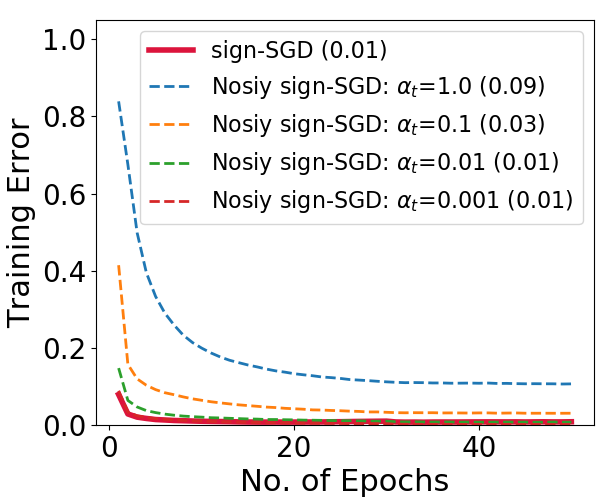}
 } 
  \subfigure[CNN, Fashion]{
 \includegraphics[width=0.23\textwidth]{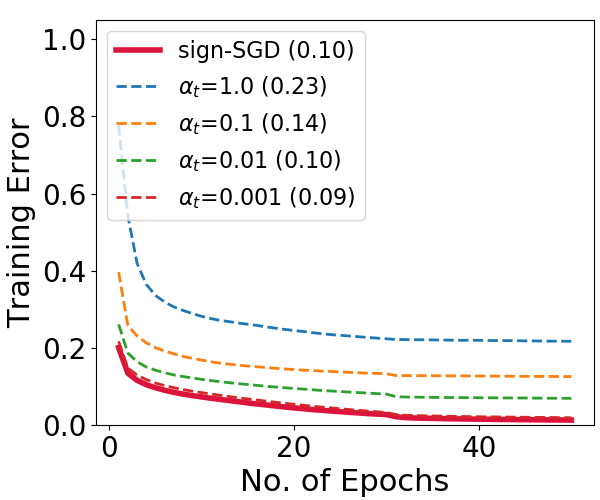}
 }\vspace{-2mm}
\subfigure[CNN, MNIST, $\alpha_t=1$]{
 \includegraphics[width=0.23\textwidth]{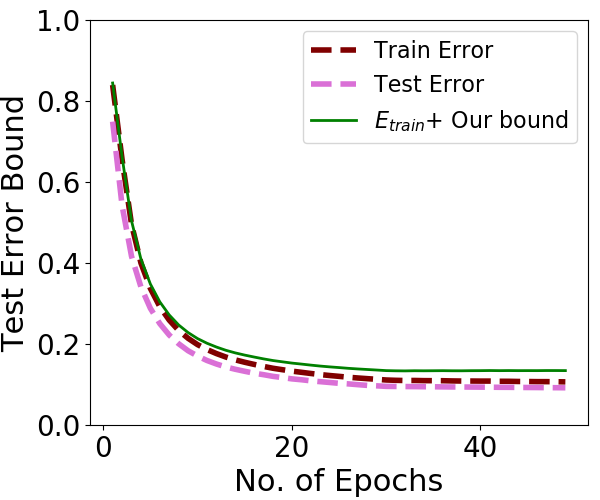}
 } 
 \subfigure[CNN, Fashion, $\alpha_t=1$]{
 \includegraphics[width=0.23\textwidth]{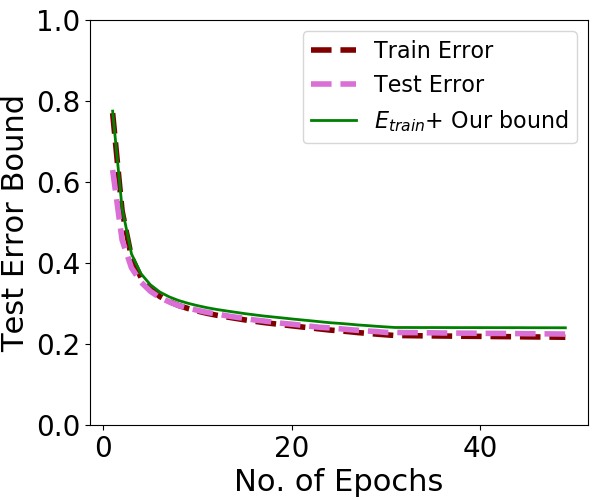}
 }
\caption[]{(a)-(b) show the training dynamics of CNN on MNIST and Fashion-MNIST using noisy sign-SGD with different scaling $\alpha_t$. Legends indicate the choice of $\alpha_t$ and the numbers in brackets are test errors at convergence. As $\alpha_t\to 0$, Nosiy sign-SGD matches both the optimization trajectory as well as the final test accuracy of the original sign-SGD \citep{bernstein2018signsgd}. (c)-(d) show that empirical test error can be bounded by our bound and the corresponding training error.}
\label{fig:noisy_signsgd}
\end{figure}

\subsection{Noisy Sign-SGD}
In this section, we present numerical results for Noisy Sign-SGD proposed in section~\ref{subsec:efld}. 
Since none of the existing bounds can give a valid generalization bound for Noisy Sign-SGD, we only present our bound here.

\noindent \textbf{Optimization.} Figure~\ref{fig:noisy_signsgd} (a)-(b) show the training dynamics of Noisy Sign-SGD under various choices of $\alpha_t$. For small $\alpha_t$, Noisy Sign-SGD matches both the optimization trajectory as well as the test accuracy of the original Sign-SGD \citep{bernstein2018signsgd}. However, as $\alpha_t$ increases, the distribution over $\{-1,+1\}$ is more spread out and the corresponding Noisy Sign-SGD seems to converge but to a sub-optimal value. 

\noindent \textbf{Generalization Bound.} Figure \ref{fig:noisy_signsgd}(c)-(d) show that our bound successfully bounds the empirical test error.
Larger $\alpha_t$ leads to sharper generalization bounds. However, larger $\alpha_t$ adversely affects the optimization, e.g., Figure~\ref{fig:noisy_signsgd} (a)-(b) blue and orange lines. The results illustrate the trade-off between the empirical optimization and the generalization bound. In practice, one needs to balance the optimization error and generalization by choosing a suitable scaling $\alpha_t$.

%% file: arXiv2022/sec/conc.tex
\vspace*{-2mm}
Inspired by recent advances in stability based and information theoretic approaches to generalization bounds \citep{mou2018generalization,pensia2018generalization,negrea2019information,Li2020On,haghifam2020sharpened}, we have presented a framework for developing such bounds based on expected stability for noisy stochastic iterative algorithms. We have also introduced Exponential Family Langevin Dynamics (EFLD), a large family of noisy gradient descent algorithms based on exponential family noise, including SGLD and Noisy Sign-SGD as two special cases.  We have developed an expected stability based generalization bound applicable to any EFLD algorithm with a $O({1}/{n})$ sample dependence and a dependence on gradient discrepancy, rather than gradient norms. Further, we have provided optimization guarantees for special cases of EFLD, viz.~Noisy Sign-SGD and SGLD. Our experiments on various benchmarks illustrate that our bounds are non-vacuous and quantitatively much sharper than existing bounds~\citep{Li2020On,negrea2019information}.


%% file: arXiv2022/sec/related.tex


\noindent \textbf{Uniform stability.} Uniform stability is a classical approach for bounding generalization error \citep{bousquet2002stability,hardt2016train,bousquet2020sharper, ShalevShwartz2009StochasticCO,feldman2018generalization,feldman2019high}, pioneered by \citet{rogers1978finite,devroye1979distribution}. Recently, uniform stability has been used in analyzing the  stability of stochastic gradient descent (SGD) \citep{hardt2016train}. 
\citet{mou2018generalization} prove the uniform stability of SGLD \citep{WellingT11, raginsky2017non} by showing that uniform stability can be bounded by the squared Hellinger distance, and further they establish discretized Fokker-Planck equations for analyzing the squared Hellinger distance. Then they provide uniform stability based generalization bounds for SGLD as $\frac{L}{n} \sqrt{\sum_t\eta^2_t/\sigma_t^2}$ which depends on $L$, the global Lipschitz constant for gradients, and the step size $\eta_t \leq \frac{\sigma_t \ln 2}{L}$ \citep{mou2018generalization,Li2020On}. 
Recently, \citet{Li2020On} followed up on \citet{mou2018generalization} and derived a data-dependent bound based on Bayes-stability, and got a bound of the form $\frac{c}{n} \sqrt{\sum_t \eta^2_t \g_e(t)/\sigma_t^2}$, where $\g_e(t)$ is the expected gradient norm square at step $t$. Their bound improves the Lipschitz constant $L$ to the expected gradient norm square. Recently, \citet{farghly2021timeindependent} provided a time-independent bound for SGLD of the form $O(1/n + \eta^{1/2}d^{1/2})$ which requires the step size scales as $O(1/(n^2d))$ to obtain an $O(1/n)$ bound.  
\citet{bassily2019private} analyze the uniform stability of  differentially private SGD (DP-SGD) for convex optimization by showing the gradient update is a non-expansive operation, which is the key fact in proving the stability of SGD \citep{hardt2016train}. 
The approach in \citet{hardt2016train} can extend to non-convex setting as well, however it requires fast decaying in step size as $\eta_t = O(1/t)$.  \citet{bassily2020stability} provide stability analysis of SGD for convex and non-smooth functions. 

\noindent \textbf{Information-theoretic bounds.} Besides the works mentioned above, other theories of deriving generalization bounds for noisy iterative algorithms have been proposed via information-theoretic approaches \citep{russo2016controlling, xu2017information}. Such results show that the generalization error of any learning algorithm can be bounded as $O(\sqrt{I(S; W)/n})$, where $I(S; W)$ is the mutual information between the algorithm input $S$ and the algorithm output $W$. Recent work following this approach focus on bounding the mutual information for a broad class of iterative algorithms, including SGLD  to obtain a $O(\sqrt{\log T/n})$ generalization bound by choosing $\eta_t = O(1/t)$, where $T$ is the total number of iterations~\citep{pensia2018generalization,bu2019tightening}. Subsequent improvements to this technique were made by 
\citet{negrea2019information, haghifam2020sharpened, rodriguez2021random} 
to prove data-dependent generalization bounds that do not depend
on the Lipschitz constant of the loss function and obtain $\sqrt{\sum_t \eta_{t}/n} $ bounds. 
Especially, \citet{haghifam2020sharpened, zhou2021individually} introduce generalization bounds based on conditional mutual information inspired by \citet{steinke2020reasoning}, leading to tighter bounds than those based on mutual information, which was extended by \citet{rodriguez2021random} from full-batch gradient to stochastic setting. Recently, \citet{wang2021analyzing} provided a bound for SGLD of the from $O(\frac{b}{n}\sum_{j=1}^m\sqrt{\sum_{t\in {\cT}_j}\eta_t^2\textit{Var}(\mathbf{g}_j(t))/\sigma_t^2})$, where $b$ is the mini-batch size, $m$ is the number of mini-batches,  ${\cT}_j$ contains the indices of iterations for mini-batch $S_j$ and $\textit{Var}(\mathbf{g}_j(t))$ is the variance of mini-batch gradient on $S_j$ evaluated on the training set. However, \citet{wang2021analyzing} requires splitting training samples into disjoint $m$ mini-batches before training and obtained a bound dependending on the sum of gradient variance. Their bound also scales inversely to the batch size since the number of mini-batches $m$ and gradient variance increase as the batch size decreases. 
\citet{neu2021informationtheoretic} extend this information-theoretic approach  to derive generalization bound for vanilla SGD. \citet{hellstrom2021fast} provide a fast-rate bound for bounded loss functions based via Conditional Information Measures \citep{grunwald2021pac, hellstrom2020generalization}, which also provides a unified view of some of the above results.


\noindent \textbf{Noisy iterative algorithms.} Introducing additional noise in the stochastic gradient has been popular in training deep nets. Noisy iterative methods have proven to be useful for machine learning applications, especially for deep neural networks in terms of escaping from saddle points \citep{jin2017escape,jin2019nonconvex}, preserving privacy  \citep{bassily2020stability,wang2019differentially}, boosting generalization and stability \citep{mou2018generalization,Li2020On}. SGLD \citep{WellingT11} has been one of the most popular noisy iterative algorithms for non-convex learning problems, where an isotropic Gaussian noise is added to the stochastic gradient. There has been some work \citep{wang2015privacy, li2019connecting} connecting SGLD with differentially private SGD  algorithm (DP-SGD) \citep{bassily2020stability,wang2019differentially} which usually adds noise with constant variance to the stochastic gradient. 
Uniform stability has also been popular in the differential privacy literature for analyzing the generalization error bound of DP-SGD algorithms. Recently, noise has been proven to be useful in Sign-SGD \citep{bernstein2018signsgd, bernstein2018signsgdnoise, chen2019distributed} which has gained popularity as it reduces communication cost in distributed learning.  Existing versions of noisy sign-SGD first adds symmetric noise to the stochastic gradient, then take the sign of the noisy stochastic gradient to update the parameters \citep{chen2019distributed, jin2020stochastic}. \citet{bernstein2018signsgdnoise, chen2019distributed, jin2020stochastic} have shown that when noise is unimodal and symmetric, sign-SGD can guarantee convergence to stationary point.  Recently, \citet{wang2021optimizing} study the anisotropic noise for SGLD, where they optimize
the information-theoretical generalization bound by manipulating the noise structure in SGLD. They prove that with constraint to guarantee low empirical risk, the optimal noise covariance is the square root of the expected gradient covariance. In recent work, \citet{Lei2020} considered ``on average stability'' based generalization bounds, which is related to our work. Their bound has a dependence either on $L$ or on some form of convexity.

%% file: arXiv2022/app/exp_stability.tex
\prophell*
\proof Let $S_n=\left(z_{1}, \ldots, z_{n}\right)$ and $\tilde S^{\prime}_n=\left(z_{1}^{\prime}, \ldots, z_{n}^{\prime}\right)$ two independent random samples and let $S^{(i)}_n=\left(z_{1}, \ldots, z_{i-1}, z_{i}^{\prime}, z_{i+1}, \ldots, z_{n}\right)$ be the sample that is identical to $S$ except in the $i$-th example where we replace $z_{i}$ with $z_{i}^{\prime}$. Note that $S^{(n)}=S'_n$ where $S'_n$ is the dataset obtained by replacing $z_n \in S_n$ with $z'_n$ as in the Proposition statement. Now, by definition we have
\begin{align*}
& \left| \E_{S_n \sim D^n}[L_D(A(S_n)) - L_S(A(S_n))]\right| \\
& = \left| \E_{S_n \sim D^n}[L_S(A(S_n)) - L_D(A(S_n))]\right| \\
&= \left|\mathbb{E}_{S_n} \mathbb{E}_{\tilde S_n^\prime} \mathbb{E}_{A} \left[\frac{1}{n} \sum_{i=1}^{n}\left(\ell(A(S_n^{(i)}), z_{i}^{\prime})-\ell(A(S_n), z_{i}^{\prime}) \right)\right] \right| \\
& = \left| \mathbb{E}_{S_n} \mathbb{E}_{\tilde S_n^\prime}  \frac{1}{n} \sum_{i=1}^{n} \mathbb{E}_{A} \left[ \ell(A(S_n^{(i)}), z_{i}^{\prime})-\ell(A(S_n), z_{i}^{\prime}) \right]\right|\\
& \overset{(a)}{=} \left| \mathbb{E}_{S_n} \mathbb{E}_{\tilde S_n^\prime} \frac{1}{n} \sum_{i=1}^{n} \left(\int_{\mathbb{R}^{d}} \ell(\mathbf{w} ; z_i^\prime) p^{(i)}(\mathbf{w}) d \mathbf{w}-\int_{\mathbb{R}^{d}} \ell(\mathbf{w} ; z_i^\prime) p(\mathbf{w}) d \mathbf{w} \right) \right|\\
& = \left| \mathbb{E}_{S_n} \mathbb{E}_{\tilde S_n^\prime} \frac{1}{n} \sum_{i=1}^{n} \int_{\mathbb{R}^{d}} \ell(\mathbf{w} ; z_i^\prime) \left( p^{(i)}(\mathbf{w}) - p(\mathbf{w}) \right)d \mathbf{w} \right| \\
& = \left| \mathbb{E}_{S_n} \mathbb{E}_{\tilde S_n^\prime} \frac{1}{n} \sum_{i=1}^{n} \int_{\mathbb{R}^{d}} \ell(\mathbf{w} ; z_i^\prime)\left(\sqrt{p^{(i)}(\mathbf{w})}+\sqrt{p(\mathbf{w})}\right)\left(\sqrt{p^{(i)}(\mathbf{w})}-\sqrt{p(\mathbf{w})}\right)  d \mathbf{w} \right| \\
& \overset{(b)}{\leq} \left| \mathbb{E}_{S_n} \mathbb{E}_{\tilde S_n^\prime} \frac{1}{n} \sum_{i=1}^{n} \left\{\left(\int_{\mathbb{R}^{d}} \ell^2(\mathbf{w} ; z_i^\prime) \left(\sqrt{p^{(i)}(\mathbf{w})}+\sqrt{p(\mathbf{w})}\right)^{2} d \mathbf{w}\right)^{\frac{1}{2}}\left(\int_{\mathbb{R}^{d}}\left(\sqrt{p^{(i)}(\mathbf{w})}-\sqrt{p(\mathbf{w})}\right)^{2} d \mathbf{w}\right)^{\frac{1}{2}}\right\} \right|
\end{align*}
where in (a) $p^{(i)}(\w), p(\w)$ are respectively the distributions obtained by $A(S_n^{(i)}),A(S_n)$, and (b) 
follows by Cauchy-Schwartz inequality. Focusing on the first integral, we have:

\begin{align*}
\left(\int_{\mathbb{R}^{d}} \ell^2(\w ; z_i^\prime) \left(\sqrt{p^{(i)}(\w)}+\sqrt{p(\w)}\right)^{2} d \w \right)^{1/2} 
& \leq \left(2 \int_{\mathbb{R}^{d}} \ell^2(\w ; z_i^\prime) p^{(i)}(\w) d\w +  2 \int_{\mathbb{R}^{d}} \ell^2(\w ; z_i^\prime) p(\w) d \w \right)^{1/2} \\
& \leq \left( c_0^2/2 + c_0^2/2 \right)^{1/2} =c_0 ~.
\end{align*}

Hence, by definition of the Hellinger divergence, we have 
\begin{align*}
\E_{S_n \sim D^n}[L_D(A(S_n)) - L_S(A(S_n))]
& \leq   \mathbb{E}_{S_n} \mathbb{E}_{\tilde {S_n}^\prime} \frac{1}{n} \sum_{i=1}^{n}c_0 \sqrt{2H^{2}\left(A(S_n), A(S_n^{(i)})\right)}\\
& \overset{(a)}{=}   \mathbb{E}_{S_n} \mathbb{E}_{\tilde S_n^\prime} \frac{1}{n} \sum_{i=1}^{n}c_0 \sqrt{2H^{2}\left(A(S_n), A(S'_n)\right)}\\
& = c_0 \E_{S_n \sim D^n} \E_{z'_n \sim D} \sqrt{2 H^2\big(A(S_n), A(S'_{n})\big)}~,
\end{align*}
where (a) follows since the samples are drawn i.i.d.~and the randomized algorithm $A(\cdot)$ is permutation invariant. That completes the proof. \qed 




\begin{prop}
For any distributions $P$ and $P'$,
$2 H^2(P,P') \leq \min \big\{ KL(P,P'),\sqrt{\frac{1}{2}KL(P,P')} \big\}$. 
\label{prop:hellkl}
\end{prop}
\proof For the first part, note that:
\begin{align*}
KL(P ,P') & = \int \left(\log \frac{p(\w)}{p'(\w)} \right) p(\w) ~d\w \\
& =  2 \int  \left( - \log \sqrt{\frac{p'(\w)}{p(\w)}} \right) p(\w)~ d\w \\
& \overset{(a)}{\geq} 2 \int \left( 1 - \sqrt{\frac{p'(\w)}{p(\w)}} \right) p(\w)~ d\w \\ 
& = 2 \int \left( p(\w) - \sqrt{p(\w) p'(\w)}) \right) ~d\w \\
& = \int \left(p(\w) + p'(\w) - 2 \sqrt{p(\w)p'(\w)} \right) ~d\w \\
& = \int \left(\sqrt{p(\w)} - \sqrt{p'(\w)}\right)^2~ d\w\\
& = 2 H^2(P,P')~,
\end{align*}
where (a) follows since for $z > -1$, $\log (1+z) \leq z$, using $x = 1+z$ and changing signs gives $-\log x \geq 1-x$.

For the second part, note that:
\begin{align*}
2H^2(P,P') & =  \int_\w ( \sqrt{p(\w)} - \sqrt{p'(\w)} )^2 d\w \\
& =  \int_\w | \sqrt{p(\w)} - \sqrt{p'(\w)}| \times | \sqrt{p(\w)} - \sqrt{p'(\w)}| ~d\w \\
& \leq  \int_\w | \sqrt{p(\w)} - \sqrt{p'(\w)}| \times | \sqrt{p(\w)} + \sqrt{p'(\w)}|~ d\w \\
& =  \int_\w | p(\w) - p'(\w)|~ d\w \\
& =  TV(P,P')~,
\end{align*}
where $TV(P,P')$ denotes the total variation distance \citep{pollard2002user}. Further, from Pinsker's inequality \citep{tsybakov2008introduction}, we have
\begin{align*}
TV(P,P') \leq \sqrt{\frac{1}{2} KL(P,P')}~.
\end{align*}
Combining the two results completes the proof. \qed



\subsection{Proofs for Section \ref{sec: exp_stab_nsi}}

Our first result establishes a bound on the KL-divergence between two component mixture models in terms of the mixing weight of the unique components. Similar results have appeared in \citep{Li2020On, bun2018composable} in related contexts. Our proof is different, simple, and self-contained.
\begin{lemm}\label{lemm:mix_bound}
Let $Q,Q',R$ be any three distributions such that $Q,Q'$ are both absolutely continuous w.r.t.~$R$. Then, for any $s \in (0,1)$
\begin{equation}
KL\left(sQ+(1-s)R ~\big\| ~sQ' + (1-s)R \right) \leq 
\frac{s^2}{1-s}~\int_{\w} \frac{(Q(\w) - Q'(\w))^2}{R(\w)} d\w~ .
\end{equation}
\end{lemm}

\proof Let $U=sQ' + (1-s)R$. Then, with $F(x) = Q(x)-Q'(x)$, we have
\begin{align*}
     KL(sQ+(1-s)R \| sQ' + (1-s)R) & =  KL(U + s(Q - Q')  \| U ) \\
     & = \int (U(x) + sF(x)) \log \left( \frac{U(x) + s F(x)}{U(x)} \right) dx \\
     & = \int (U(x) + sF(x)) \log \left( 1 + \frac{s F(x)}{U(x)} \right) dx \\
     & = \int sF(x) + \frac{s^2 F^2(x)}{2U(x)} -  \frac{s^3 F^3(x)}{6U^2(x)} + \frac{s^4 F^4(x)}{12U^3(x)}  -\cdots dx \\
     & = s^2 \int \frac{F^2(x)}{U(x)}\left(  \frac{1}{2} -  \frac{s F(x)}{6U(x)} + \frac{s^2 F^2(x)}{12U^2(x)} - \cdots \right) dx~,
\end{align*}
where the first term vanishes since $\int F(x) dx = \int (Q(x) - Q'(x)) dx = 0$. This is the reason the dependency is on $s^2$, not $s$. With $W(x) = s F(x)/U(x)$, noting that $W(x) > -1$, and 
\begin{equation}
    \left( \frac{1}{2} - \frac{W(x)}{6} + \frac{W^2(x)}{12} - \cdots \right) = \frac{(1+W(x))\log(1+W(x)) - W(x)}{W^2(x)} \leq 1~,
\end{equation}
we have
\begin{align*}
   KL(sP+(1-s)R \| sQ + (1-s)R) 
     & = s^2 \int \frac{F^2(x)}{U(x)} \left( \frac{ (1+W(x)) \log (1+W(x)) - W(x)}{W^2(x)} \right) dx \\
     & \leq s^2 \int \frac{F^2(x)}{U(x)} dx~ \\
     & \leq \frac{s^2}{(1-s)} \int \frac{(Q(x) - Q'(x))^2}{R(x)} dx~.
\end{align*}
That completes the proof. \qed

\begin{restatable}{prop}{mixmodel}
Consider the mixture models 
$Q_{t|} = \frac{1}{|G_1|} \sum_{B_t \in G_1} P_{B_t,\bxi_t}$,  $Q'_{t|} = \frac{1}{|G_1|} \sum_{B_t \in G_1} P'_{B_t,\bxi_t}$, $R_t = \frac{1}{|G_0|} \sum_{A_t \in G_0} P_{A_t,\bxi_t}$.  
Then, with 
$s = \frac{|G_1|}{|G|} = \frac{\binom{n-1}{b-1}}{\binom{n}{b}} = \frac{b}{n}$,
we have 
\begin{align*}
    P_{t|} = s Q_{t|} + (1-s) R_{t|}~, \qquad \text{and} \qquad P'_{t|} = s Q'_{t|} + (1-s) R_{t|}~.
\end{align*}
\label{prop:mixmodel}
\vspace*{-6mm}
\end{restatable}
\proof The proof follows the argument in the proof of Lemma 21 in \cite{Li2020On}.\qed

\begin{lemm}
Consider a general noisy stochastic iterative algorithm with updates of the form~\eqref{eq:nmia} with mini-batch size $B_t = b$. Then, conditioned on any trajectory $\w_{0:t-1}$, we have
\begin{align}
KL\left(P_{t|} \| P_{t|} \right) \leq  \frac{b^2}{n^2} \frac{n}{n-b} \E_{B_t \in G_1} \E_{A_t \in G_0} \left[ \bigintsss_{\bxi_t} \frac{ \left(  dP_{B_t,\bxi_t} -   dP'_{B_t,\bxi_t} \right)^2 }{ dP_{A_t,\bxi_t}} d\bxi_t \right]~.
\end{align}
\label{lemm:lsd}
\vspace*{-6mm}
\end{lemm}

\proof By definition, Proposition \ref{prop:mixmodel}, and Lemma~\ref{lemm:mix_bound} with $s=b/n$, we have 
\begin{align*}
KL\left(P_{t|} \| P'_{t|}\right) & \stackrel{(a)}{\leq} \left(\frac{b}{n} \right)^2 \left(\frac{n}{n-b}\right) \int_{\w} \frac{(Q - Q')^2}{R} d\w \\
 & = \frac{b^2}{n^2} \left(1 + \frac{b}{n-b}\right) \bigintsss_{\w} \frac{ \left(  \frac{1}{|G_1|} \left[ \sum_{B \in G_1} dP_{B,\bxi}(\w) -  \sum_{B \in G_1} P'_{B,\bxi}(\w) \right] \right)^2 }{\frac{1}{|G_0|} \sum_{A \in G_0} dP_{A,\bxi}(\w)} d\w \\
 & \stackrel{(b)}{\leq} \frac{b^2}{n^2} \left(1 + \frac{b}{n-b}\right)  \bigintsss_{\w} \frac{1}{|G_1|}\sum_{B \in G_1}  \frac{ \left(  dP_{B,\bxi}(\w) -   P'_{B,\bxi}(\w) \right)^2 }{\frac{1}{|G_0|} \sum_{A \in G_0} dP_{A,\bxi}(\w)} d\w \\
 & \stackrel{(c)}{\leq} \frac{b^2}{n^2} \left(1 + \frac{b}{n-b}\right)  \bigintsss_{\w} \frac{1}{|G_1|}  \sum_{B \in G_1} \left\{ \left(  dP_{B,\bxi}(\w) -   P'_{B,\bxi}(\w) \right)^2  \frac{1}{|G_0|}\sum_{A \in G_0} \frac{ 1 }{ dP_{A,\bxi}(\w)} \right\} d\w \\
 & = \frac{b^2}{n^2} \left(1 + \frac{b}{n-b}\right) \frac{1}{|G_1| |G_0|}  \sum_{B \in G_1} \sum_{A \in G_0}  \bigintsss_{\w} \frac{ \left( dP_{B,\bxi}(\w) -   P'_{B,\bxi}(\w) \right)^2 }{ dP_{A,\bxi}(\w)} d\w ~,
\end{align*}
where (a) is from Lemma \ref{lemm:mix_bound}, (b)(c) is from Jensen's inequality since function $f(x) = x^2$ is convex and $f(x)=\frac{1}{x}$ is convex on $(0,\infty)$. That completes the proof. \qed

Now we have all the pieces to prove the following result:

\theostab*

\proof  Based on Propositions \ref{prop:hell} and \ref{prop:hellkl}, 
we have
\begin{align*}
\E_{S \sim D^n}[L_D(A(S)) - L_S(A(S))]
& \leq  \sqrt{K L\left(P_{T} \| P_{T}^{\prime}\right)} \leq \sqrt{ \sum_{t=1}^{T} \E_{P_{0:(t-1)}}\left[KL\left(P_{t \mid} \| P_{t \mid}^{\prime}\right)\right]} 
\end{align*}
Applying Lemma \ref{lemm:lsd} to bound $KL\left(P_{t \mid} \| P_{t \mid}^{\prime}\right)$ and noting that $b/(n-b) \leq 1$ for $b \leq n/2$ completes the proof. \qed

\subsection{High Probability Generalization Bounds}
\label{ssec:highp}




Let $S \sim D^n$ with $S=(Z_1\ldots,Z_n)$ corresponding to the training data. Consider the random variable $Y(S)$, the Scaled (by $n$) Generalization Error (SGE), defined as
\begin{align}
Y(S) \triangleq \sum_{i=1}^n \E_A \big[  \E_{Z \sim D} [\ell(A(S),Z) ] - \ell(A(S),Z_i) \big] = n\big(L_D(A(S)) - L_S(A(S))\big) ~.
\end{align}
Theorem~\ref{theo:stab} establishes a bound on $|\frac{1}{n}\E_S[Y(S)]|$. We now focus on establishing a high probability bound on $\frac{1}{n}(Y(S)-\E_S[Y(S)])$. Let $S' = (Z'_1,\ldots,Z'_n)$ be such that $Z'_i$ is an independent copy of $Z_i$. Further, let $S'_i = (Z_1,\ldots,Z_{i-1},Z'_i,Z_{i+1},\ldots,Z_n)$. Then, the change in SGE
\begin{equation}
Y(S) - Y(S'_i) = n \big[ \big( L_D(A(S)) - L_D(A(S'_i)) \big) - \big( L_S(A(S)) - L_{S'_i}(A(S'_i) \big) \big]    
\end{equation}
is a symmetric random variable, and is identically distributed for $i=1,\ldots,n$.
Our analysis is based on the following assumption: 
\begin{asmp}
The random variable $(Y(S)-Y(S'_i))^2$ is sub-Gaussian with $\psi_2$-norm $\kappa_A^2$, i.e., $\| (Y-Y'_i)^2 \|_{\psi_2} = \sup_{q \geq 1} (\E_{S.S'}|Y(S)-Y(S'_i)|^{2q})^{1/q}/\sqrt{q} = \kappa_A^2$. 
\label{asmp:2moment}
\end{asmp}
Assumption~\ref{asmp:2moment} implies that $\|Y-Y'_i\|_{\psi_2} \leq \kappa_A$.
Note that since $Y(S) - Y(S'_i)$ is identically distributed for all $i$, $\kappa_A$ is the same for all $i$. Further, $\kappa_A$ is a property of the algorithm $A$, and can be viewed as a measure of stability.
If $\kappa_A=O(1)$, i.e., swapping one point effectively leads to $O(\frac{1}{n})$ change in the generalization error, then $A$ can be considered stable; on the other hand, if $\kappa_A = O(\sqrt{n})$, then $A$ is not stable since the effective change in generalization error $\frac{1}{n}(Y(S)-Y(S'_i))$ is $O(1/\sqrt{n})$, the same order as typical generalization error $\frac{1}{n}Y(S)$ itself. The sharpness of the high-probability bound we present meaningfully depends on $\kappa_A$, with smaller values implying sharper bounds.

\begin{restatable}{theo}{theohighp}
Under Assumption~\ref{asmp:2moment}, with probability at least $(1-\delta)$ over the draw $S \sim D^n$, we have
\begin{equation}
L_D(A(S)) \leq L_S(A(S)) + \E_S\bigg[L_D(A(S) - L_S(A(S))\bigg] + \max\left( \frac{1}{n}, \frac{ c_1 \kappa_A}{\sqrt{n}} \right) \log \left(\frac{16}{\delta} \right)~.      
\end{equation}
\label{theo:hp}
\vspace*{-5mm}
\end{restatable}

If $\kappa_A=O(1)$, i.e., swapping one point effectively leads to $O(\frac{1}{n})$ change in the generalization error, then $A$ can be considered very stable; if $\kappa_A=O(\sqrt[4]{n})$, then $A$ can be considered somewhat stable; and if $\kappa^2_A = O(\sqrt{n})$, then $A$ is not stable since the effective change in generalization error $\frac{1}{n}(Y(S)-Y(S'_i))$ is $O(1/\sqrt{n})$, the same order as the generalization error $\frac{1}{n}Y(S)$ itself. 


%
%
\proof Let $S=(Z_1,\ldots,Z_n)$ and let $S' = (Z'_1,\ldots,Z'_n)$ be such that $Z'_i$ is an independent copy of $Z_i$. 
With $(x)_+ = max(x,0)$, let
\begin{align}
    V^+ = V^+(S) = \sum_{i=1}^n \E_{S'}[ (Y - Y'_i)_+^2 ]~,
\end{align}

Our proof uses the following exponential version of the Efron-Stein inequality [Theorem 6.16 in \cite{boucheron2013concentration}], whose proof is based on a combination of the symmetric modified log-Sobolev inequality [Theorem 6.15 in \cite{boucheron2013concentration}] with the change of measure:
\begin{theo}\label{theo:exp_efst}
Let $Y = f(Z_1,\ldots,Z_n)$, where $Z_i,i=1,\ldots,n$ are independent. Let $\theta,\lambda > 0$ be such that $\theta \lambda < 1$ and $\E[\exp(\lambda V^+/\theta)] < \infty$. Then,
\begin{equation}
\log \E \left[ e^{\lambda(Y - \E Y)} \right] \leq \frac{\lambda \theta}{1 - \lambda \theta} \log \E\left[ e^{\lambda V^+/\theta} \right]~.
\end{equation}
\end{theo}


For the proof of Theorem~\ref{theo:hp}, note that $V^+$ is sub-Gaussian with $\| V^+ \|_{\psi_2} \leq n \kappa_A^2$.
Let $\mu_+ = \E[V^+]$ and $\kappa_+ = \| V^+\|_{\psi_2}$. Then from Theorem \ref{theo:exp_efst}, we have
\begin{align*}
\psi(\lambda) \triangleq \log \E \left[ e^{\lambda(Y - \E Y)} \right] \leq \frac{\lambda \theta}{1 - \lambda \theta} \left[ \frac{\lambda \mu_+}{\theta} + c \frac{\lambda^2 \kappa_+^2}{\theta^2} \right].
\end{align*}
Then, by Markov's inequality we have
\begin{align*}
\log \P( Y - \E Y > t ) & \leq \psi(\lambda) -\lambda t \\
 & \leq \frac{1}{1-\lambda \theta} \left[ \lambda^2 \mu_+ + c\lambda^3 \kappa_+^2/\theta - (1-\lambda \theta) \lambda t \right]~.
\end{align*}
Choosing $\theta = 1/(2\lambda)$, we have 
\begin{align*}
\log \P( Y - \E Y > t ) 
& \leq 2 \lambda^2 \mu_+ + 4c \lambda^4 \kappa_+^2 - \lambda t~.
\end{align*}
Consider choosing $\lambda = \min \left\{ 1, \frac{1}{c_0\sqrt{2\kappa_+}} \right\}$, where $c_0 = \max(1,\sqrt[4]{c})$. Since $\mu_+ \leq \kappa_+$, we have 
\begin{align*}
\log \P( Y - \E Y > t ) & \leq 2 - \min \left\{1, \frac{1}{c_0 \sqrt{2\kappa_+}} \right\} t \\
\Rightarrow \quad \P( Y - \E Y > t ) & \leq 8 \exp \left( - \min \left\{1, \frac{1}{c_0 \sqrt{2\kappa_+}} \right\} t \right)~.
\end{align*}
With $t = n\epsilon$, and noting that $\kappa_+ \leq n \kappa_A^2$, with $c_1 = c_0 \sqrt{2}$, we have
\begin{align*}
\P\left( Y - \E Y >  n \epsilon \right) & \leq 8 \exp \left( - \min \left\{ n, \frac{\sqrt{n}}{c_1\kappa_A} \right\} \epsilon \right) ~.
\end{align*}
Now, we choose $\delta$ such that
\begin{align*}
 8 \exp \left( - \min \left\{ n, \frac{\sqrt{n}}{c_1\kappa_A} \right\} \epsilon \right)   
 \leq 8 \exp \left( - n \epsilon \right)  + 8 \exp \left( - \frac{\sqrt{n}}{c_1\kappa_A} \epsilon \right) \leq \delta~.
\end{align*}
It suffices to choose $\epsilon$ such that
\begin{align*}
8 \exp \left( - n \epsilon \right) \leq \delta/2 & \qquad \Rightarrow \qquad \epsilon \geq \frac{1}{n} \log \left( \frac{16}{\delta} \right) \\
8 \exp \left( - \frac{\sqrt{n}}{c_1\kappa_A} \epsilon \right) \leq \delta/2 & \qquad \Rightarrow \qquad \epsilon \geq \frac{c_1 \kappa_A}{\sqrt{n}} \log \left( \frac{16}{\delta} \right)~.
\end{align*}
As a result,
\begin{align*}
\P\left( \frac{1}{n}(Y - \E Y) >  \max \left( \frac{1}{n}, \frac{c_1 \kappa_A}{\sqrt{n}} \right) \log \left( \frac{16}{\delta} \right) \right) & \leq \delta~.
\end{align*}
That completes the proof. \qed

%% file: arXiv2022/app/efld.tex

In this section, we provide the proofs for Section \ref{sec:efld}. We first review and show details of a few examples of exponential family.

\subsection{Examples of Exponential Family}
\label{app:example}

We show that EFLD becomes SGLD when the exponential family is Gaussian, and becomes a noisy version of sign-SGD \citep{bernstein2018signsgd, bernstein2018signsgdnoise} when the exponential family is skewed  over $\{-1,+1\}$. 
\begin{exam}[Gaussian] 
SGLD uses scaled Gaussian noise with $\psi(\btheta)= \|\btheta\|_2^2/2, \alpha_t = \sigma_t/\eta_t$, 
$\ppi_{0,\alpha_t}(\bxi) = \frac{1}{\sqrt{(2\pi)^p \alpha_t^p}} \exp(-\|\bxi\|_2^2/2\alpha_t^2)$ so that $p_{\psi}(\bxi; \btheta_{B_t,\alpha_t}) = \mathcal{N}(\btheta_{B_t}, \alpha_t^2 \I_d)$. 
Then, the distribution from the natural parameter form is:
\begin{equation}
    p_{\btheta/\sigma}(\xi) = \exp( \langle\xi, \btheta\rangle / \sigma^2 - \|\theta\|_2^2/(2\sigma^2) ) \times \frac{1}{\sqrt{2\pi}\sigma} \exp(-\|\xi\|_2^2/2\sigma^2)
    = \frac{1}{\sqrt{2\pi}\sigma} \exp(-\|\bxi-\bmu\|_2^2/2\sigma^2)~.
\end{equation}
The expectation parameter $\bmu = \nabla \psi(\btheta) = \btheta$. The scaled expectation parameter $\bmu_{\alpha} = \nabla \psi(\btheta_\alpha) = \btheta_\alpha = \btheta/\sigma = \bmu/\sigma$. Since $\xi = x/\sigma$ and the Bregman divergence $d_{\phi}(\bxi,\bmu) = \frac{1}{2}\|\bxi-\bmu\|_2^2$, the expectation parameter form
\begin{equation}
p_{\bmu_{\alpha}}(\bxi) = \exp(-\|\bxi/\sigma - \bmu/\sigma\|_2^2/2) \frac{1}{\sqrt{2\pi} \sigma} = \frac{1}{\sqrt{2\pi}\sigma} \exp(-\|\bxi-\bmu\|_2^2/2\sigma^2)~.
\end{equation}
By taking $\rho_t = \eta_t$, the update \eqref{eq:efld1} based on $\rho_t \bxi_t$ is distributed as $\mathcal{N}(\eta_t \btheta_{B_t}, \eta_t^2 \alpha_t^2 \I_d) = \mathcal{N}(\eta_t\nabla \ell(\w_{t-1}, S_{B_t}), \sigma_t^2 \I_d)$. Thus the EFLD update reduces to the SGLD update: $\w_{t} = \w_{t-1} - \eta_t \nabla \ell(\w_{t-1}, S_{B_t}) + \mathcal{N}\left(0, \sigma_t^2 \mathbb{I}_{d}\right).$
\qed 
\end{exam}
\begin{exam}[Skewed Rademacher]
\label{exam:noisy_signsgd}
For skewed Rademacher over $\{-1,1\}$, the sufficient statistic $\xi \in \{-1,1\}$, base measure is 1 on $\{-1,1\}$, and the log-partition function $\psi(\theta) = \log(\exp(-\theta)+\exp(\theta))$ for natural parameter $\theta \in \R$. The expectation parameter $\mu = \nabla \psi(\theta) = \frac{\exp(\theta)- \exp(-\theta)}{\exp(-\theta)+\exp(\theta)} = \frac{\exp(2\theta)-1}{\exp(2\theta)+1}$, then its inverse function $\theta = \nabla \phi(\mu) = \frac{1}{2}\log \left(\frac{1+\mu}{1-\mu}\right)$, by integration we have $\phi(\mu) = \frac{1+\mu}{2}\log\frac{1+\mu}{2}+ \frac{1-\mu}{2}\log\frac{1-\mu}{2}$. The expectation parameter $\mu = \nabla \psi(\theta) = \frac{\exp(\theta)}{1+\exp(\theta)} = \frac{1}{1+\exp(-\theta)}$, the sigmoid function of $\theta$. The Bregman divergence is the Bernoulli KL-divergence given by: $d_{\phi}(\xi,\mu) = \xi \log \frac{\xi}{\mu} + (1-\xi) \log \frac{1-\xi}{1-\mu}$. 
For scaled parameters $\theta_{\alpha} = \theta/\alpha$, the corresponding expectation parameter $\mu_{\alpha} = \frac{1}{1+\exp(-\theta_{\alpha})}$. 
The mean parameter form distribution is given by
\begin{equation}
p_{\mu_{\alpha}}(\xi) = \exp(- d_{\phi}(\xi,\mu_{\alpha})) = \mu_{\alpha}^{\xi} (1-\mu_{\alpha})^{1-\xi}~.
\end{equation}
Noisy Sign-SGD takes $\rho_t = \eta_t$ and componentwise $\xi_j \in\{-1,1\},~\pi_{0,\alpha_t}(\xi_j)=1,~\psi(\theta) = \log(\exp(-\theta)+\exp(\theta))$ in exponential family update equation \eqref{eq:efld1}, the $j$-th component of exponential family distribution $p_{\psi}(\bxi; \btheta_{B_t,\alpha_t})$ becomes
\begin{equation}
p_{\psi}(\xi_j; \btheta_{B_t,\alpha_t,j})  = \frac{\exp(\xi_j \btheta_{B_t,\alpha_t,j})}{\exp(-\theta_{B_t,\alpha_t,j})+\exp(\theta_{B_t,\alpha_t,j})}~.
\end{equation}
Thus, the EFLD update reduces to a noisy version of Sign-SGD: 
$\w_{t} = \w_{t-1} - \eta_t \bxi_t$, $\xi_{t,j} \sim p_{\psi}(\xi_j,\theta_{B_t,\alpha_t,j})$, 
$j\in [p]$, where $\btheta_{B_t,\alpha_t} = \nabla \ell(\w_{t-1}, S_{B_t})/\alpha_t$ is the scaled mini-batch gradient.
\qed
\end{exam}

\begin{exam}[Bernoulli over $\{0,1\}$]
For Bernoulli over $\{0,1\}$, the sufficient statistic $\xi = x \in \{0,1\}$, base measure is 1 on $\{0,1\}$, and the log-partition function $\psi(\theta) = \log(1+\exp(\theta))$ for natural parameter $\theta \in \R$. 
The expectation parameter $\mu = \nabla \psi(\theta) = \frac{exp(\theta)}{1+\exp(\theta)} = \frac{1}{1+\exp(-\theta)}$, the sigmoid function of $\theta$. The Bregman divergence is the Bernoulli KL-divergence given by: $d_{\phi}(\xi,\mu) = x \log \frac{x}{\mu} + (1-x) \log \frac{1-x}{1-\mu}$. For scaled parameters $\theta_{\alpha} = \theta/\alpha$, the corresponding expectation parameter $\mu_{\alpha} = \frac{1}{1+\exp(-\theta_{\alpha})}$. 
The mean parameter form distribution is given by
\begin{equation}
p_{\mu_{\alpha}}(\xi) = \exp(- d_{\phi}(\xi,\mu_{\alpha})) = \mu_{\alpha}^{\xi} (1-\mu_{\alpha})^{1-\xi}~.
\end{equation}
By taking the $j$-th component of exponential family distribution $p_{\psi}(\bxi; \btheta_{B_t,\alpha_t})$ becomes
$p_{\psi}(\xi_j; \theta_{B_t,\alpha_t,j})  = \frac{\exp(\xi_j \theta_{B_t,\alpha_t,j})}{1+\exp(\theta_{B_t,\alpha_t,j})}$.
Thus, the EFLD update reduces to : 
$\w_{t} = \w_{t-1} - \eta_t \bxi_t$, $\xi_{t,j} \sim p_{\psi}(\xi_j,\btheta_{B_t,\alpha_t,j})$, $j\in [p]$, where $\btheta_{B_t,\alpha_t} = \nabla \ell(\w_{t-1}, S_{B_t})/\alpha_t$ is the scaled mini-batch gradient.
\qed
\end{exam}

\subsection{Proof of Theorem~\ref{theo:efld}}

\theoefld*

To avoid clutter, we drop the subscript $t$ for the analysis and note that the analysis holds for any step $t$.  When the density $dP_{B,\bxi} = \p_{\psi}(\bxi;\btheta_{B,\alpha})$, by mean-value theorem, for each $\bxi$, we have 
\begin{equation}
p_{\psi}(\bxi; \btheta_{B,\alpha}) - p_{\psi}(\bxi; \btheta_{B',\alpha}) 
= \langle \btheta_{B,\alpha} -\btheta_{B',\alpha}, \nabla_{\tilde{\btheta}_{B,\alpha}} p_{\psi}(\bxi;\tilde{\btheta}_{B,\alpha})\rangle ~,
\end{equation}
for some $\tilde{\btheta}_{B,\alpha} = \gamma_{\bxi} \btheta_{B,\alpha} + (1-\gamma_{\bxi}) \btheta'_{B,\alpha}$ where $\gamma_{\bxi} \in [0,1]$. Then,
\begin{align}
I_{A,B} &=  \bigintsss_\bxi \frac{\big( p_{\psi}(\bxi; \btheta_{B,\alpha}) - p_{\psi}(\bxi; \btheta_{B',\alpha}) \big)^2}{p_{\psi}(\bxi; \btheta_{A,\alpha})} d\bxi \nonumber 
= \bigintsss_\bxi \frac{\langle \btheta_{B,\alpha} -\btheta_{B',\alpha}, \nabla_{\tilde{\btheta}_{B,\alpha}} p_{\psi}(\bxi;\tilde{\btheta}_{B,\alpha})\rangle ^2}{p_{\psi}(\bxi; \btheta_{A,\alpha})} d\bxi \nonumber \\
& = \bigintsss_\bxi \frac{\langle \btheta_{B,\alpha} -\btheta'_{B,\alpha}, \bxi - \nabla_{\tilde{\btheta}_{B',\alpha}} \psi(\bxi;\tilde{\btheta}_{B,\alpha}) \rangle ^2p_{\psi}^2(\bxi;\tilde{\btheta}_{B,\alpha})}{p_{\psi}(\bxi; \btheta_{A,\alpha})} d\bxi ~,
\label{eq:expcore}
\end{align}
since $p_{\psi}(\bxi;\tilde{\btheta}_{B,\alpha}) = \exp(\langle \bxi, \tilde{\btheta}_{B,\alpha} \rangle - \psi(\tilde{\btheta}_{B,\alpha})) \pi_0(\bxi)$.

\subsubsection{Handling Distributional Dependency of $\tilde{\btheta}_B$}
Note that we cannot proceed with the analysis with the density term depending on $\tilde{\btheta}_B$ since $\tilde{\btheta}_B$ depends on $\bxi$. In this step we focus on bounding the density term depending on $\tilde{\btheta}_B$ in terms of exponential family densities with parameters $\btheta_B$ and $\btheta'_B$.

\exptilde*

\proof Denoting $\gamma_{\bxi}$ as $\gamma$ for convenience (the dependence on $\bxi$ does not play a role in the analysis), we have 
\begin{align*}
\langle \bxi, & \tilde{\btheta}_{B,  \alpha} \rangle  - \psi(\tilde{\btheta}_{B,\alpha}) 
 = \langle \bxi, \gamma \btheta_{B, \alpha} + (1-\gamma) \btheta'_{B,\alpha} \rangle - \psi\left( \gamma \btheta_{B, \alpha} + (1-\gamma) \btheta'_{B, \alpha} \right) \\
& = \gamma \bigg[ \langle \bxi , \btheta_{B,\alpha} \rangle - \psi(\btheta_{B, \alpha}) \bigg]  + (1-\gamma) \bigg[ \langle \bxi , \btheta'_{B, \alpha} \rangle - \psi(\btheta'_{B,\alpha}) \bigg] \\
& ~~~~~~~~~~~~~~~
  + \bigg[ \gamma \psi(\btheta_{B, \alpha} ) + (1-\gamma) \psi(\btheta'_{B,\alpha}) - \psi( \gamma \btheta_{B,\alpha} + (1-\gamma) \btheta'_{B,\alpha} ) \bigg] \\
& \stackrel{(a)}{\leq} \max\bigg( \langle \bxi , \btheta_{B, \alpha} \rangle - \psi(\btheta_{B,\alpha}) , \langle \bxi , \btheta'_{B, \alpha} \rangle - \psi(\btheta'_{B,\alpha} )  \bigg) \\
& ~~~~~~~~~~~~~~~ 
  + \bigg[ \gamma d_{\psi}(\btheta_{B,\alpha}, \tilde{\btheta}_{B,\alpha}) + (1-\gamma) d_{\psi}(\btheta'_{B,\alpha} , \tilde{\btheta}_{B,\alpha}) \bigg]~,
\end{align*}
where the second term in (a) follows since difference between two sides of Jensen's inequality is given by the Bregman information, i.e., expected Bregman divergence to the expectation (See section 3.1.1 in \cite{banerjee2005clustering}):
\begin{align*}
  \gamma \psi(\btheta_{B, \alpha} ) + (1-\gamma) \psi(\btheta'_{B,\alpha}) - \psi( \gamma \btheta_{B,\alpha} + (1-\gamma) \btheta'_{B,\alpha} )  =  \gamma d_{\psi}(\btheta_{B,\alpha}, \tilde{\btheta}_{B,\alpha}) + (1-\gamma) d_{\psi}(\btheta'_{B,\alpha} , \tilde{\btheta}_{B,\alpha})~.
\end{align*}
Now, note that
\begin{align*}
\gamma d_{\psi}(\btheta_{B,\alpha}, \tilde{\btheta}_{B,\alpha})  + (1-\gamma) d_{\psi}(\btheta'_{B,\alpha}, \tilde{\btheta}_{B,\alpha})  
& \leq \gamma d_{\psi}(\btheta_{B,\alpha}, \btheta'_{B,\alpha}) + (1-\gamma) d_{\psi}(\btheta'_{B,\alpha}, \btheta_{B,\alpha}) \\
& \leq \gamma c_2 \| \btheta'_{B,\alpha} - \btheta_{B,\alpha} \|_2^2 + (1-\gamma) c_2 \| \btheta_{B,\alpha} - \btheta'_{B,\alpha} \|_2^2 \\
& \leq c_2 \| \btheta_{B,\alpha} - \btheta_{B',\alpha} \|_2^2~.
\end{align*}
As a result,
\begin{align*}
& \frac{\exp\left[\langle \bxi, \tilde{\btheta}_{B, \alpha} \rangle - \psi(\tilde{\btheta}_{B,\alpha}) \right]}{\max\left( \exp\left[\langle \bxi, \btheta_{B, \alpha} \rangle - \psi(\btheta_{B,\alpha}) \right] , \exp\left[\langle \bxi, \btheta'_{B, \alpha} \rangle - \psi(\btheta'_{B,\alpha}) \right] \right) }\\
& ~~~~~ \leq\exp\left[ c_2 \| \btheta_{B,\alpha} - \btheta_{B',\alpha} \|_2^2 \right]
\frac{ \exp \left[ \max\bigg( \langle \bxi , \btheta_{B, \alpha} \rangle - \psi(\btheta_{B,\alpha}) , \langle \bxi , \btheta'_{B,\alpha} \rangle - \psi(\btheta'_{B,\alpha})  \bigg) \right] }{ \max\left( \exp\left[\langle \bxi, \btheta_{B,\alpha} \rangle - \psi(\btheta_{B,\alpha}) \right] , \exp\left[\langle \bxi, \btheta'_{B, \alpha} \rangle - \psi(\btheta'_{B,\alpha}) \right] \right) } \\
& ~~~~~ = \exp\left[ c_2 \| \btheta_{B,\alpha} - \btheta_{B',\alpha} \|_2^2 \right]~.
\end{align*}
That completes the proof.\qed

\subsubsection{Bounding the Density Ratio}

Focusing on density ratio, we have

\expdr*

\proof  Note that
\begin{align*}
\frac{\exp\left[\langle \bxi, 2\btheta_{B,\alpha} \rangle - 2\psi(\btheta_{B,\alpha}) \right] }{\exp\left[\langle \bxi, \btheta_{A , \alpha} \rangle - \psi(\btheta_{A,\alpha}) \right]}  
& = \exp\left[ \langle \bxi, (2\btheta_{B,\alpha} - \btheta_{A,\alpha}) \rangle - \big( 2\psi(\btheta_{B,\alpha}) - \psi(\btheta_{A,\alpha}) \big) \right] \\
& = \beta_{B,A,\alpha} \exp \left[  \langle \bxi, (2\btheta_{B,\alpha} - \btheta_{A,\alpha} \rangle - \psi( (2\btheta_{B,\alpha} - \btheta_{A,\alpha}) \right]~,
\end{align*}
where
\begin{align*}
\beta_{B,A,\alpha} & = \exp \left[ \psi( (2\btheta_{B,\alpha} - \btheta_{A,\alpha}) - \big( 2\psi(\btheta_{B,\alpha}) - \psi(\btheta_{A,\alpha}) \big) \right] ~.
\end{align*}
Note that
\begin{align*}
\log \beta_{B,A,\alpha} & = \psi( (2\btheta_{B,\alpha} - \btheta_{A,\alpha}) - \big( 2\psi(\btheta_{B,\alpha}) - \psi(\btheta_{A,\alpha}) \big)\\
& = - \bigg( \psi(\btheta_{B,\alpha}) - \psi( (2\btheta_{B,\alpha} - \btheta_{A,\alpha}) \bigg) - \bigg( \psi(\btheta_{B,\alpha}) - \psi(\btheta_{A,\alpha}) \bigg) \\
& \stackrel{(a)}{\leq} \bigg\langle \btheta_{B,\alpha} - \btheta_{A,\alpha}, \nabla \psi( 2\btheta_{B,\alpha} - \btheta_{A,\alpha}) \bigg\rangle + \bigg\langle \btheta_{A,\alpha} - \btheta_{B,\alpha}, \nabla \psi( \btheta_{A,\alpha} \bigg\rangle   \\
& = \bigg\langle \btheta_{B,\alpha} - \btheta_{A,\alpha}, \nabla \psi( (2\btheta_{B,\alpha} - \btheta_{A,\alpha}) - \nabla \psi(\btheta_{A,\alpha})  \bigg\rangle  \\
& \stackrel{(b)}{\leq} \| \btheta_{B,\alpha} - \btheta_{A,\alpha} \|_2 \| \nabla \psi( (2\btheta_{B,\alpha} - \btheta_{A,\alpha}) - \nabla \psi(\btheta_{A,\alpha}) \|_2  \\
& \stackrel{(c)}{\leq} c_2 \| \btheta_{B,\alpha} - \btheta_{A,\alpha} \|_2 \| (2\btheta_{B,\alpha} - \btheta_{A,\alpha}) - \btheta_{A,\alpha} \|_2 \\
& = 2 c_2 \| \btheta_{B,\alpha} - \btheta_{A,\alpha} \|_2^2~,
\end{align*}
where (a) follows from the convexity of $\psi$, (b) follows from Cauchy-Schwartz, and (c) follows by smoothness of $\psi$.

As a result, we have 
\begin{align*}
\frac{\exp\left[\langle \bxi, 2\btheta_{B,\alpha} \rangle - 2\psi(\btheta_{B,\alpha}) \right] }{\exp\left[\langle \bxi, \btheta_{A, \alpha} \rangle - \psi(\btheta_{A,\alpha}) \right]}
&\leq \exp \left[ 2c_2 \| \btheta_{B,\alpha} - \btheta_{A,\alpha} \|_2^2 \right] \exp \left[  \langle \bxi, (2\btheta_B - \btheta_A)/\alpha \rangle - \psi( (2\btheta_B - \btheta_A)/\alpha) \right]~.
\end{align*}
That completes the proof.\qed

The analysis for the term involving $\btheta'_B$ is exactly the same. 



Combining Lemma~\ref{lemm:exp_step2} and \ref{lemm:exp_step3}, we have the following result:
\begin{restatable}{lemm}{expcompile}
We have
\begin{equation}
\begin{split}
\frac{p^2_{\psi}(\bxi;\tilde{\btheta}_{B,\alpha})}{p_{\psi}(\bxi;\btheta_{A,\alpha})} 
 & \leq \exp\left[ 2c_2 \| \btheta_{B,\alpha} - \btheta_{B',\alpha} \|_2^2 \right] \\
 & \phantom{...........} 
\times \max\big( \exp \left[ 2c_2 \| \btheta_{B,\alpha} - \btheta_{A,\alpha} \|_2^2 \right] p_{\psi}(\bxi; 2 \btheta_{B,\alpha} - \btheta_{A,\alpha})~,  \big. \\
& \phantom{........................} \left. 
\exp \left[2c_2 \| \btheta_{B',\alpha} - \btheta_{A,\alpha} \|_2^2 \right] p_{\psi}(\bxi; 2 \btheta_{B',\alpha} - \btheta_{A,\alpha}) \right)~.
\end{split}
\end{equation}
\label{corr:expcompile}
\end{restatable}
\proof

From Lemma \ref{lemm:exp_step2}, we have
\begin{align*}
p_{\psi}^2(\boldsymbol{\xi} ; \tilde{\boldsymbol{\theta}}_{B, \alpha}) & = \exp \left[2\left\langle\boldsymbol{\xi}, \tilde{\boldsymbol{\theta}}_{B, \alpha}\right\rangle-2\psi\left(\tilde{\boldsymbol{\theta}}_{B, \alpha}\right)\right] \\
& \leq \exp \left[c_{2}\left\|\boldsymbol{\theta}_{B, \alpha}-\boldsymbol{\theta}_{B^{\prime}, \alpha}\right\|_{2}^{2}\right] \cdot \max \left(\exp \left[\left\langle\boldsymbol{\xi}, \boldsymbol{\theta}_{B, \alpha}\right\rangle-\psi\left(\boldsymbol{\theta}_{B, \alpha}\right)\right], \exp \left[\left\langle\boldsymbol{\xi}, \boldsymbol{\theta}_{B^{\prime}, \alpha}\right\rangle-\psi\left(\boldsymbol{\theta}_{B^{\prime}, \alpha}\right)\right]\right)
\end{align*}

Thus,

\begin{align*}
\frac{p_{\psi}^{2}\left(\boldsymbol{\xi} ; \tilde{\boldsymbol{\theta}}_{B, \alpha}\right)}{p_{\psi}\left(\boldsymbol{\xi} ; \boldsymbol{\theta}_{A, \alpha}\right)} \leq \exp \left[2 c_{2}\left\|\boldsymbol{\theta}_{B, \alpha}-\boldsymbol{\theta}_{B^{\prime}, \alpha}\right\|_{2}^{2}\right] \cdot  \frac{\max \left(\exp \left[2\left\langle\boldsymbol{\xi}, \boldsymbol{\theta}_{B, \alpha}\right\rangle-2\psi\left(\boldsymbol{\theta}_{B, \alpha}\right)\right], \exp \left[2\left\langle\boldsymbol{\xi}, \boldsymbol{\theta}_{B^{\prime}, \alpha}\right\rangle-2\psi\left(\boldsymbol{\theta}_{B^{\prime}, \alpha}\right)\right]\right)}{\exp \left[\left\langle\boldsymbol{\xi}, \boldsymbol{\theta}_{A, \alpha}\right\rangle-\psi\left(\boldsymbol{\theta}_{A, \alpha}\right)\right]}
\end{align*}

Based on Lemma \ref{lemm:exp_step3}, we have
\begin{align*}
\frac{\exp \left[\left\langle\boldsymbol{\xi}, 2 \boldsymbol{\theta}_{B, \alpha}\right\rangle-2 \psi\left(\boldsymbol{\theta}_{B, \alpha}\right)\right]}{\exp \left[\left\langle\boldsymbol{\xi}, \boldsymbol{\theta}_{A, \alpha}\right\rangle-\psi\left(\boldsymbol{\theta}_{A, \alpha}\right)\right]} \leq \exp \left[2 c_{2}\left\|\boldsymbol{\theta}_{B, \alpha}-\boldsymbol{\theta}_{A, \alpha}\right\|_{2}^{2}\right] \exp \left[\left\langle\boldsymbol{\xi},\left(2 \boldsymbol{\theta}_{B}-\boldsymbol{\theta}_{A}\right) / \alpha\right\rangle-\psi\left(\left(2 \boldsymbol{\theta}_{B}-\boldsymbol{\theta}_{A}\right) / \alpha\right)\right]
\end{align*}

Combining above inequalities we have
\begin{equation}
\begin{split}
\frac{p^2_{\psi}(\bxi;\tilde{\btheta}_{B,\alpha})}{p_{\psi}(\bxi;\btheta_{A,\alpha})} 
 & \leq \exp\left[ 2c_2 \| \btheta_{B,\alpha} - \btheta_{B',\alpha} \|_2^2 \right] \\
 & \phantom{...........} 
\times \max\big( \exp \left[ 2c_2 \| \btheta_{B,\alpha} - \btheta_{A,\alpha} \|_2^2 \right] p_{\psi}(\bxi; 2 \btheta_{B,\alpha} - \btheta_{A,\alpha})~,  \big. \\
& \phantom{........................} \left. 
\exp \left[2c_2 \| \btheta_{B',\alpha} - \btheta_{A,\alpha} \|_2^2 \right] p_{\psi}(\bxi; 2 \btheta_{B',\alpha} - \btheta_{A,\alpha}) \right)~.
\end{split}
\end{equation}

That completes the proof. \qed

\subsubsection{Bounding the Integral}

Ignoring multiplicative terms which do not depend on $\bxi$ for the moment, the analysis needs to bound an integral term of the form
\begin{equation}
    \bigintsss_\bxi \langle \btheta_{B,\alpha} -\btheta'_{B,\alpha}, \bxi - \nabla \psi(\bxi;\tilde{\btheta}_{B,\alpha}) \rangle ^2 ~p_{\psi}(\bxi;2 \btheta_{B,\alpha} - \btheta_{A,\alpha}) d\bxi~,
\end{equation}
and a similar term with $p_{\psi}^2(\bxi;2 \btheta_{B',\alpha} - \btheta_{A,\alpha})$. The proof of Theorem~\ref{theo:efld} can be done by suitably bounding the integral.

{\em Proof of Theorem~\ref{theo:efld}.} From Lemma~\ref{corr:expcompile}, we have
\begin{align*}
&I_{A,B}\leq \bigintsss_\bxi \frac{ \left\langle \btheta_{B,\alpha} -\btheta'_{B,\alpha}, \bxi - \nabla \psi( \tilde{\btheta}_{B,\alpha}) \right\rangle^2 p^2_{\psi}(\bxi;\tilde{\btheta}_{B,\alpha})}{ p_{\psi}(\bxi;\btheta_{A,\alpha})} d\bxi\\
&\leq  \exp\left[ 2 c_2 \| \btheta_{B,\alpha} - \btheta_{B',\alpha} \|_2^2\right] \\
& ~~~~~~~~~~~~~~~~~~~~\times \max\left(  \exp \left[ 2 c_2 \| \btheta_{B,\alpha} - \btheta_{A,\alpha} \|_2^2 \right] 
\bigintsss_\bxi  \left\langle \btheta_{B,\alpha} -\btheta_{B,\alpha}, \bxi - \nabla \psi(\tilde{\btheta}_{B,\alpha}) \right\rangle^2 p_{\psi}(\bxi;2\btheta_{B,\alpha}-\btheta_{A,\alpha}) d\bxi \right. ~, \\
& ~~~~~~~~~~~~~~~~~~~~~~~~~~~
\left.   \exp \left[ 2c_2 \| \btheta_{B',\alpha} - \btheta_{A,\alpha} \|_2^2 \right] 
\bigintsss_\bxi  \left\langle \btheta_{B,\alpha} -\btheta_{B',\alpha}, \bxi - \nabla \psi(\tilde{\btheta}_{B,\alpha}) \right\rangle^2 p_{\psi}(\bxi;2\btheta_{B',\alpha}-\btheta_{A,\alpha}) d\bxi \right)~.
\end{align*}
Focusing on the integral in the first term (the analysis for the second term is essentially the same), we have
\begin{align*}
& \bigintsss_\bxi  \left\langle \btheta_{B,\alpha} -\btheta_{B',\alpha}, \bxi - \nabla \psi(\tilde{\btheta}_{B,\alpha}) \right\rangle^2 p_{\psi}(\bxi; 2\btheta_{B,\alpha}-\btheta_{A,\alpha}) d\bxi \\
& = \bigintsss_\bxi  \left\langle \btheta_{B,\alpha} -\btheta_{B',\alpha}, \big( \bxi - \E[\bxi] \big) - \left(\nabla \psi(\tilde{\btheta}_{B,\alpha}) - \E[\bxi] \right) \right\rangle^2 p_{\psi}(\bxi;2\btheta_{B,\alpha}-\btheta_{A,\alpha}) d\bxi \\
& \leq 2\underbrace{\bigintsss_\bxi  \left\langle \btheta_{B,\alpha} -\btheta_{B',\alpha}, \bxi - \E[\bxi] \right\rangle^2 p_{\psi}(\bxi;2\btheta_{B,\alpha}-\btheta_{A,\alpha}) d\bxi}_{T_1} \\
& \phantom{\leq} + 2 \underbrace{\bigintsss_\bxi  \left\langle \btheta_{B,\alpha} -\btheta_{B,\alpha}, \nabla \psi(\tilde{\btheta}_{B,\alpha}) - \E[\bxi] \right\rangle^2 p_{\psi}(\bxi;2\btheta_{B,\alpha}-\btheta_{A,\alpha}) d\bxi}_{T_2}~.
\end{align*}
For $T_1$, note that 
\begin{align*}
T_1 & = \E_{\bxi \sim p_{\psi}(\bxi;2\btheta_{B,\alpha}-\btheta_{A,\alpha})}\left[ \left\langle \btheta_{B,\alpha} -\btheta_{B',\alpha}, \bxi - \E[\bxi] \right\rangle^2   \right] \\
& = (\btheta_{B,\alpha} -\btheta_{B',\alpha})^T \E_{\bxi \sim p_{\psi}(2\btheta_{B,\alpha}-\btheta_{A,\alpha})}\left[ (\bxi - \E[\bxi]) (\bxi - \E[\bxi])^T \right] (\btheta_{B,\alpha} -\btheta_{B',\alpha})~\\
& = (\btheta_{B,\alpha} -\btheta_{B,\alpha})^T \nabla^2 \psi( 2\btheta_{B,\alpha}-\btheta_{A,\alpha})  (\btheta_{B,\alpha} -\btheta_{B',\alpha}) \\
& \leq c_2 \| \btheta_{B,\alpha} -\btheta_{B,\alpha} \|_2^2~,
\end{align*}
since, by smoothness, the spectral norm of $\nabla^2 \psi$ is bounded by $c_2$. 

For $T_2$, first note that 
\begin{align*}
\E_{\bxi \sim p_{\psi}(\bxi;2\btheta_{B,\alpha}-\btheta_{A,\alpha})}[\bxi] = \nabla \psi( 2\btheta_{B,\alpha} - \btheta_{A,\alpha}) ) ~.
\end{align*}
Hence, with $\tilde{\btheta}_{B,\alpha} = \gamma_{\bxi} \btheta_{B,\alpha} + (1-\gamma_{\bxi}) \btheta_{B',\alpha}$ for some $\gamma_{\bxi} \in [0,1]$, we have
\begin{align*}
T_2 & = \E_{\bxi \sim p_{\psi}(\bxi;2\btheta_{B,\alpha}-\btheta_{A,\alpha})}\left[ \left\langle \btheta_{B,\alpha} -\btheta_{B,\alpha},  \nabla \psi(\tilde{\btheta}_{B,\alpha}) - \E[\bxi] \right\rangle^2   \right] \\    
& = \E_{\bxi \sim p_{\psi}(\bxi; 2\btheta_{B,\alpha}-\btheta_{A,\alpha})}\left[ \left\langle \btheta_{B,\alpha} -\btheta_{B',\alpha},  \nabla \psi(\tilde{\btheta}_{B,\alpha}) - \nabla \psi( 2\btheta_{B,\alpha} - \btheta_{A,\alpha})  \right\rangle^2   \right]\\
& \leq  \| \btheta_{B,\alpha} -\btheta_{B,\alpha} \|_2^2 ~~~\E_{\bxi \sim p_{\psi}(\bxi;2\btheta_{B,\alpha}-\btheta_{A,\alpha})}\left[ \left\|  \nabla \psi(\tilde{\btheta}_{B,\alpha}) - \nabla \psi( (2\btheta_{B,\alpha} - \btheta_{A,\alpha}) \right\|_2^2  \right]\\
& = c_2^2 \| \btheta_{B,\alpha} -\btheta_{B',\alpha} \|_2^2~~~ \E_{\bxi \sim p_{\psi}(\bxi;2\btheta_{B,\alpha}-\btheta_{A,\alpha})}\left[ \left\|  \tilde{\btheta}_{B,\alpha} - (2\btheta_{B,\alpha} - \btheta_{A,\alpha} ) \right\|_2^2  \right] \\
& = c_2^2 \| \btheta_{B,\alpha} -\btheta_{B',\alpha} \|_2^2~~~ \E_{\bxi \sim p_{\psi}(\bxi;2\btheta_{B,\alpha}-\btheta_{A,\alpha})} \left[ \big\| \gamma_{\bxi} \btheta_{B,\alpha} + (1-\gamma_{\bxi}) \btheta_{B',\alpha} - \btheta_{B,\alpha} - (\btheta_{B,\alpha} - \btheta_{A,\alpha}) \big\|_2^2 \right] \\
& = c_2^2 \| \btheta_{B,\alpha} -\btheta_{B',\alpha} \|_2^2~~~ \E_{\bxi \sim p_{\psi}(\bxi;2\btheta_{B,\alpha}-\btheta_{A,\alpha})}
\left[ \big\|  (1-\gamma_{\bxi}) (\btheta_{B',\alpha} - \btheta_{B,\alpha}) - (\btheta_{B,\alpha} - \btheta_{A,\alpha}) \big\|_2^2 \right] \\
& \leq  2c_2^2 \| \btheta_{B,\alpha} -\btheta_{B',\alpha} \|_2^2~~~ 
\bigg( \E_{\bxi \sim p_{\psi}(\bxi;2\btheta_{B,\alpha}-\btheta_{A,\alpha})}\left[ (1-\gamma_{\bxi})^2  \big\|  \btheta_{B',\alpha} - \btheta_{B,\alpha} \big\|_2^2 \right]\\
&\qquad+  \E_{\bxi \sim p_{\psi}(\bxi;2\btheta_{B,\alpha}-\btheta_{A,\alpha})} \left[ \big\|  \btheta_{B,\alpha} - \btheta_{A,\alpha} \big\|_2^2 \right]  \bigg) \\
& = 2c_2^2 \| \btheta_{B,\alpha} -\btheta_{B',\alpha} \|_2^2~~~ 
\left( \big\|  \btheta_{B',\alpha} - \btheta_{B,\alpha} \big\|_2^2 \E_{\bxi \sim p_{\psi}(\bxi;2\btheta_{B,\alpha}-\btheta_{A,\alpha})}\left[ (1-\gamma_{\bxi})^2   \right]
+   \big\|  \btheta_{B,\alpha} - \btheta_{A,\alpha} \big\|_2^2  \right) \\
& \leq  2c_2^2 \| \btheta_{B,\alpha} -\btheta_{B',\alpha} \|_2^2~~~  \left( \| \btheta_{B,\alpha} -\btheta_{B',\alpha}\|_2^2  + \| \btheta_{B,\alpha} - \btheta_{A,\alpha} \|_2^2 \right) ~.
\end{align*}
Putting everything back together
\begin{align*}
\bigintsss_\bxi  \left\langle \btheta_{B,\alpha} -\btheta_{B',\alpha}, \bxi - \nabla \psi(\tilde{\btheta}_{B,\alpha}) \right\rangle^2 & p_{\psi}(\bxi;2\btheta_{B,\alpha}-\btheta_{A,\alpha}) d\bxi \\
& = c_2 \| \btheta_{B,\alpha} - \btheta_{B',\alpha} \|_2^2 ~~\left( 1 + 2c_2 \| \btheta_{B,\alpha} - \btheta_{B',\alpha} \|_2^2 + 2c_2 \| \btheta_{B,\alpha} - \btheta_{A,\alpha} \|_2^2  \right)~.
\end{align*}
Similarly
\begin{align*}
\bigintsss_\bxi  \left\langle \btheta_{B,\alpha} -\btheta_{B',\alpha}, \bxi - \nabla \psi(\tilde{\btheta}_{B,\alpha}) \right\rangle^2 & p_{\psi}(\bxi;2\btheta_{B',\alpha}-\btheta_{A,\alpha}) d\bxi \\
& = c_2 \| \btheta_{B,\alpha} - \btheta_{B',\alpha} \|_2^2 ~~\left( 1 + 2c_2 \| \btheta_{B,\alpha} - \btheta_{B',\alpha} \|_2^2 + 2c_2 \| \btheta_{B',\alpha} - \btheta_{A,\alpha} \|_2^2  \right)~.
\end{align*}

Then, plugging into bound on $I_{A,B}$, we have
\begin{align*}
I_{A,B} & \leq c_2 \| \btheta_{B,\alpha} - \btheta_{B',\alpha} \|_2^2 \times 
\exp\left[ 2 c_2 \| \btheta_{B,\alpha} - \btheta_{B',\alpha} \|_2^2\right] \\
& ~~~~~ \times \max\big(  \exp \left[ 2 c_2 \| \btheta_{B,\alpha} - \btheta_{A,\alpha} \|_2^2 \right] 
 \times \left( 1 + 2c_2 \| \btheta_{B,\alpha} - \btheta_{B',\alpha} \|_2^2 + 2c_2 \| \btheta_{B,\alpha} - \btheta_{A,\alpha} \|_2^2  \right) \big. ~,\\
& ~~~~~~~~~~~~  \exp \left[ 2 c_2 \| \btheta_{B',\alpha} - \btheta_{A,\alpha} \|_2^2 \right] 
 \times \left( 1 + 2c_2 \| \btheta_{B,\alpha} - \btheta_{B',\alpha} \|_2^2 + 2c_2 \| \btheta_{B',\alpha} - \btheta_{A,\alpha} \|_2^2  \right) \big)~.
\end{align*}
Since $\alpha^2 \geq 8 c_2 \Delta^2(S_{n+1})$ where $\Delta(S_{n+1}) = \max_{z,z' \in S_{n+1}} \| \nabla \ell (\w,z) - \nabla \ell (\w,z') \|_2$, recalling that $\btheta_{B,\alpha} = \nabla \ell(\w,B)/\alpha$, we have 
\begin{align*}
2 c_2 \| \btheta_{B,\alpha} - \btheta_{B',\alpha} \|_2^2 & \leq \frac{1}{4}~, ~~~
2 c_2 \| \btheta_{B,\alpha} - \btheta_{A,\alpha} \|_2^2  \leq \frac{1}{4} ~, ~~~
2 c_2 \| \btheta_{B',\alpha} - \btheta_{A,\alpha} \|_2^2  \leq \frac{1}{4} ~.
\end{align*}
As a result, we have 
\begin{align*}
I_{A,B} & \leq c_2 \| \btheta_{B,\alpha} - \btheta_{B',\alpha} \|_2^2 \times \exp\left( \frac{1}{4}\right) \times \exp\left(\frac{1}{4}\right) \times \left( 1 +\frac{1}{4} + \frac{1}{4} \right) \\
& \leq \frac{5c_2}{2 \alpha^2} \| \nabla \ell(\w,S_B) - \nabla \ell(\w,S'_B) \|_2^2~.
\end{align*}
That completes the proof. \qed


\subsection{Expected Stability of EFLD}

\theoefldstab*

\proof 

Based on Lemma \ref{theo:stab}, we have
\begin{equation}
\hspace*{-5mm}
| \E_{S}[L_D(A(S)) - L_S(A(S))] | \leq c_1 \frac{b}{n} \E_{S} \E_{z'_n} \sqrt{ \sum_{t=1}^T  \underset{W_{0:(t-1)}}{\E} \underset{B_t \in G_1}{\E} \underset{A_t \in G_0}{\E} \left[ I_{A_t, B_t} \right] }~.
\end{equation}
with $I_{A_t, B_t} = \bigintsss_{\bxi_t} \frac{ \left(  dP_{B_t,\bxi_t} -   dP'_{B_t,\bxi_t} \right)^2 }{ dP_{A_t,\bxi_t}} d\bxi_t$. 

From Theorem \ref{theo:efld}, we have 
\begin{align*}
I_{A_{t}, B_{t}} & \leq \frac{5 c_{2}}{2 \alpha_{t \mid \mathbf{w}_{t-1}}^{2}}\left[\left\|\nabla \ell\left(\mathbf{w}_{t-1}, S_{B_{t}}\right)-\nabla \ell\left(\mathbf{w}_{t-1}, S_{B_{t}}^{\prime}\right)\right\|_{2}^{2}\right] \\
& =\frac{5 c_{2}}{2  \alpha_{t \mid \mathbf{w}_{t-1}}^{2}}\left[\left\| \frac{1}{b}  \sum_{z\in S'_{B_t}}\nabla \ell\left(\mathbf{w}_{t-1}, z \right)- \frac{1}{b}  \sum_{z\in S_{B_t}} \nabla \ell\left(\mathbf{w}_{t-1}, z \right)\right\|_{2}^{2}\right]  \\
& =\frac{5 c_{2}}{2 b \alpha_{t \mid \mathbf{w}_{t-1}}^{2}}\left[\left\| \nabla \ell\left(\mathbf{w}_{t-1}, z_n \right)-\nabla \ell\left(\mathbf{w}_{t-1}, z_n^\prime \right)\right\|_{2}^{2}\right]~,
\end{align*}
where the last equation holds because $S_{B_t}$ and $S'_{B_t}$ only differ at $z_n$ and $z_n^\prime$. 

Combining the above two inequalities, we have
\begin{align*}
| \E_{S}[L_D(A(S)) - L_S(A(S))] | \leq c_1 \frac{b}{n} \E_{S} \E_{z'_n} \sqrt{ \sum_{t=1}^T  \underset{W_{0:(t-1)}}{\E} \left[\left\| \nabla \ell\left(\mathbf{w}_{t-1}, z_n \right)-\nabla \ell\left(\mathbf{w}_{t-1}, z_n^\prime \right)\right\|_{2}^{2}\right] }~.
\end{align*}
That completes the proof. 
\qed

%% file: arXiv2022/app/signSGD_opt.tex
\subsection{Optimization Guarantees for Noisy Sign-SGD}

The ``density'' for a mini-batch $B$ at scale $\alpha$ is:
\begin{equation}
p_{\psi}(\bxi; \btheta_{B,\alpha}) = \exp(\langle \bxi, \btheta_{B,\alpha} \rangle - \psi(\btheta_{B,\alpha})) \ppi_0(\bxi)   ~, \qquad \btheta_{B,\alpha} \triangleq \frac{\btheta_B}{\alpha} = \frac{\nabla \ell(\w_{t}, S_{B})}{\alpha}~.
\end{equation}
Note that the corresponding expectation parameter
\begin{equation}
\m_{B,\alpha} = \nabla_{\btheta_{B,\alpha}} \psi( \theta_{B,\alpha} )~.
\end{equation}
The full-batch Noisy Sign-SGD update the parameters as 
\begin{equation}
\label{eq:app_signSGD_full_batch}
    \w_{t+1} = \w_{t} - \eta_t \bxi_t~, \qquad \bxi_{t,i} \sim \text{Rad}\left(\frac{1}{1+\exp \left(-2\nabla L_S(\w_t)/\alpha_{t}\right)}\right)~, ~ \forall i\in [d]~,
\end{equation}
where $\text{Rad}(x)$ is the parametric Rademacher distribution with density $x$ at $1$ and density $1-x$ at $-1$.
For mini-batch $B_t$ and scaling $\alpha_t$, mini-batch Noisy Sign-SGD updates the parameters as  
\begin{equation}\label{eq:app_signSGD_mini_batch}
    \w_{t+1} = \w_{t} - \eta_t \bxi_t~, \qquad \bxi_{t,i} \sim \text{Rad}\left(\frac{1}{1+\exp \left(-2\boldsymbol{\theta}_{B_{t}, \alpha_{t}}\right)}\right)~, ~ \forall i\in [d]~.
\end{equation}
We make the following smoothness assumptions of the empirical loss function $L_S(\w)$:
\asmpsmooth*
The assumption on on the empirical loss $L_S(\w)$ is common in optimization analysis, besides that, we also assume some natural statistical properties of the batch gradient of the loss $\nabla \ell(\w_{t}, S_{B_t})$, where the randomness comes from batches conditioned on $\w_t$, satisfies the following assumptions:
\asmpbatch*
With Assumption \ref{asmp: smooth} and Assumption \ref{asmp:batch}, we have the following guarantee for convergence of noisy signSGD under full batch and mini batch settings. The following is a restate theorem from the main paper for the mini-batch Noisy Sign-SGD
\begin{theo}
The following holds for any $S$, any initialization $\w_0$, and the expectation is taken over the randomness of algorithm: if the loss satisfies Assumption~\ref{asmp: smooth}, for full-batch noisy Sign-SGD with step size $\eta_t = 1/\sqrt{T}$ and $\alpha_t$ satisfying $c\geq \alpha_t \geq \|\nabla L_S(\w_t)\|_{\infty}$, we have
\vspace*{-3mm}
\begin{equation}
\label{eq:sign_SGD_full_batch_convergence_1}
    \E\left[\frac{1}{T}\sum_{t=1}^T \|\nabla L_S(\w_t)\|_2^2\right] \leq \frac{5c}{3\sqrt{T}}\left(L_S(\w_{0}) - L_S(\w^*) + \frac{1}{2}\|\vec{K}\|_1\right).
\vspace*{-3mm}
\end{equation}
Further, if Assumption~\ref{asmp:batch} holds, for mini-batch noisy Sign-SGD with step size $\eta_t = 1/\sqrt{T}$, and $\alpha_t$ satisfying $c \geq \alpha_t \geq \max[\sqrt{2}\kappa_t, 4\|\nabla L_S(\w_t)\|_{\infty}]$, we have
\vspace*{-3mm}
\begin{equation}
\label{eq:sign_SGD_mini_batch_convergence_1}
    \E\left[\frac{1}{T}\sum_{t=1}^T \|\nabla L_S(\w_t)\|_2^2  \right] \leq  \frac{4c}{\sqrt{T}}\left(L_S(\w_{0}) - L_S(\w^*) + \frac{1}{2}\|\vec{K}\|_1\right).
\vspace{-5mm}
\end{equation}
\end{theo}

\proof First we prove Equation \eqref{eq:sign_SGD_full_batch_convergence_1} for full-batched settings.
Conditioned at $t$-th iteration, with Assumption \ref{asmp: smooth}, we have 
\begin{align*}
    L_S(\w_{t+1}) -L_S(\w_{t}) & \leq \nabla L_S(\w_t)^T(\w_{t+1} - \w_{t}) + \frac{1}{2} \sum_{i=1}^d K_i (\w_{t+1,i} - \w_{t,i})^2\\
    & = - \eta_t  \nabla L_S(\w_t)^T \bxi_t  + \eta_t^2 \sum_{i=1}^d \frac{K_i}{2}.
\end{align*}
Then taking conditional expectation on both side for above equation we have
\begin{align*}
\mathbb{E}[ L_S(\w_{t+1}) -L_S(\w_{t})|\w_{t}] &\leq   - \eta_t  \nabla L_S(\w_t)^T \E_{\bxi_t|\w_t}\bxi_t  + \frac{\eta_t^2}{2} \|\vec{K}\|_1\\
&= - \eta_t  \sum_i \nabla L_S(\w_t)_i \left(\frac{2}{1+\exp \left(-2\nabla L_S(\w_t)_i/\alpha_t\right)}-1\right)  + \frac{\eta_t^2}{2} \|\vec{K}\|_1\\
&= - \eta_t  \sum_i \nabla L_S(\w_t)_i \left(\frac{\exp \left(2\nabla L_S(\w_t)_i/\alpha_t\right) -1 }{\exp \left(2\nabla L_S(\w_t)_i/\alpha_t\right)+1}\right)  + \frac{\eta_t^2}{2} \|\vec{K}\|_1\\
&= - \eta_t  \sum_i \nabla L_S(\w_t)_i \tanh (\nabla L_S(\w_t)_i/\alpha_t)  + \frac{\eta_t^2}{2} \|\vec{K}\|_1\\
&= - \eta_t  \sum_i |\nabla L_S(\w_t)_i| |\tanh (\nabla L_S(\w_t)_i/\alpha_t)|  + \frac{\eta_t^2}{2} \|\vec{K}\|_1.
\end{align*}
By taking $c\geq \alpha_t \geq \|\nabla L_S(\w_t)\|_{\infty}$, and $\eta_t=1/\sqrt{T}$, we have $\nabla L_S(\w_t)_i/\alpha_t\leq 1$ so we can apply Lemma \ref{lemm:tanh_bound} to have
\begin{align*}
\mathbb{E}[ L_S(\w_{t+1}) -L_S(\w_{t})|\w_{t}] &\leq  - \frac{e^2-1}{(e^2+1)\alpha_t\sqrt{T}}  \|\nabla L_S(\w_t)\|_2^2  + \frac{1}{2T} \|\vec{K}\|_1\\
&\leq - \frac{3}{5c\sqrt{T}}  \|\nabla L_S(\w_t)\|_2^2  + \frac{1}{2T} \|\vec{K}\|_1.
\end{align*}
By telescope sum we have
\begin{equation*}
    \E\left[\frac{1}{T}\sum_{t=1}^T \|\nabla L_S(\w_t)\|_2^2\right] \leq \frac{5c}{3\sqrt{T}}\left(L_S(\w_{0}) - L_S(\w^*) + \frac{1}{2}\|\vec{K}\|_1\right),
\end{equation*}
which completes the proof of full-batch updates.

Then we turn to prove Equation \eqref{eq:sign_SGD_mini_batch_convergence_1} for mini-batch settings.
From smoothness condition Assumption \ref{asmp: smooth}, we have
\begin{align*}
\mathbb{E}[ L_S(\w_{t+1}) -L_S(\w_{t})|\w_{t}] &\leq \E_{\w_{t+1}|\w_t}\nabla L_S(\w_t)^T(\w_{t+1} - \w_{t}) + \frac{1}{2} \E_{\w_{t+1}|\w_t} \sum_{i=1}^d L_i (\w_{t+1,i} - \w_{t,i})^2\\
&= \E_{\bxi_t|\w_t} \nabla L_S(\w_t)^T(-\eta_t \bxi_t) + \frac{1}{2} \E_{\bxi_t|\w_t} \sum_{i=1}^d L_i (-\eta_t \bxi_t)^2\\
    &=- \eta_t  \nabla L_S(\w_t)^T \E_{\bxi_t|\w_t}\bxi_t  + \frac{\eta_t^2}{2} \|\vec{K}\|_1\\
&= - \eta_t  \sum_i \nabla L_S(\w_t)_i \E_{B_t|\w_t}\left[\frac{2}{1+\exp \left(-2\nabla_i \ell(\w_t, S_{B_{t}}/\alpha_t)\right)}-1\right]  + \frac{\eta_t^2}{2} \|\vec{K}\|_1\\
&= - \eta_t  \sum_i \nabla L_S(\w_t)_i \E_{B_t|\w_t}\left[\frac{\exp \left(2\nabla_i \ell(\w_t, S_{B_{t}})/\alpha_t\right) -1 }{\exp \left(2\nabla_i \ell(\w_t, S_{B_{t}})/\alpha_t\right)+1}\right]  + \frac{\eta_t^2}{2} \|\vec{K}\|_1\\
& = - \eta_t  \sum_i \E_{B_t|\w_t} [\theta_{B_{t},i}] \E\left[\frac{\exp \left(2\theta_{B_{t},i}/\alpha_t\right) -1 }{\exp \left(2\theta_{B_{t},i}/\alpha_t\right)+1}\right]  + \frac{\eta_t^2}{2} \|\vec{K}\|_1.
\end{align*}

Focus on each individual term in the sum, we have
\begin{align*}
\E_{B_t|\w_t}&[\theta_{B,i}] \E_{B_t|\w_t}\left[\frac{\exp \left(2\theta_{B_t,i}/\alpha_t\right) -1 }{\exp \left(2\theta_{B_t,i}/\alpha_t\right)+1}\right]\\ 
& = \E_{B_t|\w_t}[\theta_{B_t,i}] \E_{B_t|\w_t}\left[\frac{\exp \left(2\theta_{B_t,i}/\alpha_t - 2\E_{B_t|\w_t}[\theta_{B_t,i}/\alpha_t] \right) -\exp - 2\E_{B_t|\w_t}\theta_{B_t,i}/\alpha_t }{\exp \left(2\theta_{B_t,i}/\alpha_t - 2\E_{B_t|\w_t}[\theta_{B_t,i}/\alpha_t]\right)+\exp - 2\E_{B_t|\w_t}\theta_{B_t,i}/\alpha_t}\right]
\end{align*}

For ease of notation, denote $2\theta_{B,i}/\alpha_t - \E_{B_t|\w_t}[2\theta_{B,i}/\alpha_t] = \theta,~\E_{B_t|\w_t}[2\theta_{B,i}/\alpha_t] =\mu$, 
and the pdf of $\theta$ is $p_{\theta}$ for the moment, then from Assumption \ref{asmp:batch}, we have $\theta$ is mean zero, symmetric around zero, and subgaussian with $\psi_2$ norm $2\kappa_t/\alpha_t$ by taking $\lambda=1,~v = \1_i$ in the sub-Gaussian assumption: $\E_\theta[\exp\theta] \leq \exp(2\kappa_t^2/\alpha_t^2)$.

Therefore, by changing notation we have 
\begin{align*}
     \E_{B_t|\w_t}[\theta_{B,i}] \E_{B_t|\w_t}\left[\frac{\exp \left(2\theta_{B_t,i}/\alpha_t\right) -1 }{\exp \left(2\theta_{B_t,i}/\alpha_t\right)+1}\right] = \frac{\alpha_t \mu}{2} \E_\theta\left[\frac{\exp \theta -\exp(-\mu) }{\exp \theta +\exp(-\mu)}\right].
\end{align*}
By symmetry of the distribution of $\theta$, we have
\begin{align*}
\frac{\alpha_t \mu}{2} \mu\E&_\theta\left[\frac{\exp \theta -\exp(-\mu) }{\exp \theta +\exp(-\mu)}\right]\\
    &= \frac{\alpha_t \mu}{2}\int_{-\infty}^{\infty}p_\theta(x)\left[\frac{\exp x-\exp(-\mu) }{\exp x +\exp(-\mu)}\right]dx\\
    &=\frac{\alpha_t \mu}{2}\int_{0}^{\infty}p_\theta(x)\left[\frac{\exp x -\exp(-\mu) }{\exp x +\exp(-\mu)}+\frac{\exp (-x) -\exp(-\mu) }{\exp (-x) +\exp(-\mu)}\right]dx\\
    &=\frac{\alpha_t \mu}{2}\int_{0}^{\infty}p_\theta(x)\frac{(\exp x -\exp(-\mu))(\exp (-x) +\exp(-\mu)) + (\exp (-x) -\exp(-\mu))(\exp x +\exp(-\mu))}{(\exp x +\exp(-\mu))(\exp (-x) +\exp(-\mu))}dx\\
    &=\alpha_t/2 \int_{0}^{\infty}p_\theta(x)\frac{2\mu (1-\exp(-2\mu))}{(\exp x +\exp(-\mu))(\exp (-x) +\exp(-\mu))}dx
\end{align*}
By symmetry, we have $p_\theta(x) = p_\theta(-x)$, and therefore
\begin{align*}
\alpha_t/2 & \int_{0}^{\infty}p_\theta(x)\frac{2\mu (1-\exp(-2\mu))}{(\exp x +\exp(-\mu))(\exp (-x) +\exp(-\mu))}dx\\
    &=\alpha_t/2\int_{-\infty}^{\infty}p_\theta(x)\frac{\mu (1-\exp(-2\mu))}{(\exp x +\exp(-\mu))(\exp (-x) +\exp(-\mu))}dx\\
    &=\frac{\alpha_t }{2}(\mu (1-\exp(-2\mu))) \E_\theta\frac{1}{(\exp \theta +\exp(-\mu))(\exp (-\theta) +\exp(-\mu))}
\end{align*}
Since $\frac{\alpha_t }{2}(\mu (1-\exp(-2\mu)))\geq 0$, and $\frac{1}{x}$ is convex on $\R_+$, we have
\begin{align*}
   & \frac{\alpha_t }{2}(\mu (1-\exp(-2\mu))) \E_\theta\frac{1}{(\exp \theta +\exp(-\mu))(\exp (-\theta) +\exp(-\mu))}\\ \geq &\frac{\alpha_t }{2}(\mu (1-\exp(-2\mu))) \frac{1}{\E_\theta(\exp \theta +\exp(-\mu))(\exp (-\theta) +\exp(-\mu))}.
\end{align*}

Using the sub-Gaussian property of $\theta$: $\E_\theta[\exp\theta] \leq \exp(2\kappa_t^2/\alpha_t^2)$, and symmetry so $\E_\theta[\exp -\theta] \leq \exp(2\kappa_t^2/\alpha_t^2)$, we have
\begin{align*}
    &\frac{\alpha_t }{2}(\mu (1-\exp(-2\mu))) \frac{1}{\E_\theta(\exp \theta +\exp(-\mu))(\exp (-\theta) +\exp(-\mu))}\\
    &= \frac{\alpha_t }{2}(\mu (1-\exp(-2\mu))) \frac{1}{1 +\exp(-2\mu)+\exp(-\mu)(\E\exp \theta +\E\exp (-\theta))}\\
    &\geq  \frac{\alpha_t }{2}(\mu (1-\exp(-2\mu))) \frac{1}{1 +\exp(-2\mu)+2\exp(-\mu)\exp(2\kappa_t^2/\alpha_t^2)}\\
    &\geq  \frac{\alpha_t }{2}(\mu (1-\exp(-2\mu))) \frac{1}{2+2\exp(2\kappa_t^2/\alpha_t^2)}.
\end{align*}
Switching back to our previous notation:
\begin{align*}
    \frac{\alpha_t }{2}(\mu (1-\exp(-2\mu))) \frac{1}{2+2\exp(2\kappa_t^2/\alpha_t^2)}
  & = \E_{B_t|\w_t} \theta_{B,i} (1-\exp(-4\E_{B_t|\w_t}\theta_{B,i}/\alpha_t)) \frac{1}{2+2\exp(2\kappa_t^2/\alpha_t^2)}\\
  & = \nabla L_S(\w_t)_i (1-\exp(-4\nabla L_S(\w_t)_i/\alpha_t)) \frac{1}{2+2\exp(2\kappa_t^2/\alpha_t^2)},
\end{align*}

which implies we have
\begin{align*}
    \mathbb{E}[ L_S(\w_{t+1}) -L_S(\w_{t})|\w_{t}] \leq - \frac{\eta_t}{2+2\exp(2\kappa_t^2/\alpha_t^2)}  \sum_i \nabla L_S(\w_t)_i (1-\exp(-4\nabla L_S(\w_t)_i/\alpha_t))  + \frac{\eta_t^2}{2} \|\vec{K}\|_1.
\end{align*}
Using Lemma \ref{lemm:mix_norm}, we have
\begin{equation}
    \mathbb{E}[ L_S(\w_{t+1}) -L_S(\w_{t})|\w_{t}] \leq - \frac{\eta_t}{2(1+\exp(2\kappa_t^2/\alpha_t^2))}  \sum_i |\nabla L_S(\w_t)_i| \min \bigg[ |2\nabla L_S(\w_t)_i/\alpha_t| ,0.5\bigg]  + \frac{\eta_t^2}{2} \|\vec{K}\|_1.
 \end{equation}

We choose $\alpha_t$ such that $\alpha_t \geq 4\|\nabla L_S(\w_t)\|_{\infty}$, then we have
\begin{equation}
\mathbb{E}[ L_S(\w_{t+1}) -L_S(\w_{t})|\w_{t}] \leq - \frac{\eta_t}{\alpha_t (1+\exp(2\kappa_t^2/\alpha_t^2))}  \|\nabla L_S(\w_t)\|_2^2  + \frac{\eta_t^2}{2} \|\vec{K}\|_1.
\end{equation}
We choose $\alpha_t$ such that $c\geq \alpha_t \geq \max[\sqrt{2} \kappa_t, 4\|\nabla L_S(\w_t)\|_{\infty}]$, then we have 
\begin{equation}
    \mathbb{E}[ L_S(\w_{t+1}) -L_S(\w_{t})|\w_{t}] \leq - \frac{\eta_t}{c (1+e)}  \|\nabla L_S(\w_t)\|_2^2  + \frac{\eta_t^2}{2} \|\vec{K}\|_1.
\end{equation}
Therefore, if we choose $\eta_t = 1/\sqrt{T}$, we have
\begin{equation}
    \mathbb{E}[ L_S(\w_{t+1}) -L_S(\w_{t})|\w_{t}] \leq - \frac{1}{c (1+e)\sqrt{T}}  \|\nabla L_S(\w_t)\|_2^2  + \frac{1}{2T} \|\vec{K}\|_1.
\end{equation}
By telescope sum, we have
\begin{equation}
    \E\left[\frac{1}{T}\sum_{t=1}^T \|\nabla L_S(\w_t)\|_2^2  \right] \leq  \frac{(1+e)c}{\sqrt{T}}\left(L_S(\w_{0}) - L_S(\w^*) + \frac{1}{2}\|\vec{K}\|_1\right).
\end{equation}
With $\w_R$ to be uniformly randomly sampled from $\{\w_1, .., \w_T\}$, we have
\begin{equation}
    \E \|\nabla L_S(\w_R)\|_2^2\leq  \frac{(1+e)c}{\sqrt{T}}\left(L_S(\w_{0}) - L_S(\w^*) + \frac{1}{2}\|\vec{K}\|_1\right),
\end{equation}
and note that $1+e<4 $ which completes the proof.   \qed

\begin{lemm}\label{lemm:tanh_bound}
For any $-1\leq x\leq 1$, the following holds:
\begin{equation}
    |\tanh x|\geq \frac{e^2-1}{e^2+1}|x|
\end{equation}
\end{lemm}
\proof Without loss of generality, we focus on $0<x\leq 1$. we prove $\tanh x/ x$ is decreasing function on $\R^+$, which is equivalent to
\begin{align*}
    (\frac{\tanh x}{x})' =  \frac{x(1-\tanh^2 x) - \tanh x }{x^2}\leq 0,
\end{align*}
and is equivalent to 
\begin{align*}
    x\leq \frac{\tanh x}{1 -\tanh^2 x},
\end{align*}
where the right hand side is $\sinh x$ and use the fact that $x\leq \sinh x$ for $x > 0$ implies $\tanh x/x$ is decreasing function, so 
\begin{align*}
    \tanh x/x\geq \tanh 1 = \frac{e^2-1}{e^2 +1},
\end{align*} which completes the proof. \qed

\begin{lemm}\label{lemm:mix_norm} For any $x$,
$$|1-\exp(-2x))|\geq \min (|x|,\frac{1}{2}).$$
\end{lemm}
\proof 
Since $\exp(-x)\geq 1-x$, so we have for $x<0$:
$$1-\exp (-2x)\leq 2x\leq x.$$

For $x>0$, since $\exp x \geq 1+x$, we have $\exp -x\leq \frac{1}{1+x}$, so
$$1-\exp (-2x)\geq \frac{2x}{1+2x}.$$
Then when $\frac{1}{2}>x>0$, $$\frac{2x}{1+2x} > x,$$
and when $x\geq \frac{1}{2}$, $$\frac{2x}{1+2x} \geq \frac{1}{2},$$
which completes the proof.\qed

\theoSGLDopt*

\proof Recall that the update of SGLD is
\begin{equation}
\mathbf{w}_{t+1}=\mathbf{w}_{t}-\eta_{t} \nabla \ell\left(\mathbf{w}_{t}, S_{B_{t}}\right)+ \sigma_{t}  \mathcal{N}\left(0, \mathbb{I}_{d}\right)~.
\end{equation}
By smoothness of the loss, taking expectation w.r.t.~the randomness of the mini-batch $B_t$ and the Gaussian draw $\g_t \sim \cN(0,\I_d)$ conditioned on $\w_{1:t}$, we have
\begin{align*}
\mathbb{E} [L_S(\w_{t+1})] &\leq L_S(\w_t) + \mathbb{E} [\left\langle \nabla L_S(\w_t), \w_{t+1} - \w_t \right\rangle ]  + \frac{K}{2} \mathbb{E} [ \|  \w_{t+1} - \w_t\|^2] \\
& =  L_S(\w_t) + \mathbb{E} [\left\langle \nabla L_S(\w_t), -\eta_t \nabla \ell\left(\mathbf{w}_{t}, S_{B_{t}}\right) + \sigma_t \g_t \right\rangle] + \frac{K}{2} \E[\| -\eta_t \nabla \ell\left(\mathbf{w}_{t}, S_{B_{t}}\right) + \sigma_t \g_t \|^2] \\
& \overset{(a)}{\leq}  L_S(\w_t) -\eta_t \|\nabla L_S(\w_t)\|^2 + \frac{K \eta_t^2}{2} \left(c_3 \kappa_t^2 + p\sigma_t^2/\eta_t^2\right)~,
\end{align*}
where (a) follows since $\E[\nabla \ell\left(\mathbf{w}_{t}, S_{B_{t}}\right)]=0, \E[\g]=0, \E[\|\g\|^2] = p$, and $\E[\|\nabla \ell\left(\mathbf{w}_{t}, S_{B_{t}}\right)\|^2] \leq c_3 \kappa_t^2$ for some absolute constant $c_3$ since $\nabla \ell\left(\mathbf{w}_{t}, S_{B_{t}}\right)$ is sub-Gaussian with $\psi_2$-norm $\kappa_t$.

Rearranging the above inequality and using $\alpha_t = \sigma_t/\eta_t$ we have
\begin{align} \label{eq:one_step_descend}
    \eta_t \| \nabla L_S(\w_t)\|^2 \leq  L_S(\w_t) - \mathbb{E} [L_S(\w_{t+1})]  + \frac{K \eta_k^2}{2}\left(c_3 \kappa_t^2 + p\alpha_t^2\right)~.
\end{align}
Summing over $t =1$ to $t =T$ and apply expectation over the trajectory at each step, we have
\begin{align*}
\sum_{t =1}^T \eta_t\mathbb{E}\| \nabla L_S(\w_t)\|^2 
\leq L_S(\w_1) - L_S(\w^\star) +  \sum_{t=1}^T \frac{K \eta_t^2}{2} (p \alpha_t^2 + \kappa_t^2)~,
\end{align*}
where $\w^*$ is a minima of $L_S(\w)$. With $\eta_t= \frac{1}{\sqrt{T}}$ for all $t \in [T]$, we have
\begin{align*}
\frac{1}{T} \sum_{t =1}^T \mathbb{E} \| \nabla L_S(\w_t)\|^2 
\leq\frac{L_S(\w_1) - L_S(\w^\star)}{\sqrt{T}} + \frac{  \frac{K}{2T} \sum_{t=1}^T (p \alpha_t^2 +  c_3 \kappa_t^2) }{\sqrt{T}}~.  
\end{align*}
With $\w_R$ to be uniformly randomly sampled from $\{\w_1, .., \w_T\}$, we have

\begin{align*}
\mathbb{E} \| \nabla L_S(\w_R)\|^2 
\leq\frac{L_S(\w_1) - L_S(\w^\star)}{\sqrt{T}} + \frac{  \frac{K}{2T} \sum_{t=1}^T (p \alpha_t^2 +  c_3 \kappa_t^2) }{\sqrt{T}}~.  
\end{align*}

That completes the proof.  \qed 

%% file: arXiv2022/app/app_exp.tex
\subsection{Datasets} 
We use MNIST \cite{mnist}, Fashion-MNIST \cite{fashion-mnist}, CIFAR-10 \cite{cifar10} and CIFAR-100 \cite{cifar10} in our experiments.


\noindent {\bf MNIST dataset:} 60,000 black and white training images, including handwritten digits 0 to 9. We use a subset of MNIST with $n = 10,000$ data points where 1,000 samples from each class are randomly selected. Each image of size $28 \times 28$ is first re-scaled into [0,1] by dividing each pixel value by 255, then z-scored by subtracting the mean and dividing the standard deviation of the training set.

\noindent {\bf Fashion-MNIST dataset:} 60,000 gray-scale training images and 10,000 test images, including 10 clothing categories such as shirts, dresses, sandals, etc. Each image of size $28 \times 28$ is first re-scaled into [0,1] by dividing each pixel value by 255, then z-scored by subtracting the mean and dividing the standard deviation of the training set.

\noindent {\bf CIFAR-10/-100 dataset:} 60,000 color images consisting of 10/100 categories, e.g., airplane, cat, dog etc. The training set includes 50,000 images while the test set contains the rest 10,000 images. Each image of size $32 \times 32$ has 3 color channels. We first re-scale each image into [0, 1] by dividing each pixel value by 255, then each image is normalized by subtracting the mean and dividing the standard deviation of the training set for each color channel. We also use \textit{RandomCrop} and \textit{RandomHorizontalFlip} for data augmentation.

\subsection{Network Architectures} 
For experiments on both MNIST and Fashion-MNIST, we use a convolutional neural network with two convolutional layers followed by two fully connected layers with ReLU activations. For experiments use the CIFAR-10 dataset, we consider CNN architecture with two convolutional layers
and three fully connected layers. The detail of each CNN architecture can be found in Table \ref{tab:cnn1} and Table~\ref{tab:cnn2}.

\begin{table}[h]
\caption{CNN architecture for MNIST and Fashion MNIST.}
\centering
\begin{tabular}{cc} 
Layer & Parameters \\
\hline \hline Convolution & 32 filters of $5 \times 5$ \\
Max-Pooling & $2 \times 2$ \\
Convolution & 64 filters of $5 \times 5$\\
Max-Pooling & $2 \times 2$ \\
Fully connected & 1024 units \\
Softmax & 10 units \\
\hline
\end{tabular}
    \label{tab:cnn1}
\end{table}

\begin{table}[h]
\caption{CNN architecture for CIFAR-10.}
\centering
\begin{tabular}{cc} 
Layer & Parameters \\
\hline \hline Convolution & 64 filters of $5 \times 5$ \\
Max-Pooling & $2 \times 2$ \\
Convolution & 192 filters of $5 \times 5$\\
Max-Pooling & $2 \times 2$ \\
Fully connected & 384 units \\
Fully connected & 192 units \\
Softmax & 10 units \\
\hline
\end{tabular}
    \label{tab:cnn2}
\end{table}


\subsection{Experimental Setup}  
We are interested in stochastic gradient Langevin dynamics,
whose iterative updates are given by $\w_{t} = \w_{t-1} - \eta_t \nabla \ell(\w_{t-1}, S_{B_t}) + \mathcal{N}\left(0, \sigma_t^2 \mathbb{I}_{d}\right).$ 
We also denote $\beta_t =2\eta_t/\sigma_t^2$ as the inverse temperature at time t. 
For MNIST and Fashion-MNIST, the initial learning rate is $\eta_0=0.004$ and it decays by 0.96 after every 5 epochs. For CIFAR-10,the initial learning rate is $\eta_0=0.005$ and it decays by 0.995 after every 5 epochs. We use batch size $|B_t| = 100$ for MNIST and Fashion-MNIST, and $|B_t| = 200$ for CIFAR-10. 

Motivated by \cite{zhang2017understanding}, we train CNN with SGLD on a smaller subset of MNIST dataset ($n=10000$) with randomly corrupted labels. The corruption fraction varies from $0\%$ (without label corruption) to $60\%$. For different level of randomness, we use the same training setting with batch size $|B_t|=100$, initial step size $\eta_0 = 0.005$, noise variance $\sigma_t = 0.2 \cdot \eta_t$, and we decay $\eta_t$ by 0.995 for every 30 epochs.

We are also interested in Noisy Sign-SGD whose iterative updates are given by 
$\w_{t+1} = \w_{t} - \eta_t \bxi_t$, where $\bxi_{t,j} \sim p_{\btheta_{B_t,\alpha_t,j}}(\xi_j)=\frac{\exp(\xi_j \btheta_{B_t,\alpha_t})}{\exp(-\btheta_{B_t,\alpha_t})+\exp(\btheta_{B_t,\alpha_t})}$. The initial learning rate is $\eta_0=10^{-4}$ and it decays by 0.1 after every 30 epochs. We use batch size $|B_t| = 100$ for all benchmarks.


All experiments minimize cross-entropy loss for a fixed number of epochs and have been run on NVIDIA Tesla K40m GPUs. For CNN, we repeat each experiment 30 times, and for ResNet-18, we repeat 5 times.

\subsection{Evaluation of the bound in Theorem~\ref{theo:efldstab}} 

{\bf Estimation of $c_0$ in Proposition 1.} Something similar to $c_0$ exists in all prior bounds, e.g., see Lemma 1 of \citet{xu2017information}, Theorem 9 in \citet{Li2020On}, etc. If the loss is bounded, then one can get $c_0$, but in general, it is difficult. Empirically, we chose $c_0$ based on maximum observed training loss for our bound as well as baseline approaches. 

{\bf Computation of $\Delta_t^2( \bar{S}_{n+1} )$ and the expectation over $S_{n+1}$ in Theorem~\ref{theo:efldstab}.}
We start with the uniform analysis. Let $L$ be the Lipschitz constant so $\| \nabla \ell(w,z)\|_2 \leq L$. Then $\Delta_{t|}(\bar{S}_{n+1}) \leq 2L$. So, suffices to have $\alpha_t = 4\sqrt{2} c_2 L$.  This will always work if $L$ is known or ensured using gradient clipping. 
%

More generally, for any given $\bar{S}_{n+1}$, $\Delta_{t|}(\bar{S}_{n+1})$ can be computed by definition
from Theorem~\ref{theo:efld} by taking the maximum discrepancy over  all $\binom{n+1}{2}$ pairs of points in $\bar{S}_{n+1}$. The argmax for the discrepancy will arguably have $z \in S_n$ and $z'=z_{n+1}$, which simplifies the argmax to just consider $n$ pairs rather than $\binom{n+1}{2}$. We verified and used this simpler calculation for the experiments.

\begin{figure}[t]
    \centering
      \subfigure[Estimates of $\Delta_t^2( \bar{S}_{n+1} )$]{\includegraphics[width=0.49\textwidth]{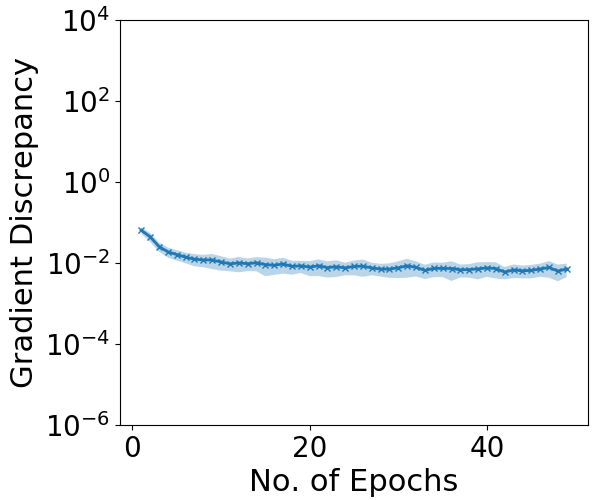}}  
      \subfigure[Effect of batch size]{\includegraphics[width=0.49\textwidth]{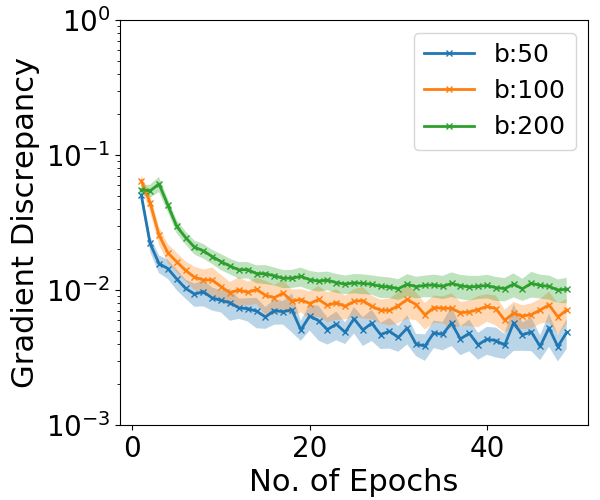}}
    \vspace{-3mm}
    \caption[]{Results for training CNN using SGLD on MNIST. (a) Estimation of $\Delta_t^2( \bar{S}_{n+1} )$ using re-sampled $(S_n,z_{n+1})$. 
    The variance of such estimation is small. (b) The gradient discrepancy for different batch sizes. Overall, the differences are not significant. }
    \label{fig:batch_size}
    \vspace{-4mm}
\end{figure}

The expectation over $\bar{S}_{n+1}$ is hard to compute but (re)sampling based estimates can be used. The term entails sampling $\bar{S}_{n+1}$, running the training with $n$ samples $S_n$, using $z_{n+1}$ to compute the discrepancy. 
To get Monte Carlo estimates, the analysis has to be repeated for different $z_{n+1}$ (done in the main paper) or by resampling $(S_n,z_{n+1})$, and computing the average which we show in Figure \ref{fig:batch_size} (a). The estimates are shown to have rather low variance.  

{\bf Dependence of the bound on batchsize $b$.} Our bound in Theorem~\ref{theo:efldstab} does not have a dependence on $b$, because of what Remark 3.2 shows, i.e., the $1/b$ scale factor neutralizes the leading b term in Lemma 1. The gradient discrepancy term itself may have a mild dependence on $b$, with smaller batch sizes having mildly smaller empirical gradient discrepancy as shown in Figure~\ref{fig:batch_size} (b).

\begin{table}[h]
\caption{Details of Experiments reported in Figure~\ref{fig:sgld_bound} and \ref{fig:sgld_mnist} for MNIST and Fashion-MNIST with CNN}
\centering
\begin{tabular}{c|c} 
\hline
Parameter & Values \\
\hline
Dataset & MNIST/ Fashion-MNIST \\\hline
Architecture & CNN with 2 conv. layers and 2 linear layers \\\hline
Batch Size & $100$\\\hline
Learning Rate & $\eta_0=4\times10^{-3}$, decay epochs=5, decay rate=0.96 \\\hline
Inverse Temperature& $\beta \in [5000, 55000]$ \\\hline
Number of Epochs & 50\\\hline
No. of training examples & 55000 \\\hline
Number of Repeated Runs & 30 \\
\hline
\end{tabular}
\label{tab:mnist}
\end{table}

\begin{table}[h]
\caption{Details of Experiments reported in Figure~\ref{fig:sgld_bound} and \ref{fig:sgld_cifar10} for CIFAR-10 with CNN}
\centering
\begin{tabular}{c|c} 
\hline
Parameter & Values \\
\hline
Dataset & CIFAR-10\\\hline
Architecture & CNN with 2 conv. layers and 3 linear layers \\\hline
Batch Size & $200$\\\hline
Learning Rate & $\eta_0=5\times10^{-3}$, decay epochs=5, decay rate=0.995 \\\hline
Inverse Temperature& $\beta \in [5000, 55000]$ \\\hline
Number of Epochs & 1000\\\hline
No. of training examples & 55000 \\\hline
Number of Repeated Runs & 20 \\
\hline
\end{tabular}
\label{tab:cifar10}
\end{table}

\begin{table}[h]
\caption{Details of Experiments reported in Figure~\ref{fig:bound_random} for MNIST with CNN ($\sigma_t = 0.2 \cdot \eta_t$)}
\centering
\begin{tabular}{c|c} 
\hline
Parameter & Values \\
\hline
Dataset & MNIST\\\hline
Architecture & CNN with 2 conv. layers and 2 linear layers \\\hline
Batch Size & $100$\\\hline
Learning Rate & $\eta_0=5\times10^{-3}$, decay epochs=30, decay rate=0.995 \\\hline
Number of Epochs & 1000\\\hline
No. of training examples & 10000 \\\hline
Number of Repeated Runs & 30 \\
\hline
\end{tabular}
\label{tab:random}
\end{table}

\begin{figure*}[t] 
\centering
 \subfigure[MNIST, $\alpha_t^2 \approx 0.1$, $\beta_t=55000$]{
 \includegraphics[width=0.4\textwidth]{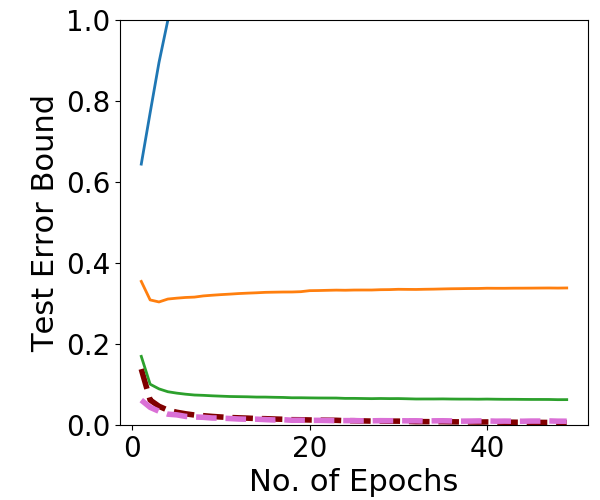}
 }
  \subfigure[MNIST, $\alpha_t^2 \approx 0.1$, $\beta_t=5000$]{
 \includegraphics[width=0.4\textwidth]{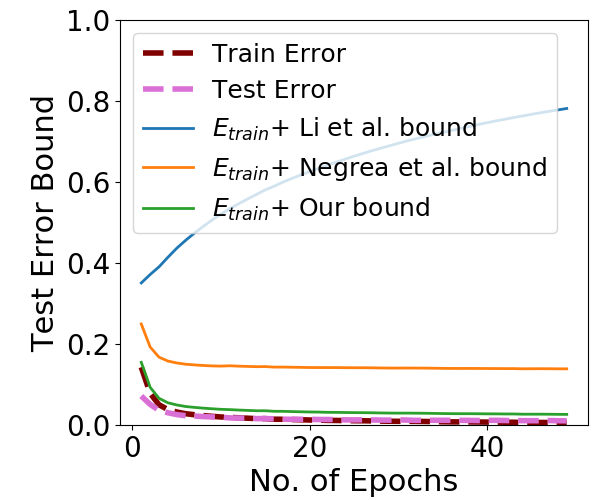}
 }
\subfigure[MNIST, $\alpha_t^2 \approx 0.1$, $\beta_t=55000$]{
\includegraphics[width=0.4\textwidth]{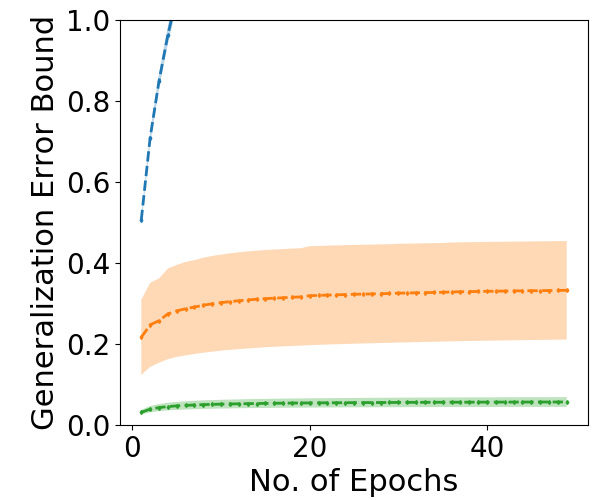}
 }
\subfigure[MNIST, $\alpha_t^2 \approx 0.1$, $\beta_t=5000$]{
\includegraphics[width=0.4\textwidth]{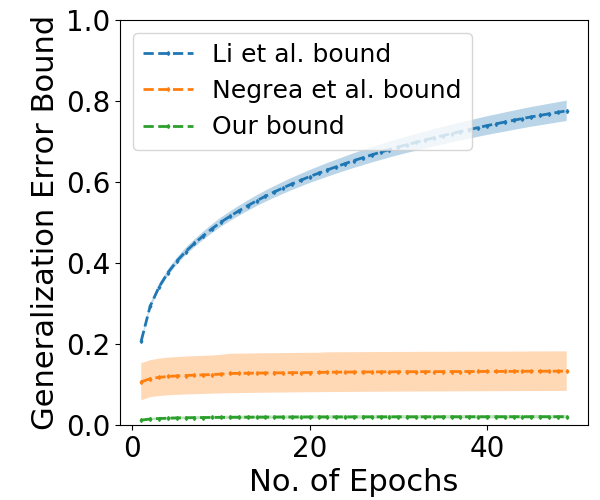}
 }   
 \subfigure[MNIST, $\alpha_t^2 \approx 0.1$, $\beta_t=55000$]{
 \includegraphics[width=0.4\textwidth]{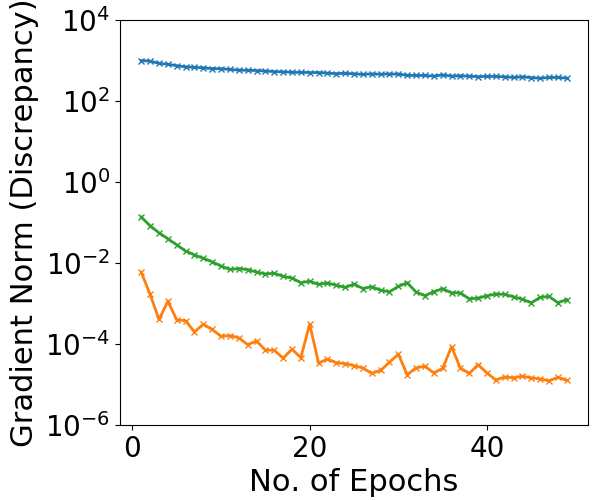}
 }
 \subfigure[MNIST, $\alpha_t^2 \approx 0.1$, $\beta_t=5000$]{
 \includegraphics[width=0.4\textwidth]{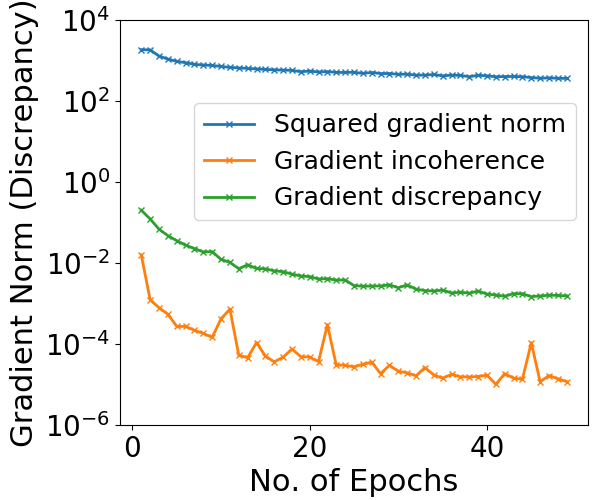}
 } 
\caption[]{Numerical results for CNN trained on MNIST using SGLD 
with a small noise variance $\sigma_t\approx 10^{-4}$.
}
\label{fig:sgld_mnist}
\end{figure*}

\begin{figure*}[t] 
\centering
  \subfigure[CIFAR-10, $\alpha_t^2 \approx 0.01$, $\beta_t=55000$]{
 \includegraphics[width=0.4\textwidth]{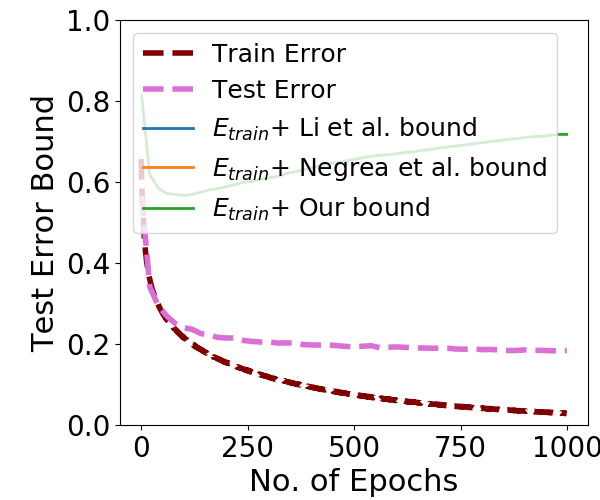}
 }
\subfigure[CIFAR-10, $\alpha_t^2 \approx 0.1$, $\beta_t=5000$]{
 \includegraphics[width=0.4\textwidth]{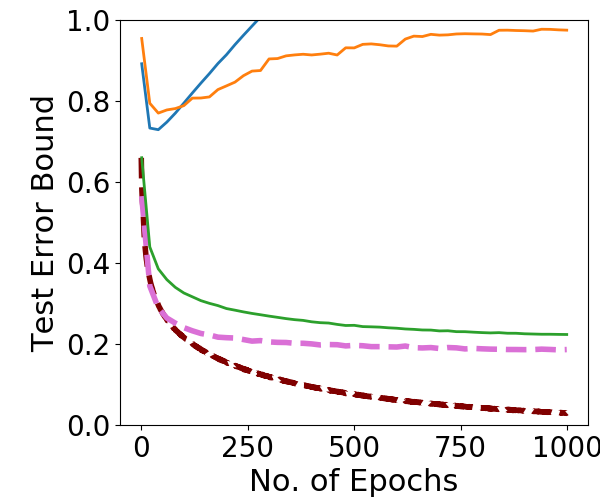}
 } 
 \subfigure[CIFAR-10, $\alpha_t^2 \approx 0.01$, $\beta_t=55000$]{
 \includegraphics[width=0.4\textwidth]{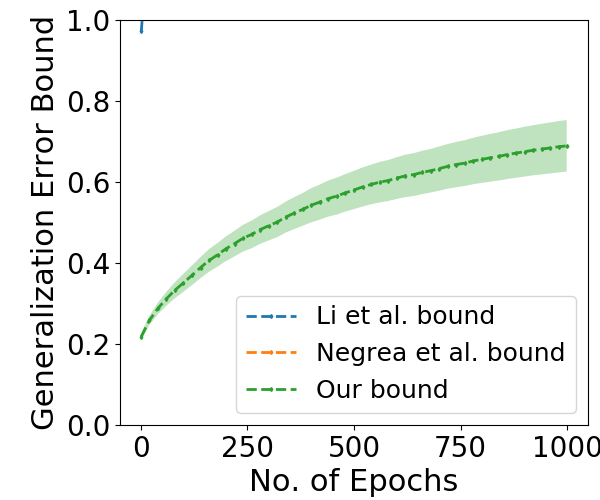}
 } 
  \subfigure[CIFAR-10, $\alpha_t^2 \approx 0.1$, $\beta_t=5000$]{
 \includegraphics[width=0.4\textwidth]{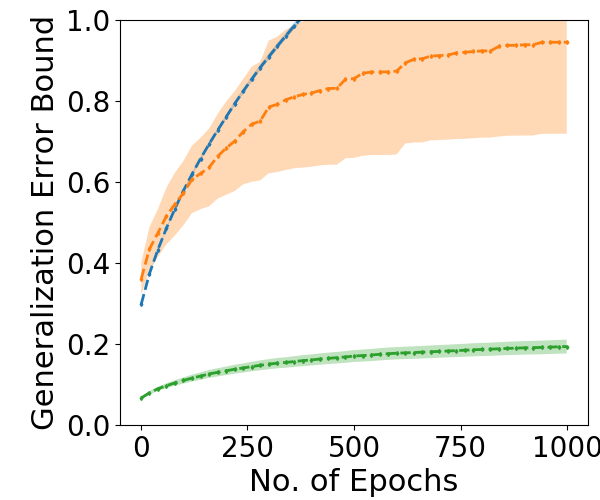}
 }
  \subfigure[CIFAR-10, $\alpha_t^2 \approx 0.01$, $\beta_t=55000$]{
 \includegraphics[width=0.4\textwidth]{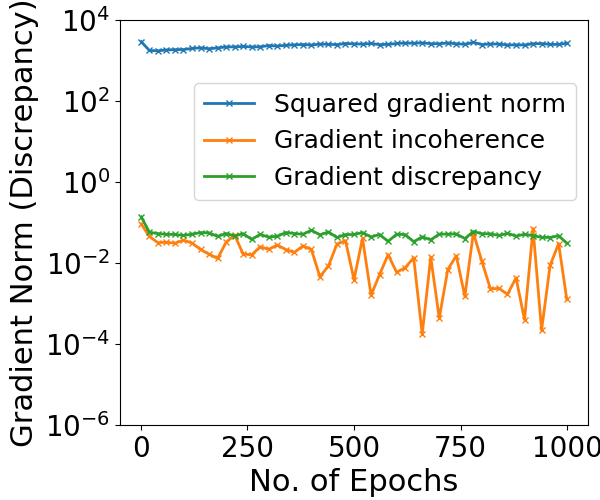}
 }
  \subfigure[CIFAR-10, $\alpha_t^2 \approx 0.1$, $\beta_t=5000$]{
 \includegraphics[width=0.4\textwidth]{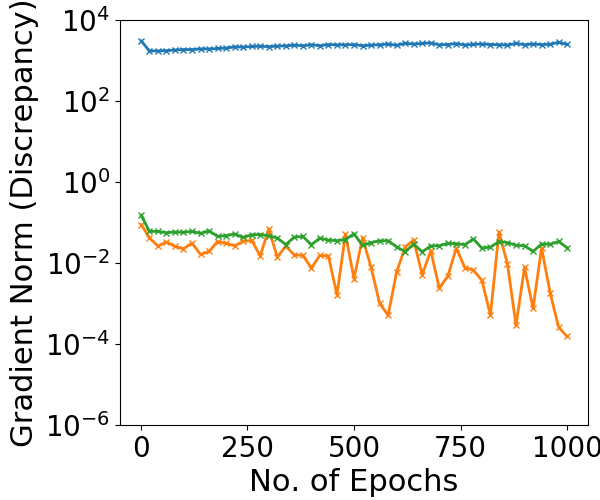}
 } 

\caption[]{Numerical results for training CNN using SGLD 
on CIFAR-10. 
}
\label{fig:sgld_cifar10}
\end{figure*}

\begin{figure*}[th] 
\centering
  \subfigure[CNN, MNIST, Noisy-SGD, $\alpha_t = 0.2$, $\sigma_t\approx 10^{-4}$]{
 \includegraphics[width=0.85\textwidth]{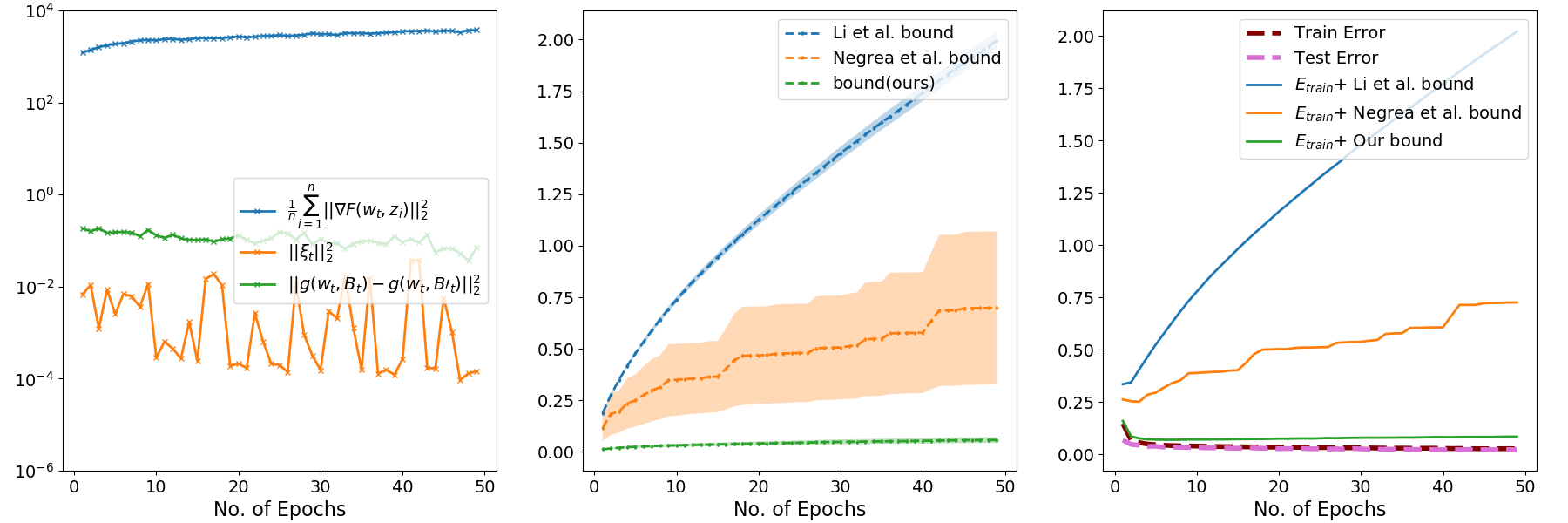}
 } 
  \subfigure[CNN, MNIST, Noisy-SGD, $\alpha_t = 0.002$, $\sigma_t\approx 10^{-6}$]{
 \includegraphics[width=0.85\textwidth]{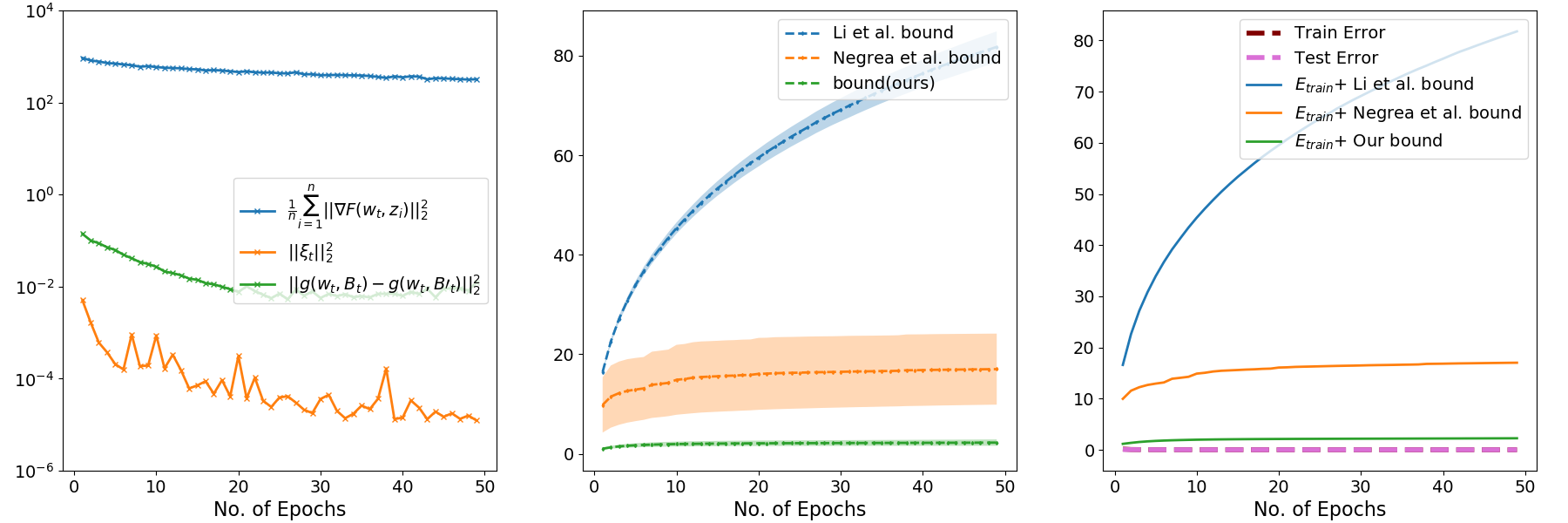}
 }
\caption[]{Numerical results for CNN trained on MNIST using Noisy SGD. 
We follow the setting described in \cite{Li2020On}. The initial $\eta_t$ is $4\times 10^{-3}$, and decays by 0.95 ($\alpha_t=0.2$) or 0.995 ($\alpha_t=0.002$) for every 5 epochs. 
}
\label{fig:mnist_noisysgd}
\end{figure*}

\begin{figure*}[th!] 
\centering
  \subfigure[CNN, MNIST]{
 \includegraphics[width=0.4\textwidth]{ICLR2022/images/cnn1-mnist-n0-bs100-sign-sgd-train-dynamics-reduced.png}
 } 
  \subfigure[CNN, Fashion]{
 \includegraphics[width=0.4\textwidth]{ICLR2022/images/cnn1-fashion-n0-bs100-sign-sgd-train-dynamics-reduced.png}
 }
 \subfigure[ResNet-18, CIFAR10]{
 \includegraphics[width=0.4\textwidth]{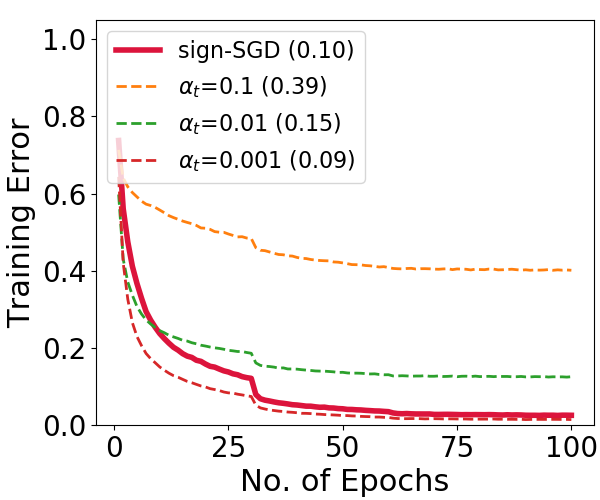}
 }
 \subfigure[ResNet-18, CIFAR100]{
 \includegraphics[width=0.4\textwidth]{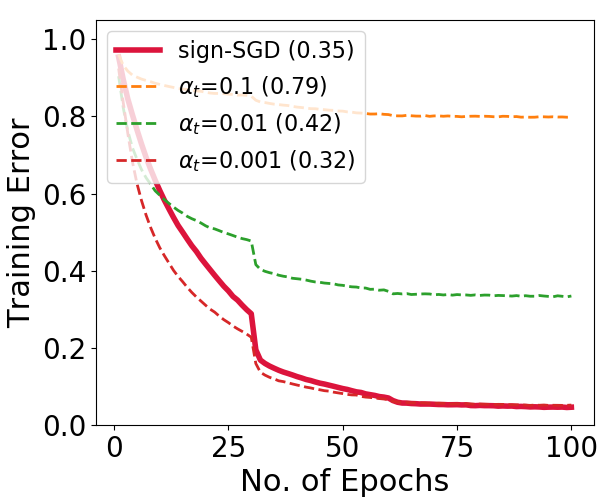}
 }
\caption[]{Training dynamics of CNN on MNIST and Fashion-MNIST, and ResNet-18 on CIFAR-10 and CIFAR-100 using noisy sign-SGD with different scaling $\alpha_t$. Legends indicate the choice of $\alpha_t$ and the numbers in brackets are test errors at convergence. As $\alpha_t\to 0$, Nosiy sign-SGD matches both the optimization trajectory as well as the final test accuracy of the original sign-SGD \citep{bernstein2018signsgd}.} 
\label{fig:noisy_signsgd_train}
\end{figure*}

\begin{figure*}[t] 
\centering
 \subfigure[MNIST, $\alpha_t = 1.0$]{
 \includegraphics[width=0.3\textwidth]{ICLR2022/images/smallcnn-mnist-bounds-noisy-signSGD-c.png}
 }
 \subfigure[MNIST, $\alpha_t = 0.1$]{
 \includegraphics[width=0.3\textwidth]{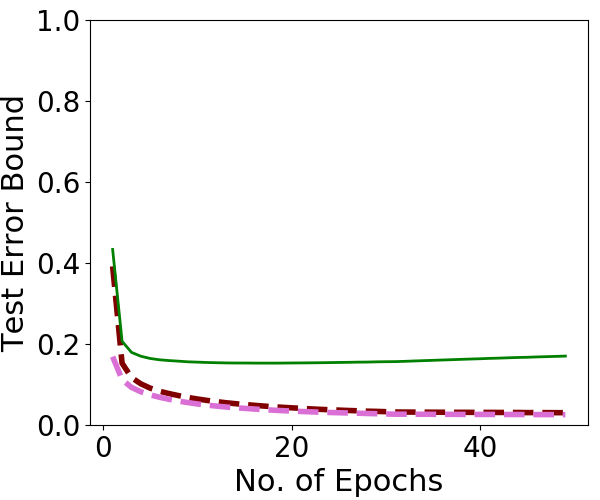}
 }
 \subfigure[MNIST, $\alpha_t = 0.01$]{
 \includegraphics[width=0.3\textwidth]{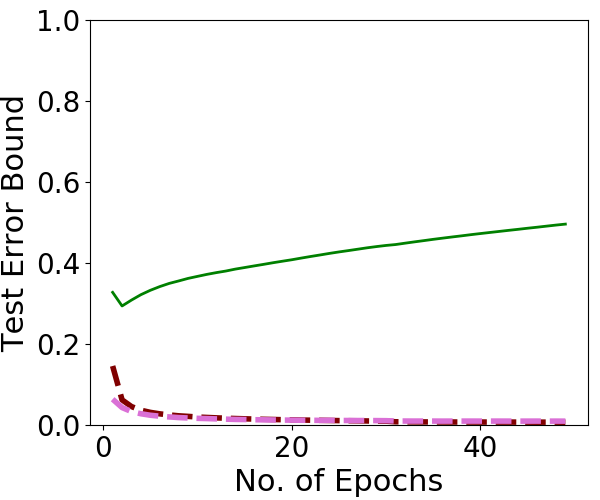}
 }
  \subfigure[MNIST, Our Bound]{
 \includegraphics[width=0.3\textwidth]{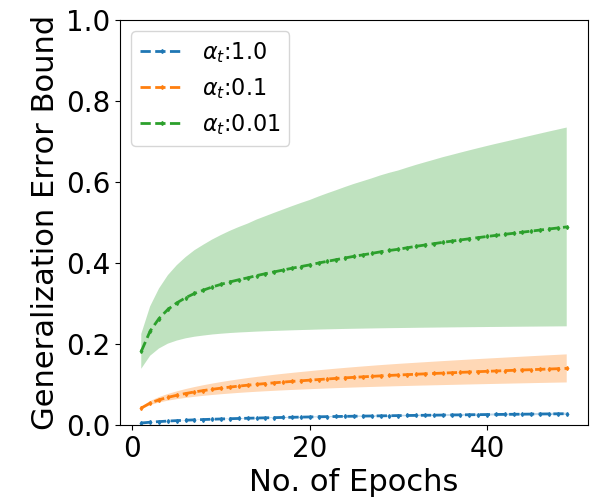}
 }  
 \subfigure[MNIST, Gradient Discrepancy]{
 \includegraphics[width=0.3\textwidth]{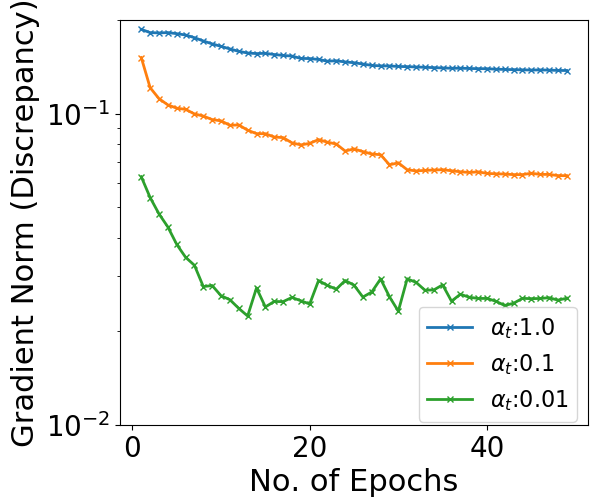}
 }
 
 \subfigure[Fashion, $\alpha_t = 1.0$]{
 \includegraphics[width=0.3\textwidth]{ICLR2022/images/smallcnn-fashion-bounds-noisy-signSGD-c.png}
 }
 \subfigure[Fashion, $\alpha_t = 0.1$]{
 \includegraphics[width=0.3\textwidth]{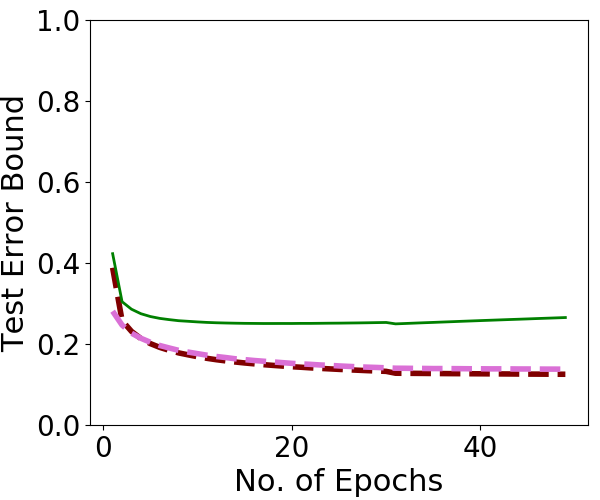}
 }
 \subfigure[Fashion, $\alpha_t = 0.01$]{
 \includegraphics[width=0.3\textwidth]{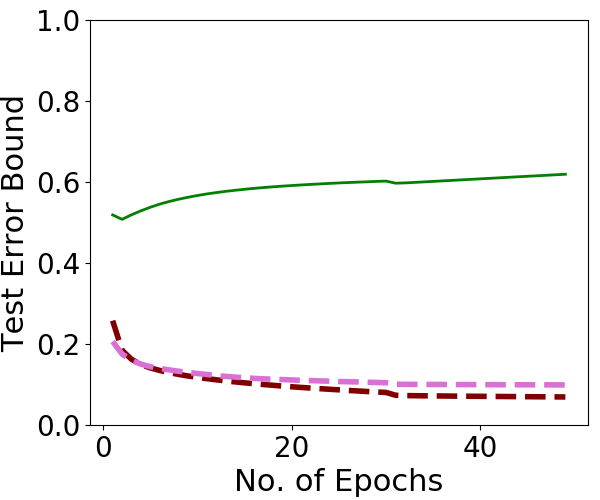}
 }
  \subfigure[Fashion, Our Bound]{
 \includegraphics[width=0.3\textwidth]{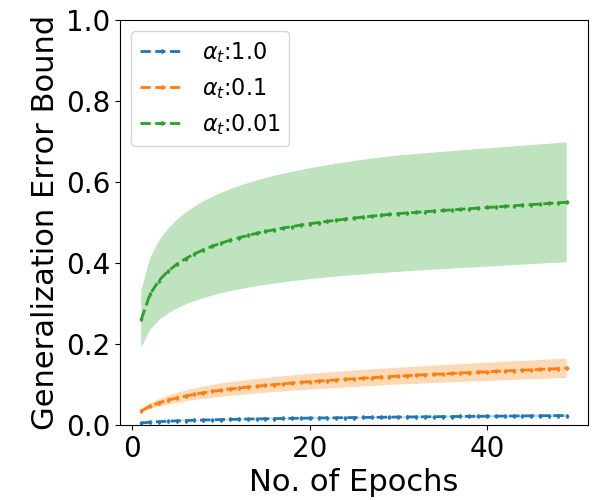}
 }  
 \subfigure[Fashion, Gradient Discrepancy]{
 \includegraphics[width=0.3\textwidth]{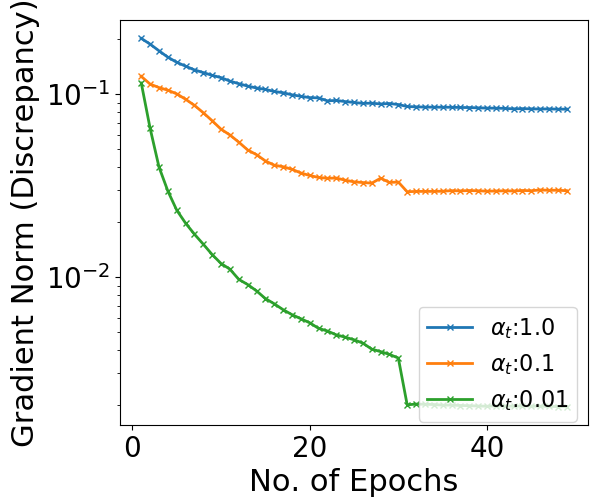}
 }  
 
\caption[]{Numerical results for training CNN on MNIST and Fashion-MNIST using Noisy Sign-SGD. Increasing $\alpha_t$ leads to a tighter generalization bound even though it leads to a slightly larger gradient discrepancy. 
}
\label{fig:bound_noisy_signsgd}
\end{figure*}